\documentclass[%
 aip,
 apl,%
 amsmath,amssymb,
 reprint,%
]{revtex4-2}

\usepackage{graphicx}% Include figure files
\usepackage{dcolumn}% Align table columns on decimal point
\usepackage{bm}% bold math
\usepackage{subfigure}
\usepackage{subcaption}
\usepackage{ragged2e}
\usepackage{caption}
\usepackage{comment}
\usepackage{hyperref}
\usepackage{xcolor}
\usepackage{makecell}
\usepackage{booktabs}

\begin{document}

\preprint{AIP/123-QED}

\title{{\normalfont\bfseries\itshape RTNinja}: A generalized machine learning framework for analyzing random telegraph noise signals in nanoelectronic devices}

\author{Anirudh Varanasi}
\email{Anirudh.Varanasi@imec.be}
 \affiliation{imec, 3001 Leuven, Belgium}
 \affiliation{Department of Materials Engineering, KU Leuven, 3001 Leuven, Belgium}
\author{Robin Degraeve}%
 \affiliation{imec, 3001 Leuven, Belgium}
\author{Philippe Roussel}
\affiliation{imec, 3001 Leuven, Belgium}
\author{Clement Merckling}
\affiliation{imec, 3001 Leuven, Belgium}
\affiliation{Department of Materials Engineering, KU Leuven, 3001 Leuven, Belgium}

%\date{\today}% It is always \today, today, but any date may be explicitly specified

\begin{abstract}
Random telegraph noise is a prevalent variability phenomenon in nanoelectronic devices, arising from stochastic carrier exchange at defect sites and critically impacting device reliability and performance. Conventional analysis techniques often rely on restrictive assumptions or manual interventions, limiting their applicability to complex, noisy datasets. Here, we introduce \textit{RTNinja}, a generalized, fully automated machine learning framework for the unsupervised analysis of random telegraph noise signals. \textit{RTNinja} deconvolves complex signals to identify the number and characteristics of hidden individual sources without requiring prior knowledge of the system. The framework comprises two modular components: \textit{LevelsExtractor}, which uses Bayesian inference and model selection to denoise and discretize the signal, and \textit{SourcesMapper}, which infers source configurations through probabilistic clustering and optimization. To evaluate performance, we developed a Monte Carlo simulator that generates labeled datasets spanning broad signal-to-noise ratios and source complexities; across 7000 such datasets, \textit{RTNinja} consistently demonstrated high-fidelity signal reconstruction and accurate extraction of source amplitudes and activity patterns. Our results demonstrate that \textit{RTNinja} offers a robust, scalable, and device-agnostic tool for random telegraph noise characterization, enabling large-scale statistical benchmarking, reliability-centric technology qualification, predictive failure modeling, and device physics exploration in next-generation nanoelectronics.
\end{abstract}

\maketitle

\section{Introduction}

For over six decades, the semiconductor industry has been at the forefront of technological advancements, driving exponential performance growth and transforming society at a rapid pace. This progress has been propelled by the consistent miniaturization of device dimensions, toward the nanometer scale, with each new technology node \cite{cmosscale}. However, this advancement has come with significant challenges. Each technological node has introduced new scientific obstacles, compelling researchers to develop innovative solutions to overcome these limitations. One such obstacle is random telegraph noise (RTN) which has been observed by the device research community since the 1980s~\cite{ralls1984,uren1985}. This noise is observed in transistors and capacitors, which are fundamental building blocks of logic and memory devices. RTN is an intrinsic noise phenomenon observed as random discrete level fluctuations in the progression of device characteristics, such as drain current and gate current. As a result, RTN leads to non-ideal device behavior, degrades long-term performance, and induces failures~\cite{yaney1987meta,mueller1996conductance,mueller1997statistics,fukuda2007random,qiu2015impact,goda2015time}. The origin of RTN is attributed to defects in the gate dielectric material that are capable of exchanging charge carriers (electrons or holes) with the device substrate via quantum mechanical tunneling~\cite{grasser2020noise}. This mechanism is then observed as charge fluctuations in device characteristics. Interest in this phenomenon grew when it was investigated in flash memories, revealing current fluctuations of up to 60\% and threshold-voltage shifts as high as 700~mV in 90-nm technology node devices~\cite{fantini2007,tega2006anomalously,kurata2007random,spinelli2021random}. Subsequently, over the last two decades, RTN has become a vital phenomenon to investigate for any nanoelectronic device as it serves a dual purpose: (i)~an important mechanism for investigating the microscopic physics of carrier-defect interactions, while (ii)~also posing as an indicator of the reliability and stochastic variability in performance. Moreover, RTN affects not only the reliability of individual devices, but also the overall reliability of circuit designs~\cite{kaczer2016defect,simoen2011random,matsumo2012,luo2015}. To effectively tackle device technology issues and benchmark different technologies, the development of advanced RTN analysis frameworks is essential, as this leads to a robust understanding of the physical and statistical aspects. 

Over the years, numerous techniques have been developed to analyze RTN signals and extract information such as the observed number of individual RTN sources and their corresponding activities. Typically, but not exclusively, an individual RTN source is either a single defect or a chain of defects in nanoelectronic devices. Basic analysis techniques include histograms and time-lag plots, with various adaptations of time-lag plots documented in the literature~\cite{yuzhelevski2000random,nagumo2009new,nagumo2010statistical,martin2014new}. More advanced approaches, such as the Canny edge detection algorithm, hidden Markov models, Kohonen networks, and deep-learning-assisted methods, have also been developed~\cite{stampfer2018characterization,stampfer2020semi,miki2012statistical,puglisi2014factorial,grill2019electrostatic,xiao2024new,gonzalez2020neural,xu2024deep}. Some fundamental and practical limitations encountered across these techniques, though not necessarily present in every single one, include the following:\\
(i) Applicable only for simple signals with very high signal-to-noise ratios (SNR) with reasonable sampling of RTN fluctuations.\\
(ii) Extracted information limited to observed levels and not individual sources.\\
(iii) Applicable to specific cases of RTN signals, such as the activity of single defects with exponentially distributed activity.\\
(iv) Analysis limited to signals involving limited sources, as complexity of more than two sources is often not feasible.\\
(v) Manual interaction is required, meaning the number of sources and their activity distribution parameters are user-defined inputs. This is extremely challenging for signals with more than two sources, for signals with low SNR, or for analyzing datasets at large scale.\\
(vi) Artifacts are introduced since some techniques must account for all possible levels originating from observable sources, even if some levels are not observed during the measurement window.\\
(vii) Computation time for analysis increases exponentially with the number of levels. 

With recent rapid advancements in data analytics and machine learning, which are beneficial across various scientific and technological domains, the analysis of RTN signals can be classified as an unsupervised learning problem of a descriptive nature. Descriptive types of ML algorithms are designed to conduct exploratory data analysis and extract meaningful information (in-depth characteristics). This contrasts with commonly used predictive ML algorithms, which make predictions about future outputs based on unseen inputs after training on a set of known inputs and outputs. In the context of RTN signals, the observable data typically represent the progression of a complex signal as a function of time, which is a convolution of unknown sources, their activities, and intrinsic experimental noise. In this work, we propose a machine learning (ML) framework\textemdash \textit{RTNinja}\textemdash that can conduct automated data-driven analysis of RTN signals. As shown in Figure~\ref{fig_focus}, an observable RTN signal is provided as the input information and \textit{RTNinja} extracts the hidden information comprising individual sources along with their corresponding activity. Compared to existing RTN analysis techniques, \textit{RTNinja} has the following major advantages:\\
(i) It can deconvolve complex signals constituting up to seven or more individual sources (subject to SNR resolution).\\
(ii) The process is completely automated, where decision-making is primarily driven by data, without the need to predefine the number of sources or their activity distribution parameters.\\
(iii) It can identify individual sources with limited activity, meaning sources that are undersampled in the observation window.\\
\textit{RTNinja} is a generalized framework as it can analyze RTN signals of any nanoelectronic device since no tailor-made assumptions are taken about the underlying sources and their activity distribution parameters. In the following sections, different modules of the \textit{RTNinja} framework are explained in detail by means of an example dataset. Furthermore, the robustness of the framework is discussed by applying it to a large number of synthetic (and therefore labeled) data with varying SNR levels and data complexity. Finally, given that no ML framework can be invincible, a comprehensive outline of the limitations and scope of application is provided.

%\begin{comment}
	
\section{Monte Carlo-based RTN Simulator}
\begin{figure*} %[htbp!]
	%\centering
	\subfigure[]{\includegraphics[width=0.75\textwidth]{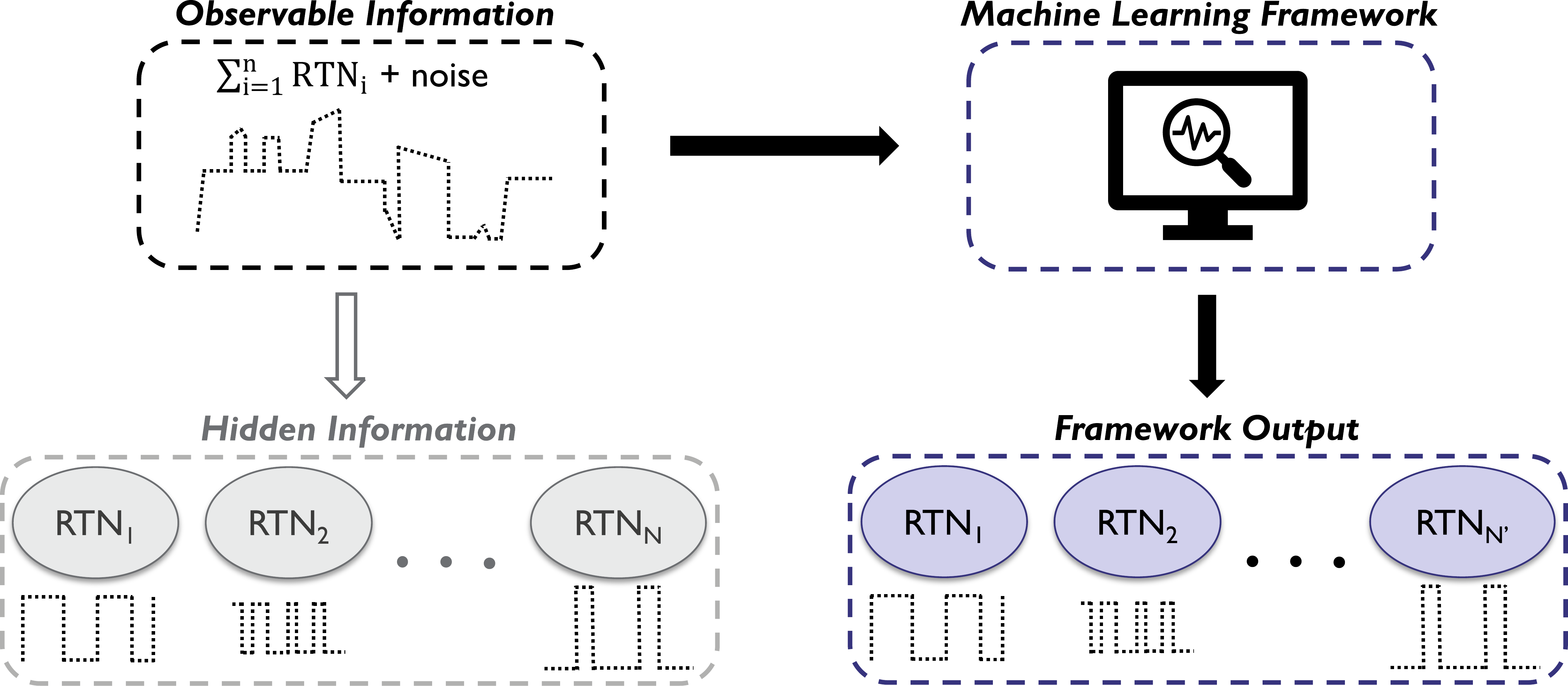}\label{fig_focus}}
	\vfill
	\vspace{-1.2em}
	\subfigure[]{\includegraphics[width=0.63\textwidth]{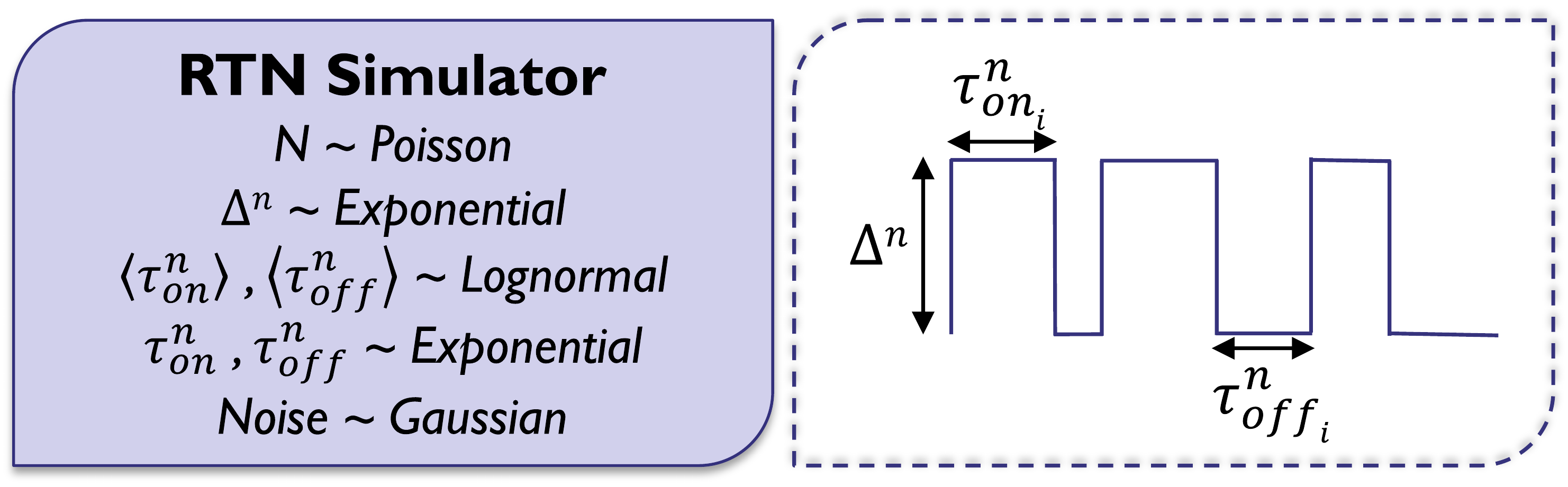}\label{fig_mcsim}}
	\hfill
	\subfigure[]{\includegraphics[width=0.33\textwidth]{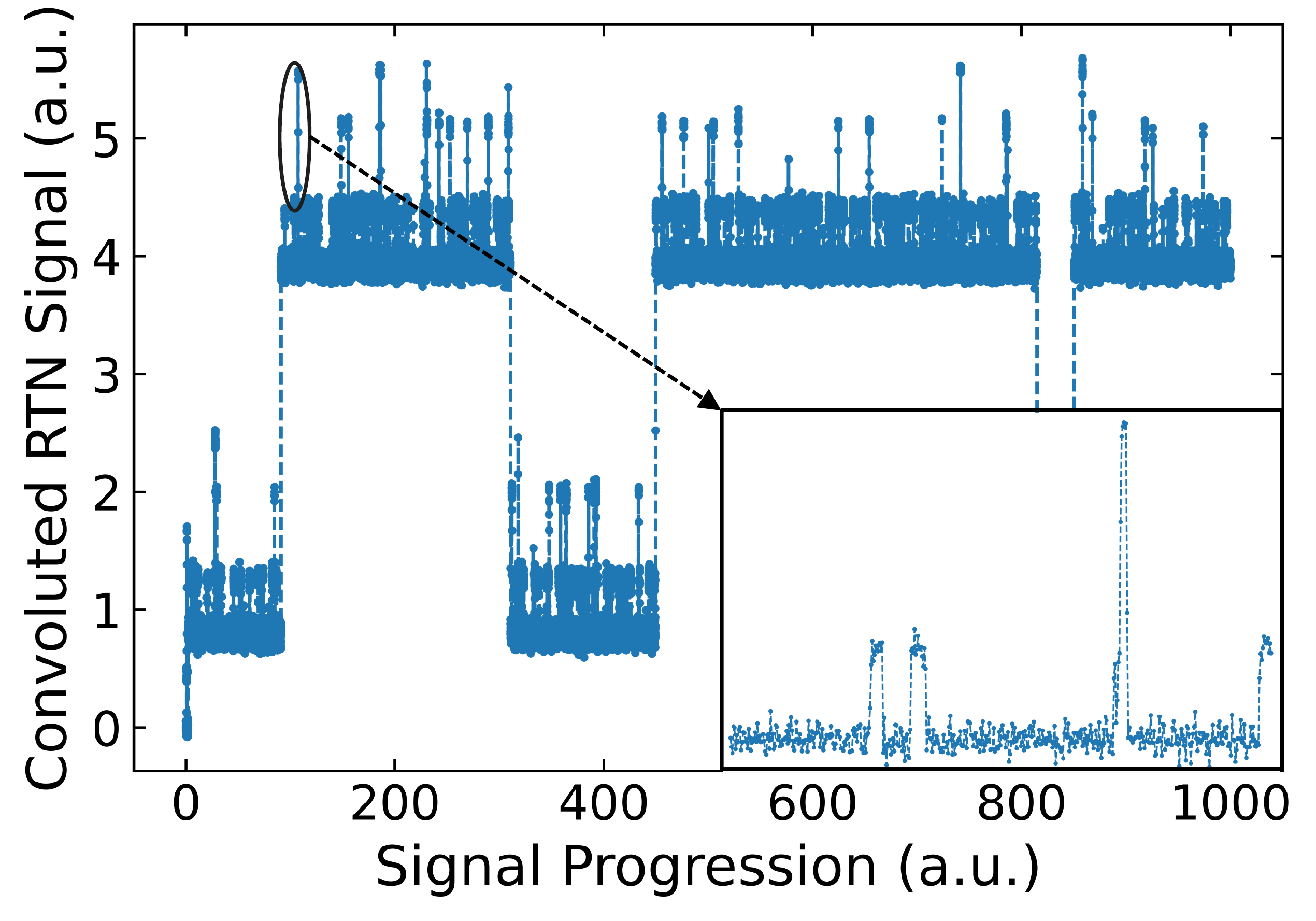}\label{fig_mcsimexample}}
	\vfill
	\vspace{-1.2em}
	\subfigure[]{\includegraphics[width=0.8\textwidth]{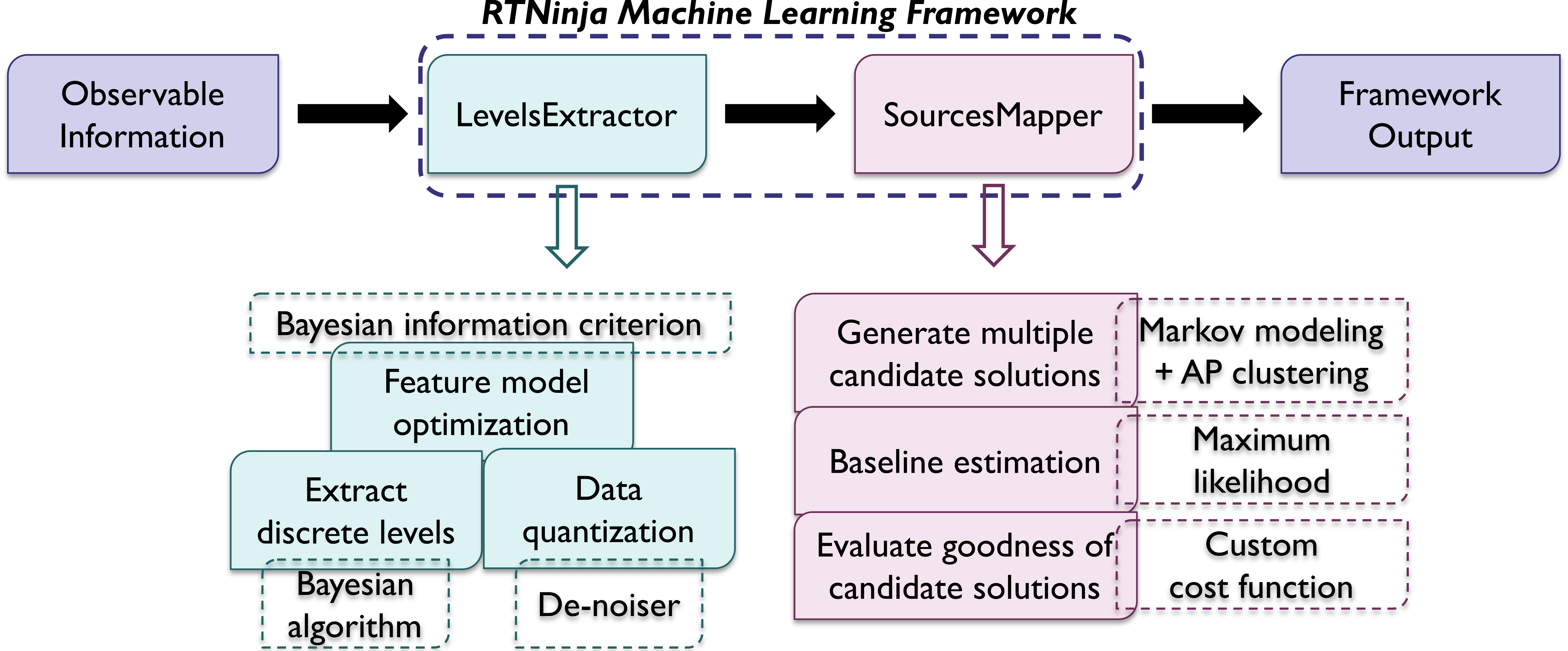}\label{fig_mlflow}}
	\vspace{-1.2em}
	\caption{\justifying(a)~Schematic illustration of RTN signal analysis, a descriptive type of unsupervised learning problem, using the \textit{RTNinja} machine learning framework. Input to the framework is a convoluted RTN signal resulting from the activity of a hidden number of individual sources, along with intrinsic noise fluctuations. Without any predefined estimates of the underlying information, \textit{RTNinja} analyzes the signal and outputs the total number of individual sources and their corresponding activity. Ideally, the output is an exact reproduction of the input sources ($N = N'$) with their corresponding activity. (b)~Monte Carlo-based RTN simulator. The number of sources~($N$) is sampled from a Poisson distribution, where each source has three attributes: amplitude ($\Delta^n$), on-state activity, and off-state activity. Both state activities follow exponential distributions with the primary distribution parameters, on-state mean value~($\langle\tau_{on}^{n}\rangle$) and off-state mean value~($\langle\tau_{off}^{n}\rangle$), are sampled from a log-normal distribution. Extrinsic noise is added to the RTN signal in proportion to the overall signal magnitude. (c)~The example dataset is simulated for a total time of $10^{3}$ units (e.g., seconds) at a sampling rate of 50 samples per unit. This particular signal constitutes four sources, with amplitudes within the range ($4\times10^{-1}$,~$3$) and mean activities within the range ($3\times10^{-1}$,~$5\times10^{2}$). (d)~Workflow of the \textit{RTNinja} machine learning framework. Observable information, that is, an experimental RTN signal, is provided as input data to the framework. \textit{RTNinja} analyzes this information and outputs the hidden individual RTN sources and their corresponding progression characteristics. The framework consists of two modules: \textit{LevelsExtractor} and \textit{SourcesMapper}. \textit{LevelsExtractor} identifies discrete constant levels and builds an optimal model of the input data. \textit{SourcesMapper} finds the most likely solution, that is, the number of sources along with their amplitudes and activities, that explains the optimal model produced by \textit{LevelsExtractor}. Each module consists of three sub-modules, with the objectives and underlying approaches of each sub-module clearly outlined.}%\label{fig_concept}
\end{figure*}

RTN signal analysis exemplifies an unsupervised learning problem, where there is no availability of ground-truth labels for the input data. In such scenarios, it is challenging to objectively evaluate the performance of an ML framework in identifying underlying patterns or structures in the data. Researchers frequently rely on heuristic methods, such as clustering quality metrics or visual inspection, to assess the performance of ML frameworks \cite{fergus2022performance}. These methods are less reliable and can lead to subjective evaluations, unlike in supervised learning problems where the ground-truth labels are available, which enables the direct comparison between ground truth and predicted output, facilitating a more straightforward and objective performance evaluation. In the context of RTN signals, it is possible to convert this unsupervised learning problem into a supervised learning scenario. It is a common practice in the device research community to generate physics-based or statistical simulations of RTN signals, especially for individual defect studies, with predefined characteristics~\cite{grasser2020noise,martin2014new,puglisi2014factorial,xiao2024new,xu2024deep,vecchi2023unified}. In this work, we adopted the practice of building a Monte Carlo (MC) simulator that generates synthetic RTN signals, with all underlying attributes sampled from specific statistical distributions, as illustrated in Figure~\ref{fig_mcsim}. The choice of these distributions aims at mimicking common research studies in the device research community, particularly those focused on understanding the impact of single defects on transistor characteristics. However, the \textit{RTNinja} framework is designed to be agnostic to the underlying distributions. Figure~\ref{fig_mcsimexample} presents an example dataset generated by the MC-based RTN simulator, which is used to explain the \textit{RTNinja} ML framework.

%\captionsetup[subfigure]{font={normal}, skip=0pt, margin=0cm, singlelinecheck=false}

\section{{\normalfont\bfseries\itshape \NoCaseChange{RTNinja}} Framework}
\subsection{Overview}
The workflow representation of the \textit{RTNinja} ML framework is illustrated in Figure~\ref{fig_mlflow}. \textit{RTNinja} consists of two modules, namely, \textit{LevelsExtractor} and \textit{SourcesMapper}. 

The \textit{LevelsExtractor} module is designed to optimally model observable input data by identifying transitions between discrete constant levels. This process involves three key sub-modules. Initially, an algorithm based on Bayesian inference is designed to determine a set of discrete levels where each level is characterized by a Gaussian distribution defined by its mean and standard deviation. Following this, the input data are reconstructed to reflect transitions between these levels, effectively filtering outliers or experimental variations through a de-noising condition. The combined application of these two sub-modules results in a characteristic feature model in which the input data are represented by an integer number of levels and the transitions between them. This model is refined based on a standard deviation value, which is estimated a priori from the data and used as an input for the Bayesian algorithm. Multiple feature models are generated by varying the standard deviation value, and the optimal model is selected using the Bayesian information criterion (BIC), ensuring the most accurate representation of the signal evolution. 

The \textit{SourcesMapper} module aims at identifying the optimal set of $N$ sources that can physically explain the observed feature model. To achieve this, the module addresses the challenge of associating each observed level with a specific state configuration of the individual sources, where each individual source is treated as a two-state entity comprising an on-state and an off-state. A source's amplitude is defined by the difference between the mean values of two levels, while a state configuration refers to a superposition of the states of $N$ sources. The process is carried out by three sub-modules. First, the available information from \textit{LevelsExtractor}, including the list of possible differences between all levels and the transitions between levels based on the quantized data, is utilized. Markov modeling is employed to extract the underlying transition probability matrix, and affinity propagation (AP) clustering is used to generate multiple  $N$-source sets, referred to as candidate solutions. Second, for each candidate solution, the optimal baseline is estimated using a maximum likelihood method to align the state configurations with the Bayesian levels. Finally, a custom-designed cost function is developed to identify the optimal solution from all candidate solutions. The modules and their underlying sub-modules are explained in detail in the following sections using an example dataset.

\subsection{\normalfont\bfseries\itshape LevelsExtractor}
The \textit{LevelsExtractor} module is designed to effectively capture and model the transitions within observable input data by identifying discrete constant levels. The Bayesian algorithm and De-noiser sub-modules are applied together to construct a characteristic feature model of the input signal. By systematically varying the standard deviation parameter, estimated a priori from the data, multiple feature models are generated. To ensure the most accurate representation of the signal evolution, the optimal model is selected based on the Bayesian Information Criterion (BIC), which balances model complexity with data fidelity. 

\begin{figure*} %[htbp!]
	%\centering
	\subfigure[]{\includegraphics[width=0.61\textwidth]{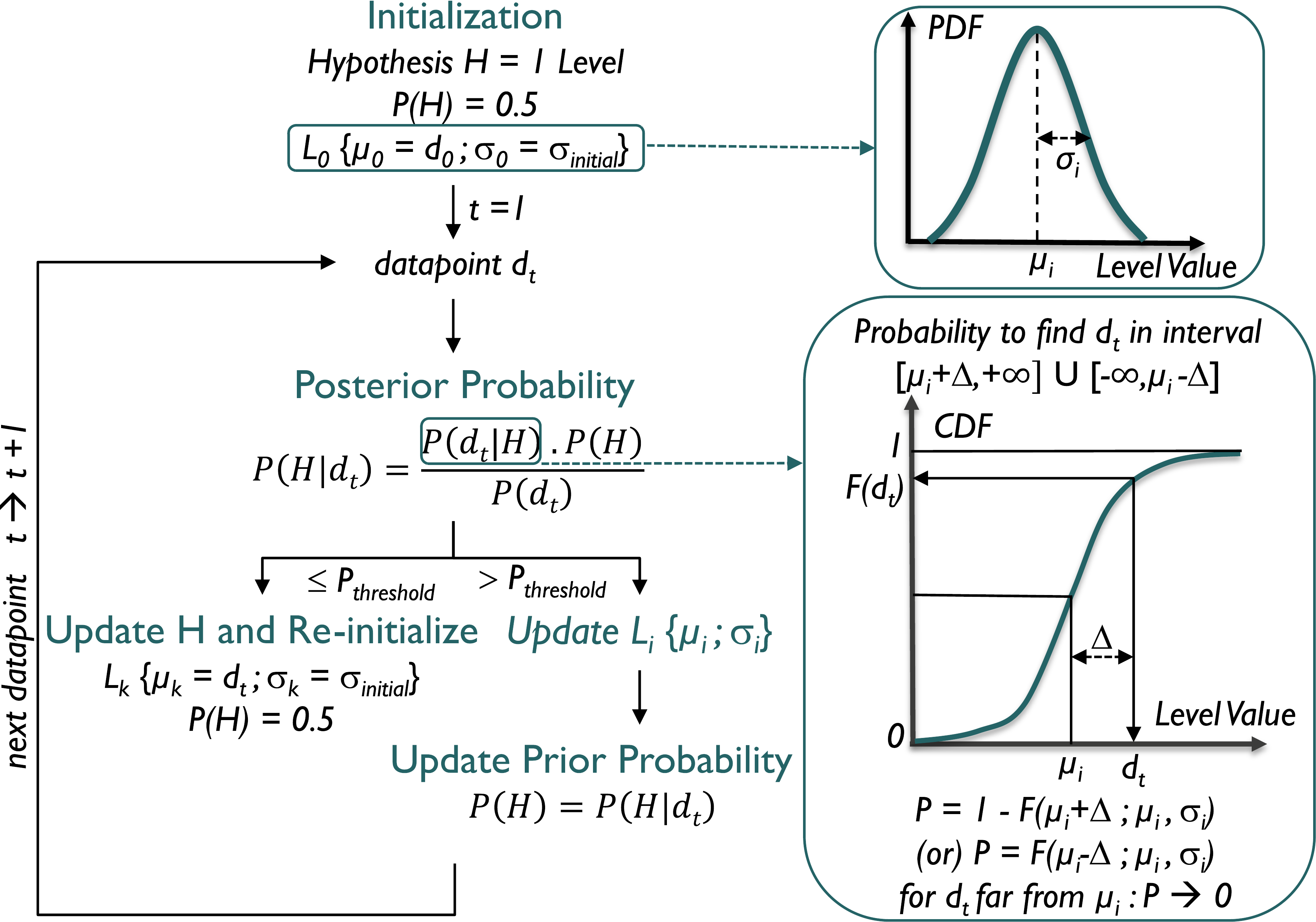}\label{fig_bayesian}}
	\hfill%
	\subfigure[]{\includegraphics[width=0.38\textwidth]{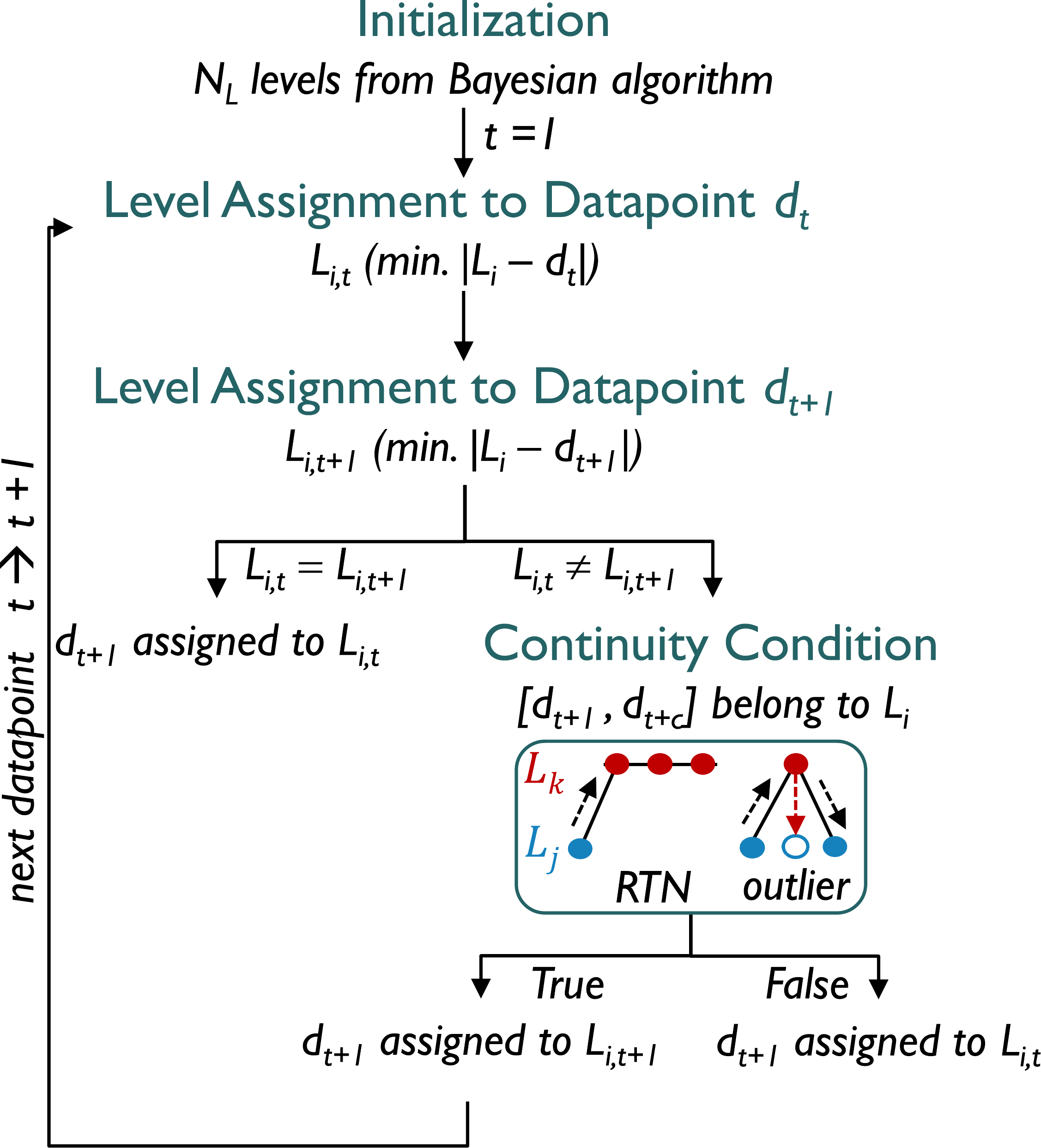}\label{fig_noisefilter}}
	%\vfill
	%\vspace{-1em}
	\subfigure[]{\includegraphics[width=0.4\textwidth]{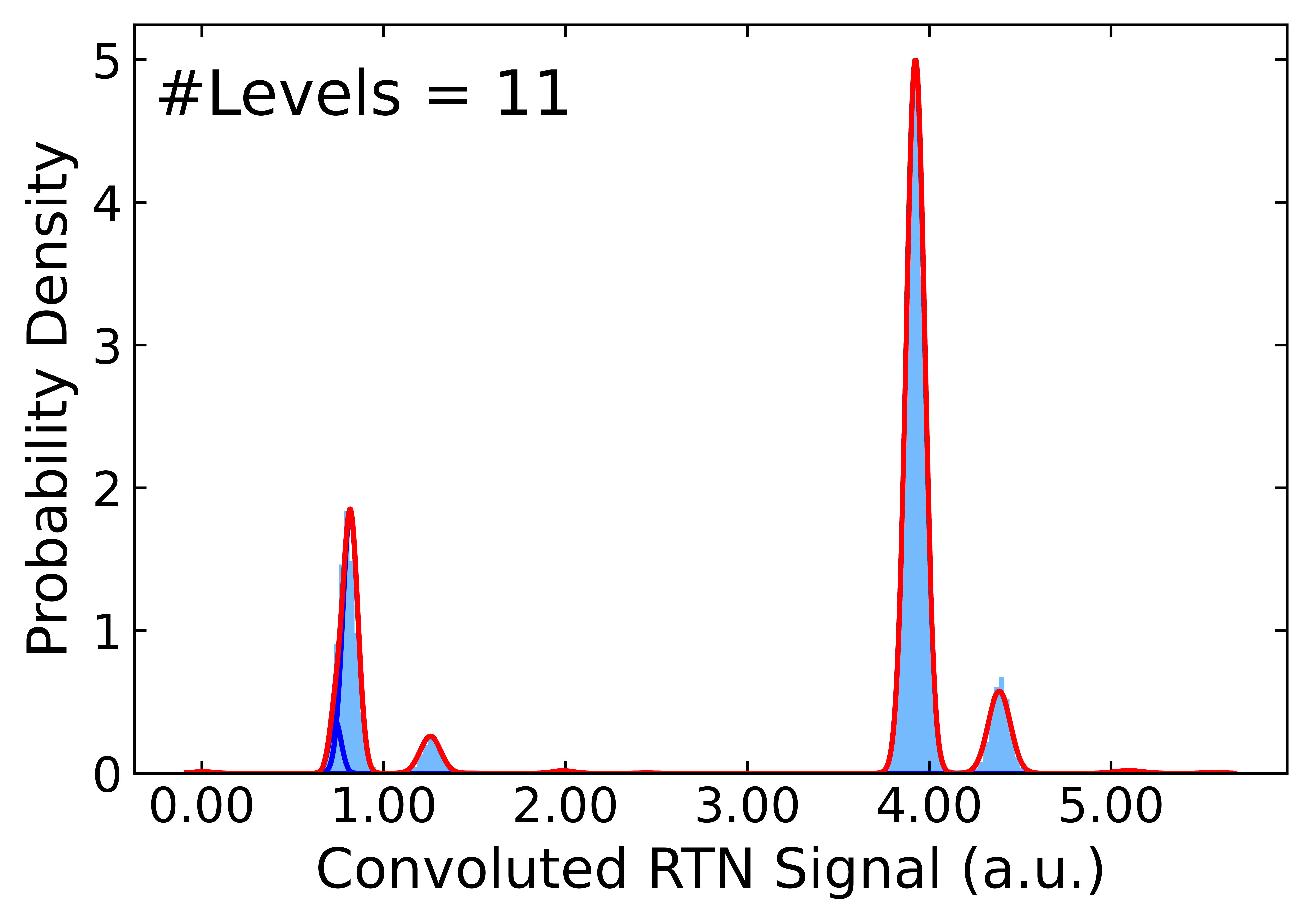}\label{fig_levels}}
	%\hfill%
	\subfigure[]{\includegraphics[width=0.4\textwidth]{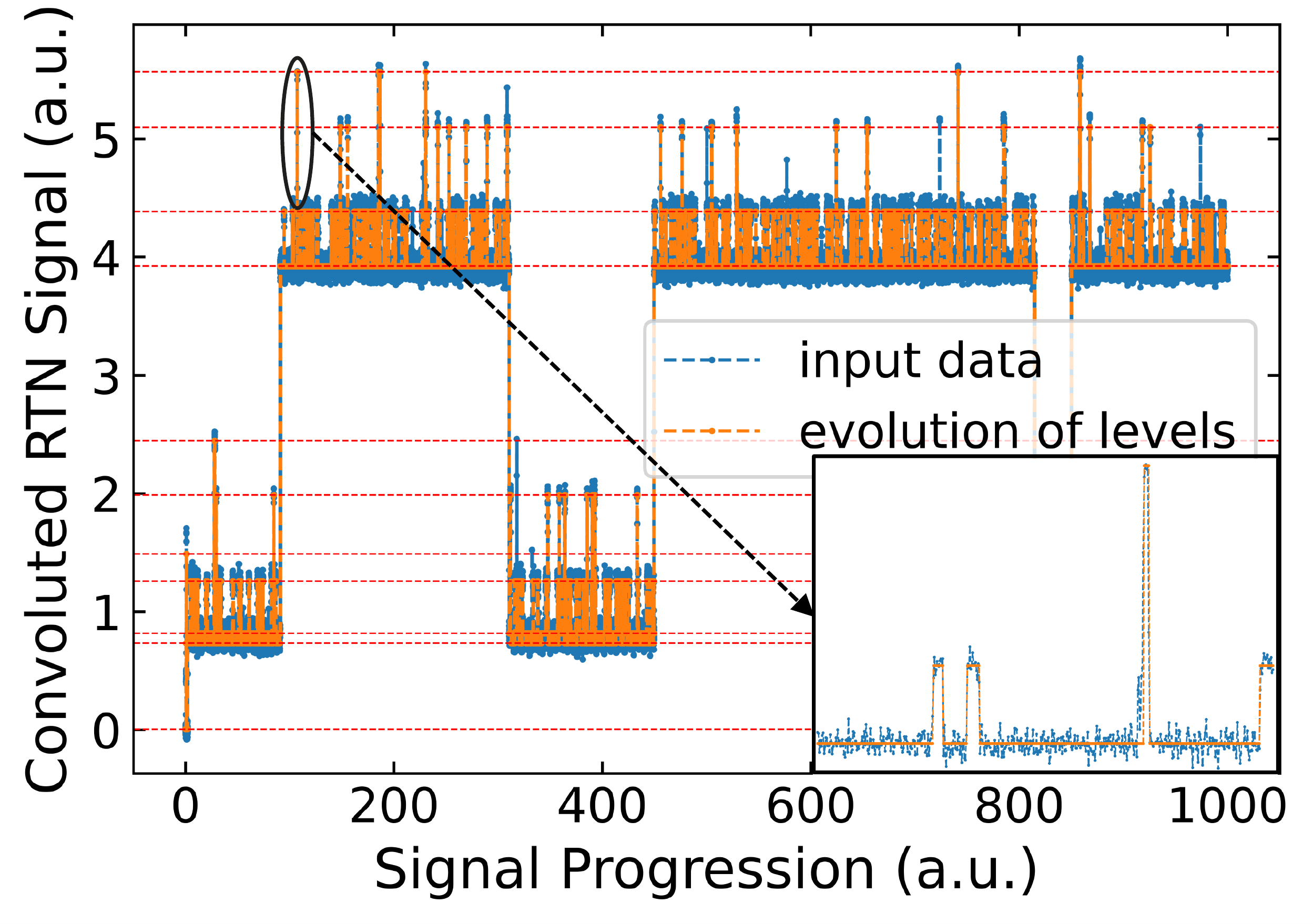}\label{fig_levelsrecon}}
	%\vfill
	\subfigure[]{\includegraphics[width=\textwidth]{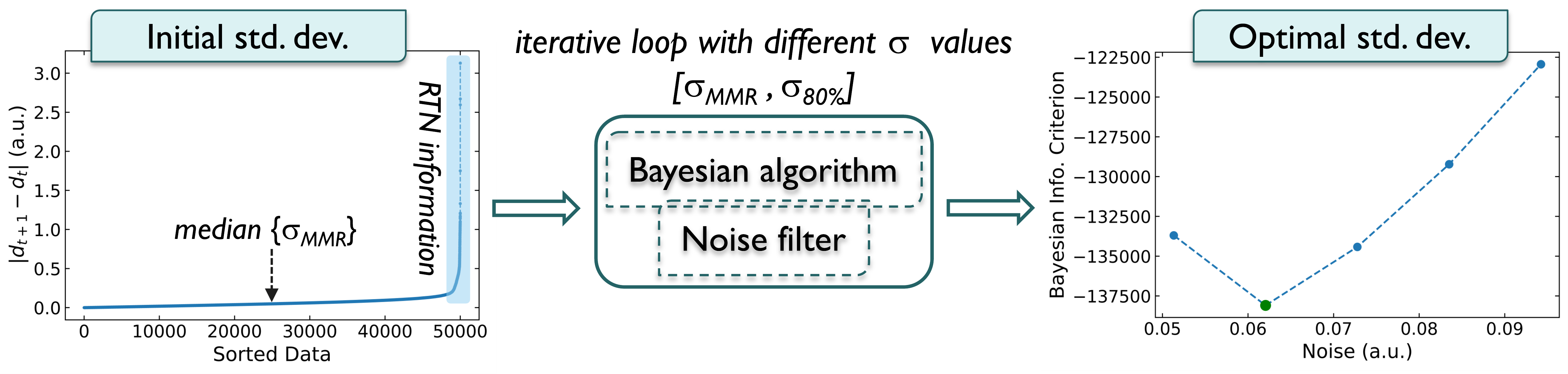}\label{fig_sigopt}}
	\vspace{-1em}
	\caption{\justifying Working details and results of the \textit{LevelsExtractor} module. (a)~Flowchart of the Bayesian algorithm, which identifies the most suitable hypothesis (number of levels and their model parameters) based on observed data using Bayes' theorem. The hypothesis is updated if the posterior probability falls below a user-defined threshold (\(10^{-15}\) in this case). The conditional probability, \(P(d_t \mid H)\), is calculated using a custom CDF-based proximity estimator, representing the likelihood of a data point belonging to its nearest level (\(P = 1\) if \(d_{t} = \mu_{i}\)). (b)~Flowchart of the De-noiser, which assigns data points to their nearest levels, applying a custom continuity condition (hyperparameter set to $c$~=~3 data points) to ensure statistical significance for transitions identified as RTN signatures. (c)~The Bayesian algorithm extracts 11 levels, with the total weighted distribution of these levels (red) overlapping the dataset histogram (blue), and \(\mu\) values of levels represented by dashed lines. (d)~Input data are successfully reconstructed, with \(\mu\) values of the 11 levels represented by dashed lines. (e)~Standard deviation is estimated a priori using a median moving range method (\(\sigma_{MMR}\)). While a minority of the input data contains the RTN transition information, the remaining portion corresponds to experimental noise at a constant level, leading to \(\sigma_{MMR}\) estimation. Multiple feature models are built by varying the initial \(\sigma\) value input to the Bayesian algorithm. BIC values are computed for all models, where the optimal \(\sigma\) value and the corresponding feature model are determined by the lowest BIC value. This does not necessarily coincide with \(\sigma_{MMR}\) since fast and small-amplitude RTN sources might remain undetectable.}\label{fig_levelsextractor}
\end{figure*}

Figures~\ref{fig_bayesian}~and~\ref{fig_noisefilter} illustrate the flowcharts for the Bayesian algorithm and the De-noiser, respectively. The Bayesian algorithm initiates with the formulation of a hypothesis, starting with a single constant level, which is modeled as a Gaussian distribution characterized by two parameters: mean (\(\mu\)) and standard deviation (\(\sigma\)). These parameters are initialized using the first data point and the  \textit{a priori} estimated standard deviation value, with the prior probability set at 0.5. The next observed data point is then used to update the hypothesis through Bayes' theorem, which in turn updates the prior probability via the likelihood function. This function evaluates how well the observed data fits the present hypothesis, and by combining it with the prior probability, the posterior probability is obtained. Depending on whether the posterior probability exceeds or falls below a predefined threshold $P_{threshold}$ (i.e., a hyperparameter), the belief is updated either by adding a new level (if below the threshold) or by refining the existing hypothesis by updating the model parameters based on new evidence. This iterative approach leverages the posterior probability from each step as the prior for the next, continually improving parameter estimates and enhancing the understanding of \(\mu\) and \(\sigma\). Figure~\ref{fig_levels} demonstrates the application of the Bayesian algorithm on the example dataset, revealing that 11 levels are required to explain the full dataset, with four levels being predominant while the remaining levels have fewer data points.

%\captionsetup[subfigure]{font={normal}, skip=0pt, margin=0cm, singlelinecheck=false}

The levels extracted from the Bayesian algorithm are then used to reconstruct the input signal, with a hyperparameter acting as a constraint to filter out noise. The De-noiser flowchart begins by revisiting the first two data points and assigning them to their closest Bayesian level. A 'continuity' condition is imposed, whereby a transition between levels is considered a 'true' RTN only if a predefined number of data points $c$ (i.e., a hyperparameter) belong to the same level after the transition. This condition serves as a qualitative measure of stochasticity, as the likelihood of a transition being an outlier is low if three data points ($c$~=~3) are observed close to the same level post-transition. It is important to note that any hidden source with faster RTN activity will also be considered noise, and such data points will be forcefully assigned to the present level as if no transition occurred. The time resolution of activity that can be extracted is limited by this condition. For example, if the measurement sampling time is 20 ms, the fastest activity that can be captured is 60 ms, given the continuity constraint is set to three data points~($c$~=~3). This iterative process is used to reconstruct the full RTN signal [Figure~\ref{fig_levelsrecon}].

The Bayesian algorithm, in conjunction with the De-noiser, is employed to construct a characteristic feature model of the data. The only information required a priori is the initial standard deviation, which serves as input for the Bayesian algorithm. This initial value is estimated using a median-based 2-by-2 range standard deviation estimator applied to the entire observed dataset, as illustrated in Figure~\ref{fig_sigopt}. This value establishes the resolution threshold for the smallest detectable source amplitude and influences the spacing between adjacent levels. To construct the optimal model, the initial standard deviation value is systematically varied within an evenly spaced interval, ranging from \(\sigma_{MMR}\) to the \(80^{th}\) percentile in the 2-by-2 range-based data, generating multiple feature models. A critical consideration is ensuring that higher \(\sigma\) values are sufficiently separated from potential source amplitudes to avoid masking important RTN information. Once the feature model is developed for each \(\sigma\) value, the next step is to select the optimal model. This is done using the Bayesian Information Criterion (BIC), a statistical metric for model selection among a finite set of candidates. It is defined as:
\[BIC = -2\log(\hat{L}) + \log(n)d\] 
\[\text{where,} \quad
\log(\hat{L}) = -\frac{n}{2} \log(2\pi) - \frac{n}{2} \ln(\sigma^2) - \frac{\sum_{i=1}^n (y_i - \hat{y}_i)^2}{2\sigma^2},\] 
where $\hat{L}$ represents the likelihood of the model given the data, $n$ is the number of data points, and $d$ is the number of parameters (in this case, the number of Bayesian levels). The likelihood term, \(-2\log(\hat{L})\), evaluates the goodness of fit, where $y_i$ represents the observed data point and $\hat{y}_i$ denotes the corresponding reconstructed value. The complexity penalty term, \(\log(n)d\), mitigates overfitting by imposing a penalty on models with a greater number of parameters. The model with the lowest BIC is selected, as it provides the optimal trade-off between data fit and model complexity. The previously discussed Figures~\ref{fig_levels}~and~\ref{fig_levelsrecon} depict the optimal feature model associated with the lowest BIC value, offering the most accurate representation of the data.

\subsection{\normalfont\bfseries\itshape SourcesMapper}
The \textit{SourcesMapper} module is designed to establish the optimal $N$-source set that can explain the extracted 11 Bayesian levels along with the quantized signal (level-to-level transitions) information. An important assumption is made: each individual RTN source is considered to be independent and exhibits a two-state activity. This assumption implies that an $N$-source set can have a maximum of $2^{N}$ state configurations, representing the superposition of the states of the sources. The objective is to establish the optimal $N$ sources, which includes determining the sources' amplitudes (the difference between two distinct levels) and their activity, to achieve accurate reconstruction of the quantized data. A brute-force approach would involve trying all combinations of possible amplitudes and finding the most likely fit for the sequence of transitions. This approach, however, becomes a combinatorial problem and is computationally infeasible to solve. In addition, since a source's amplitude is relative to the levels, it is necessary to determine the optimal baseline value, which is the signal that would be observed when all $N$ sources are in the off-state.

%\captionsetup[subfigure]{font={normal}, skip=0pt, margin=0cm, singlelinecheck=false}
\begin{figure*}%[t!]
	\begin{minipage}{0.49\textwidth}
		\vspace{-8em}
		\subfigure[]{\includegraphics[width=\textwidth]{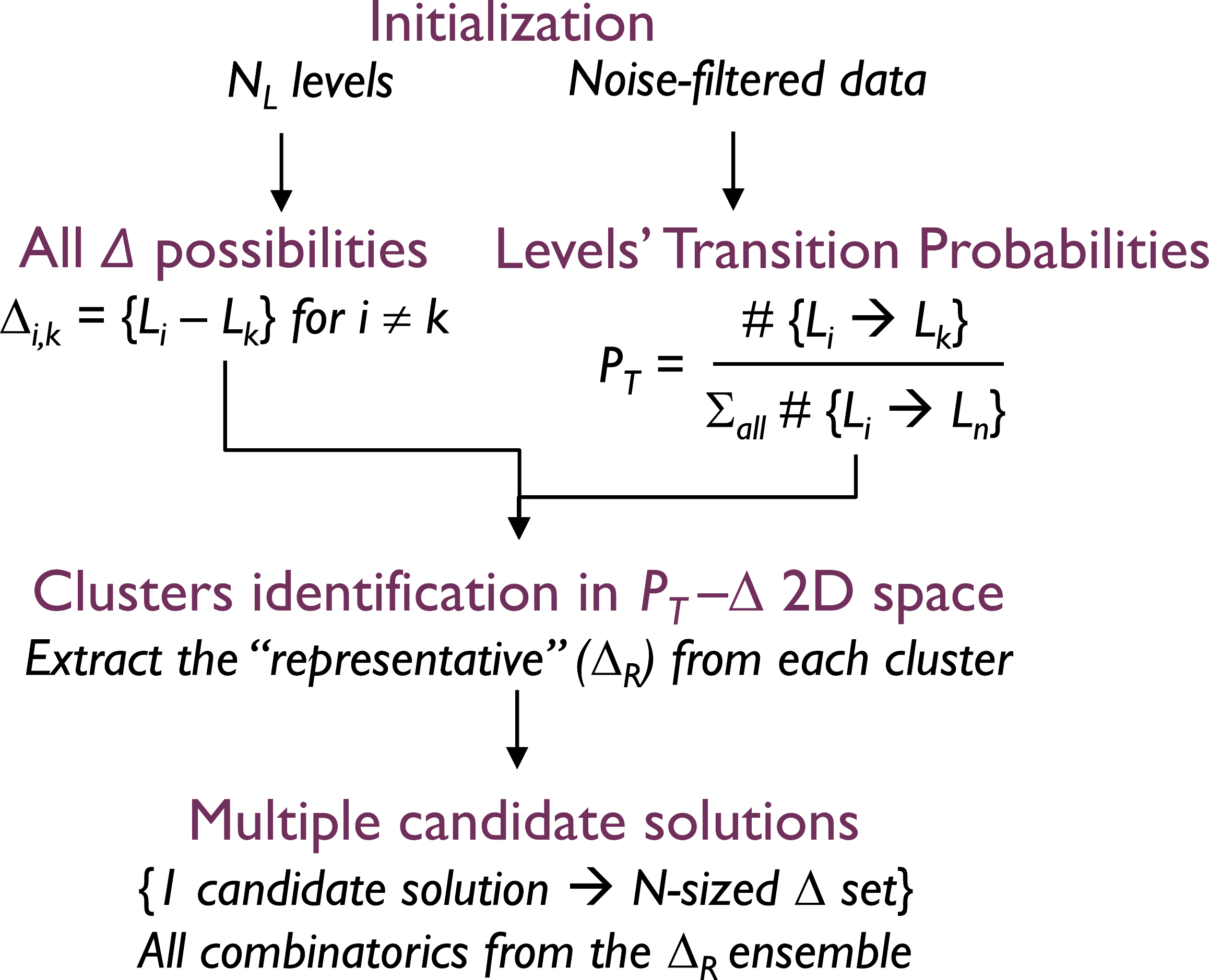}\label{fig_transclust}}
	\end{minipage}
	\begin{minipage}{0.47\textwidth}
		\subfigure[]{\includegraphics[width=0.85\textwidth]{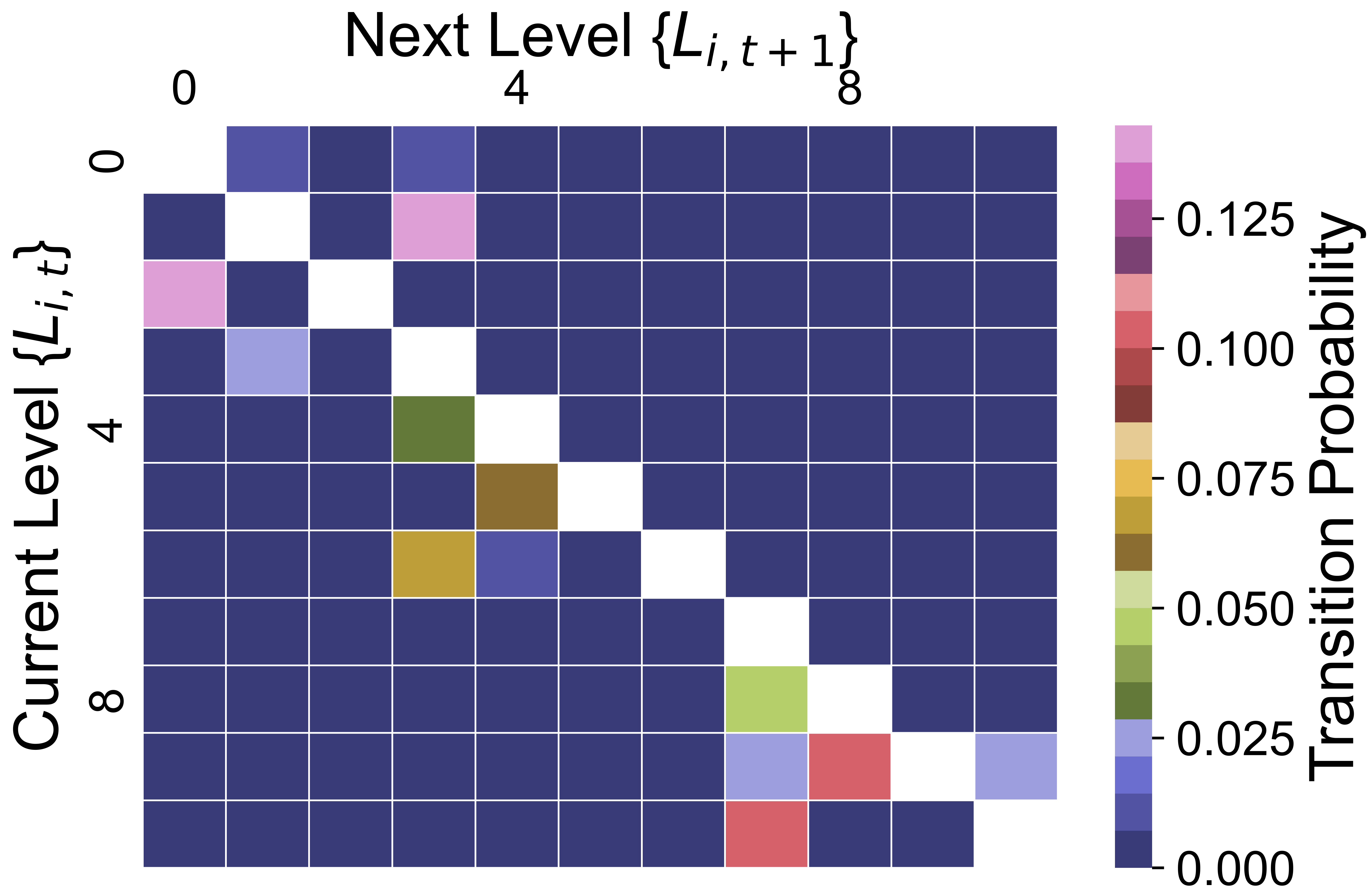}\label{fig_transmatrix}}
		\subfigure[]{\includegraphics[width=0.85\textwidth]{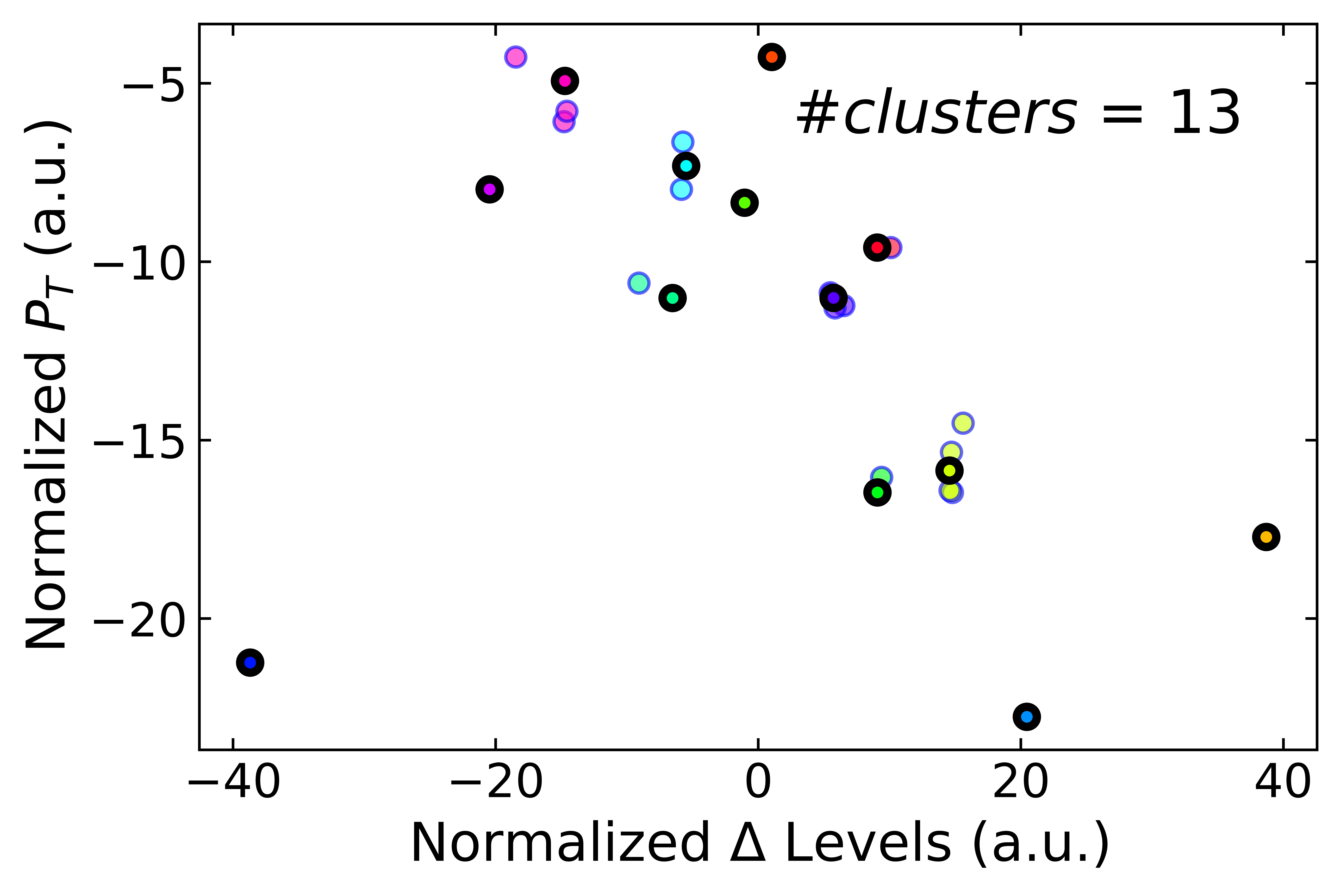}\label{fig_clustering}}
	\end{minipage}
	\begin{minipage}{0.49\textwidth}
		\vspace{-8em}
		\subfigure[]{\includegraphics[width=0.77\textwidth]{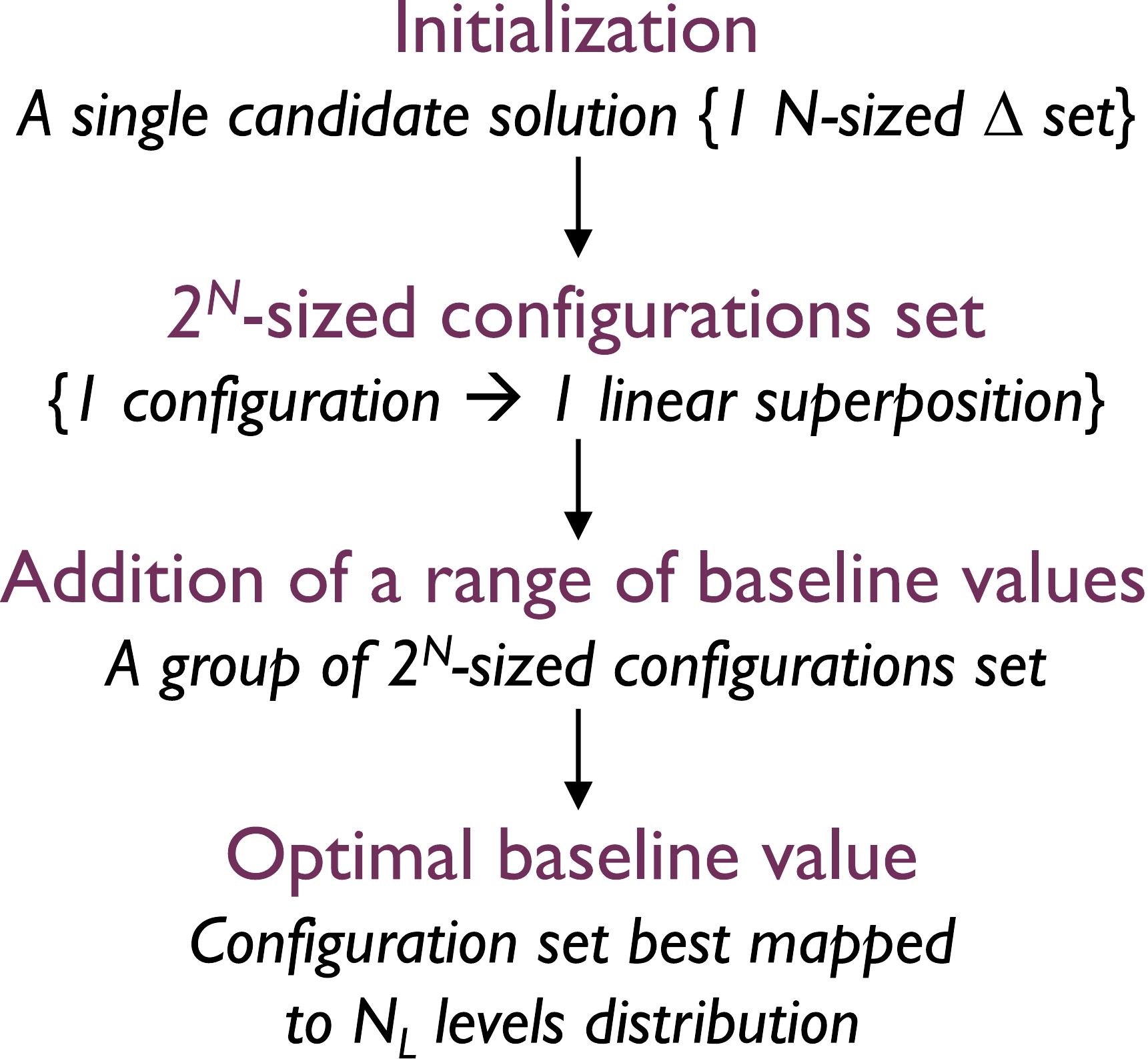}\label{fig_baseline}}
	\end{minipage}
	\begin{minipage}{0.47\textwidth}
		\subfigure[]{\includegraphics[width=0.85\textwidth]{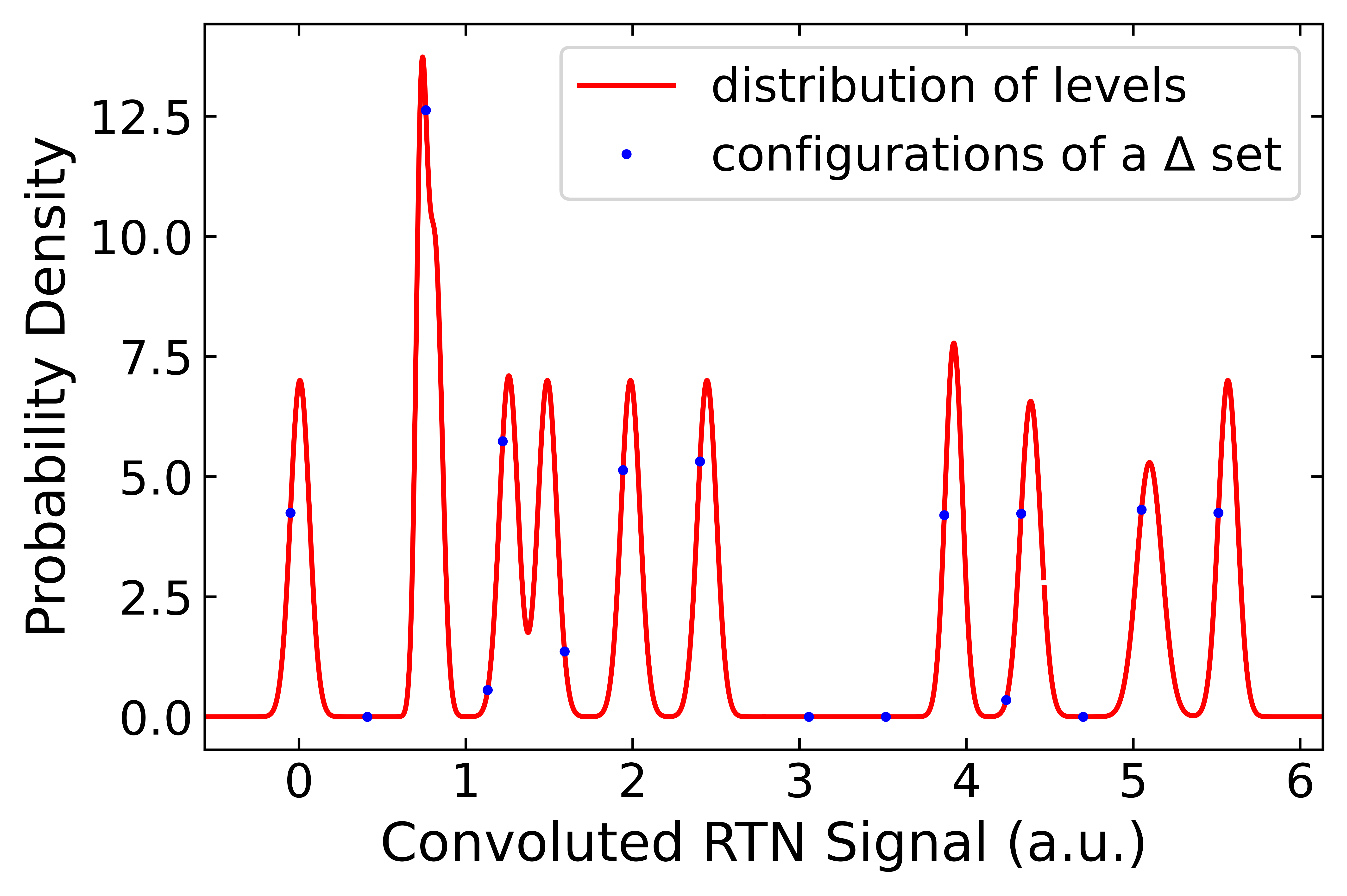}\label{fig_basemlle}}
	\end{minipage}
	\vspace{-1em}
	\caption{\justifying (a)~Flowchart illustrating the generation of multiple $N$-source sets. The output information from the \textit{LevelsExtractor} module is used to extract transition probabilities between levels and all possible source amplitudes. Clusters and their representative amplitudes are identified in the normalized \(P_{T}-\Delta\) space, enabling the combinatorial generation of $N$-sized \(\Delta\) sets. (b)~Markov transition matrix depicting the transition probabilities between the 11 Bayesian levels. Diagonal elements are masked for clarity, as they represent the high likelihood of remaining at the same level. The limited observability of transitions is evident in the broad distribution of activity rates associated with the hidden sources. (c)~Result of the AP clustering algorithm applied to the normalized \(P_{T}-\Delta\) space, revealing 13 clusters and their representative points (indicated with bold markers). (d)~Flowchart for baseline estimation. For each candidate solution, the corresponding $2^{N}$-configuration set is generated. A range of baseline values is tested for each set, with the optimal baseline defined as the one that best matches the total distribution of Bayesian levels, determined using a maximum log-likelihood approach. (e)~Example of a 16-configuration set with a specific baseline value derived from a candidate solution (4-source set), mapped onto the Bayesian levels distribution. The intersectional probability density values are used to compute the log-likelihood. Since only 11 levels are observed, some configurations may not align with any level, which does not pose an issue.}\label{fig_sourcesmapper}
\end{figure*}

The first step involves extracting meaningful features from the data generated by \textit{LevelsExtractor}, as outlined in Figure~\ref{fig_transclust}. The quantized data, representing level-to-level transitions, is modeled as a Markov process—a mathematical system in which transitions occur between states without memory of past states, a valid assumption for independent RTN signals. From this, a transition matrix is derived, containing the probabilities of transitions (\(P_{T}\)) between levels [Figure~\ref{fig_transmatrix}]. Simultaneously, the potential source amplitudes (\(\Delta\)) are calculated as the differences between the \(\mu\) values of distinct levels. This results in a two-dimensional \(P_{T}-\Delta\) space where unique groupings emerge, as illustrated in Supplementary Figure~1. These groupings reflect the distinct amplitude and activity characteristics of hidden sources. To extract an $N$-source set, the AP clustering algorithm is employed on a normalized \(P_{T}-\Delta\) space to ensure equal-sized average error bars in both dimensions~[Figure~\ref{fig_clustering}]. AP operates on the principle of "message-passing" between data points to identify distinct clusters~\cite{apc}. It then automatically selects a data point from each cluster to act as the "representative" of that specific cluster. Unlike traditional clustering algorithms that require the number of clusters to be specified in advance, AP autonomously discovers both the clusters and their representatives. The results of the AP algorithm applied to the \(P_{T}-\Delta\) space are shown in Figure~\ref{fig_clustering}, where a total of 13 clusters are identified. Adjusting the AP hyperparameter settings led to slight variations in the number of clusters and representatives, as detailed in Supplementary Figure~1. Based on all these variants, an ensemble of one-dimensional representative potential amplitudes (\(\Delta_{R}\)) is prepared, without utilizing the representative \(P_{T}\) information. To generate an $N$-source set, we first need to define the value of $N$. Given that the Bayesian levels are 11, the minimum number of hidden sources in the signal is 4 (\(2^{4} = 16\) state configurations). Multiple 4-source sets, referred to as candidate solutions, are then generated from the \(\Delta_{R}\) ensemble, making the combinatorial number feasible ($\sim$700 in this case) for computational workload. 

The next step is to determine the optimal candidate solution. This process begins by identifying the optimal baseline value for each candidate solution, ensuring that the total amplitudes of its state configurations align with the Bayesian levels. Figure~\ref{fig_baseline} illustrates the flowchart for this process. Initially, a 16-configuration set is generated from a candidate solution. A range of baseline values is then added to this set, producing a group of configuration sets. The range of baseline values is chosen such that the resulting sets comprehensively cover the distribution of the 11 Bayesian levels. For each configuration set, a log-likelihood value is computed by projecting its amplitudes onto the total distribution of Bayesian levels, as demonstrated in Figure~\ref{fig_basemlle} for a random configuration set. The configuration set with the highest log-likelihood value determines the optimal baseline.

%\captionsetup[subfigure]{font={normal}, skip=0pt, margin=0cm, singlelinecheck=false}
\begin{figure*}%[htbp!]
	\begin{minipage}{\textwidth}
		\subfigure[]{\begin{minipage}{0.55\textwidth}\includegraphics[width=\textwidth]{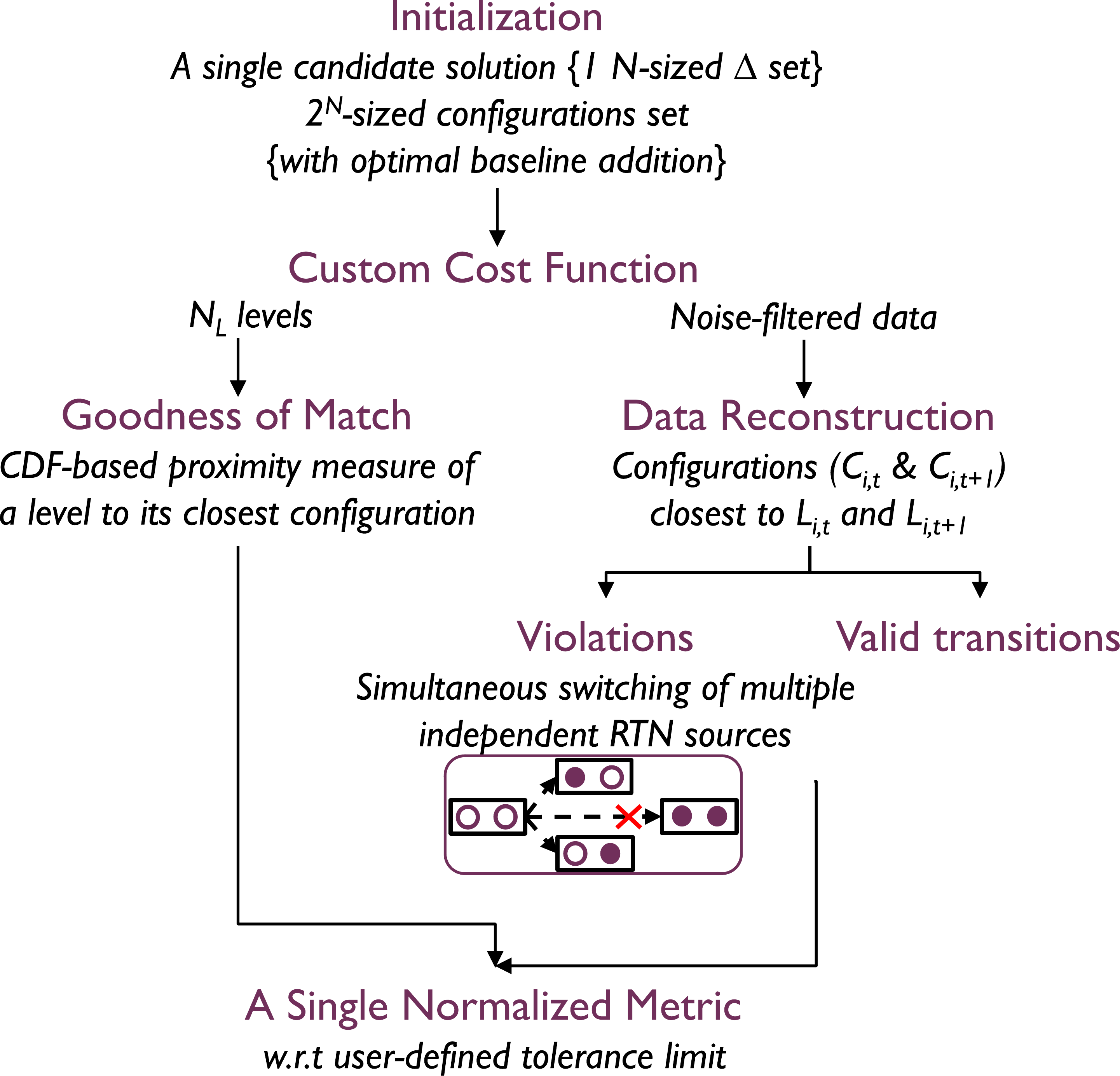}\end{minipage}\label{fig_costfn}}
		\subfigure[]{\begin{minipage}{0.38\textwidth}\includegraphics[width=\textwidth]{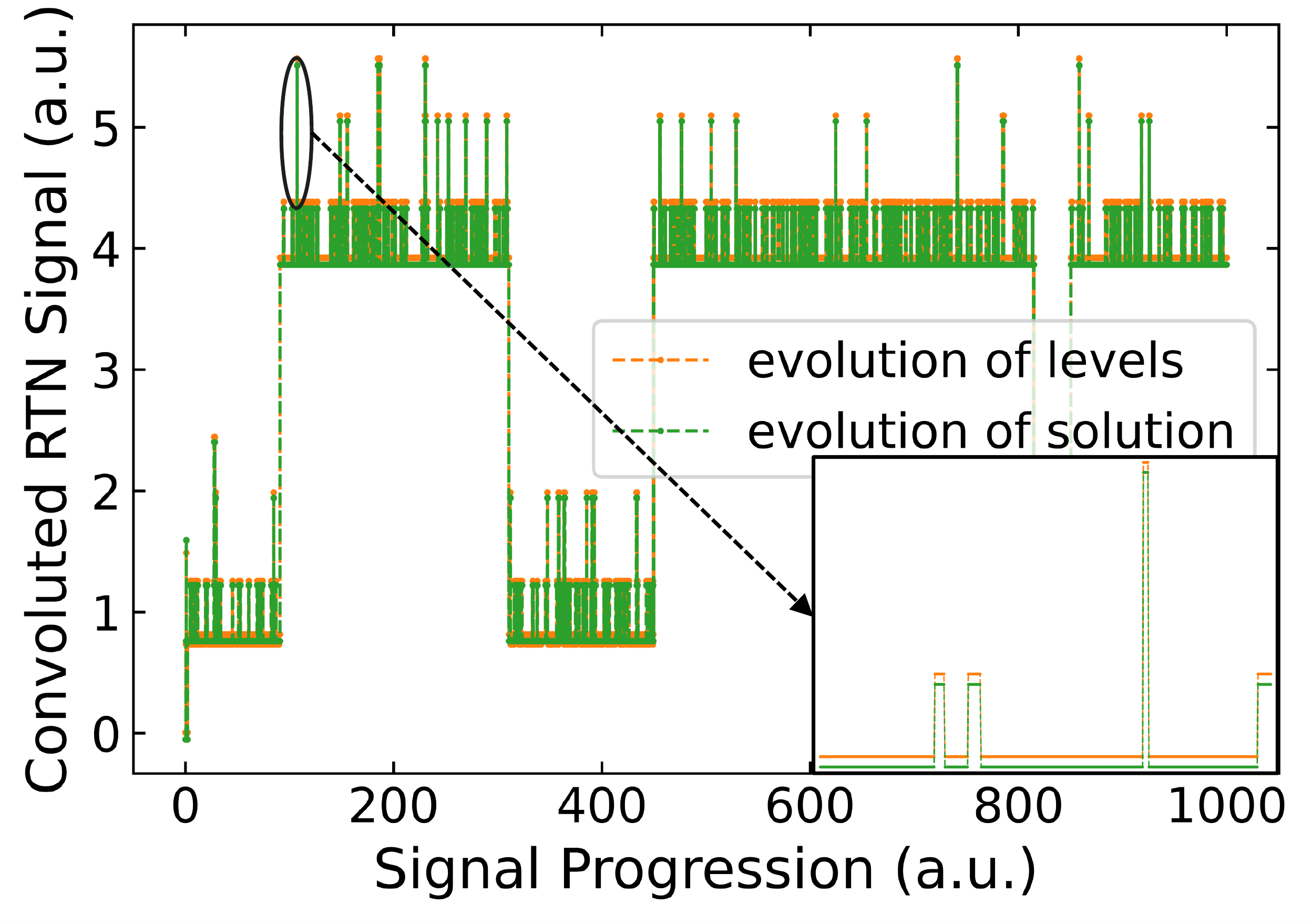}\\[-0.5em]\includegraphics[width=\textwidth]{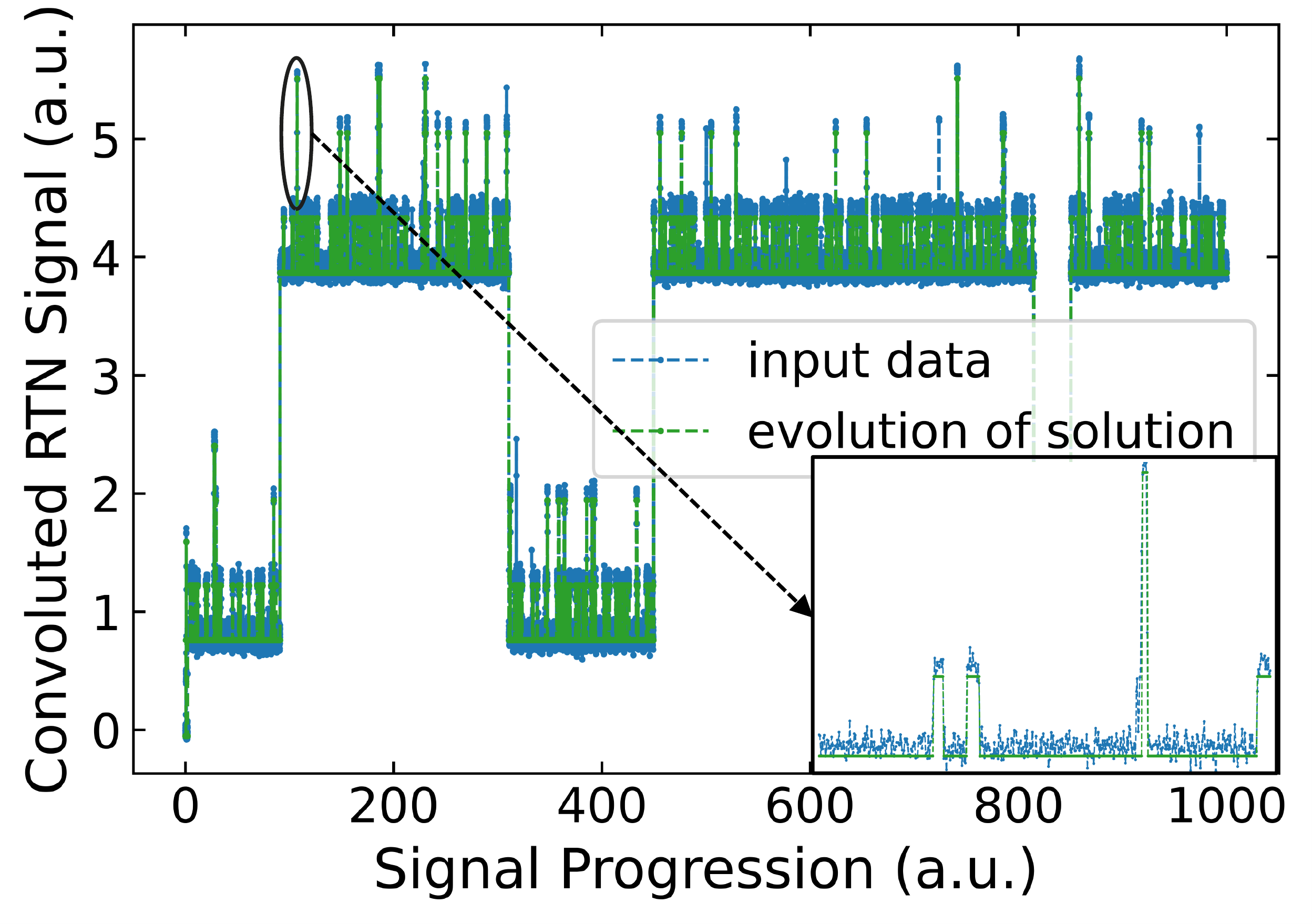}\end{minipage}\label{fig_sol}}
	\end{minipage}
	%\vspace*{0.2em}
	\subfigure[]{\includegraphics[width=0.7\textwidth]{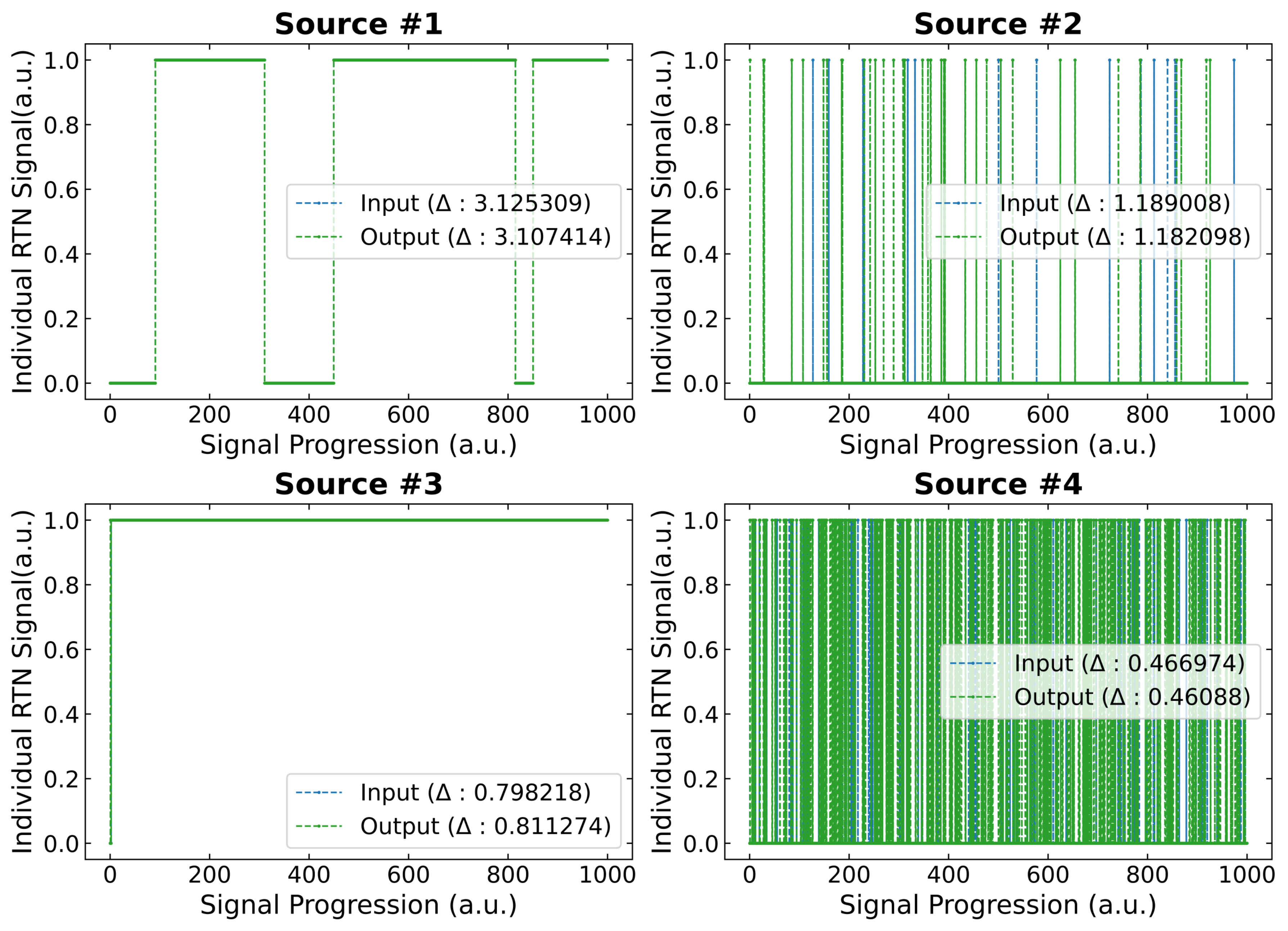}\label{fig_solact}}
	\vspace*{-0.5em}
	\caption{\justifying (a)~Cost function designed to evaluate candidate solutions, incorporating both static and dynamic metrics. The static metric maps each Bayesian level to its closest state configuration, quantifying the goodness of match based on proximity. The dynamic metric involves data reconstruction using these mapped state configurations, where simultaneous switching of multiple sources during a transition is treated as a violation. This process is exemplified for a 2-source system with four possible configurations, illustrating that transitions involving both sources changing their activity state are counted as violations. (b)~Data reconstruction results for the optimal candidate solution, demonstrating successful signal reconstruction with only 5 violations out of a total of 1086 transitions. (c)~Deconvolution of the optimal candidate solution into its four hidden sources, compared with the ground truth. The slow-activity sources (\#1 and \#3) are accurately extracted, with amplitude deviations within \(\pm 0.5\%\). For the fast-activity sources (\#2 and \#4), the extracted amplitudes are within \(\pm 1.7\%\), although some activity signatures are either not captured or shifted in occurrence instances. These discrepancies may arise from the de-noiser, where continuity constraints affect the quantized data, and/or from cost function constraints related to violated transitions.}\label{fig_solution}
\end{figure*}

Following this approach, each candidate solution is paired with a 16-configuration set that incorporates its corresponding optimal baseline value. To evaluate the effectiveness of a candidate solution, a custom cost function is developed, as illustrated in Figure~\ref{fig_costfn}. This cost function incorporates two metrics:\\ (i)~The \textit{goodness of match} metric assesses how well the 11 Bayesian levels align with their closest state configurations. A cumulative distribution function (CDF)-based proximity measure is used to quantify this match, assigning a value of 1 for an exact match and approaching 0 as the discrepancy increases.\\
(ii)~The \textit{reconstruction accuracy with transition constraints} metric assesses the validity of reconstructed quantized data under a specific rule: transitions between configurations must involve only the switching of a single source between on/off states. Simultaneous switching of multiple sources is considered a violation. This constraint is based on the fact that RTN sources can exhibit average activity rates ranging from the sampling rate to the full dataset length, for instance, from milliseconds to hours. In addition, since the RTN sources are considered independent, it is highly improbable for multiple sources to switch at the same instance. This approach helps maintain the fidelity of the reconstructed signal while accounting for realistic transition dynamics. Violations are counted during the reconstruction process to quantify adherence to the constraint.

The two metrics are combined into a single normalized metric relative to a predefined tolerance limit, which is a hyperparameter. In this scenario, the tolerance limits are set such that up to 2\% of total transitions are allowed as violations, and a mismatch tolerance of one standard deviation is permitted. Both metrics are equally weighted, resulting in a maximum cost function value of 2. An ideal candidate solution would perfectly match all Bayesian levels with no violations, yielding a cost function value of 0. Using this cost function, all candidate solutions are evaluated, and the optimal solution is identified as the one with the minimum cost function value. In this case, the optimal solution has a cost function of 1.249. Figure~\ref{fig_sol} illustrates the reconstructed RTN signal using this optimal candidate solution, which can be directly deconvolved into the activities of the four hidden sources [Figure~\ref{fig_solact}]. Since the dataset used in this example is synthetic and labeled, the \textit{RTNinja} framework's result is compared with the ground truth, demonstrating successful analysis of the convoluted RTN signal. 

If no candidate solution achieves a cost function value below 2, the \textit{SourcesMapper} module is automatically reinitialized under the hypothesis that the signal comprises one more source. This iterative process continues until the cost function condition is satisfied, ensuring an accurate representation of the RTN signal.

\section{Performance Evaluation}
To evaluate the robustness and limitations of the \textit{RTNinja} framework, we conducted extensive testing on a range of MC-simulated datasets. The framework was assessed based on two key criteria: data complexity — quantified by the number of sources in the signal — and the signal-to-noise ratio (SNR). To ensure a systematic and unbiased evaluation, the number of sources in each dataset was sampled from a uniform distribution rather than a physically motivated Poisson distribution. Source amplitudes were drawn from a unit exponential distribution, while noise was modeled using a Gaussian distribution with standard deviations set as percentages at 1\%, 5\%, 10\%, 20\%, and 30\%, respectively. In total, 1400 base datasets were generated, with 200 datasets corresponding to each source count from 1 to 7. For each of these base datasets, five noise levels were applied to simulate varying SNR conditions, resulting in a total of 7000 datasets. 

\begin{figure*} [htbp!]
	%\centering
	\vspace{-2em}
	\makebox[\textwidth][c]{%
	\begin{minipage}{0.33\textwidth}
		\centering Noise Level: 1\%
	\end{minipage}%
	\begin{minipage}{0.33\textwidth}
		\centering 5\%
	\end{minipage}%
	\begin{minipage}{0.33\textwidth}
		\centering 30\%
	\end{minipage}}
	\vspace{-1em}
	\subfigure[]{\includegraphics[width=0.32\textwidth]{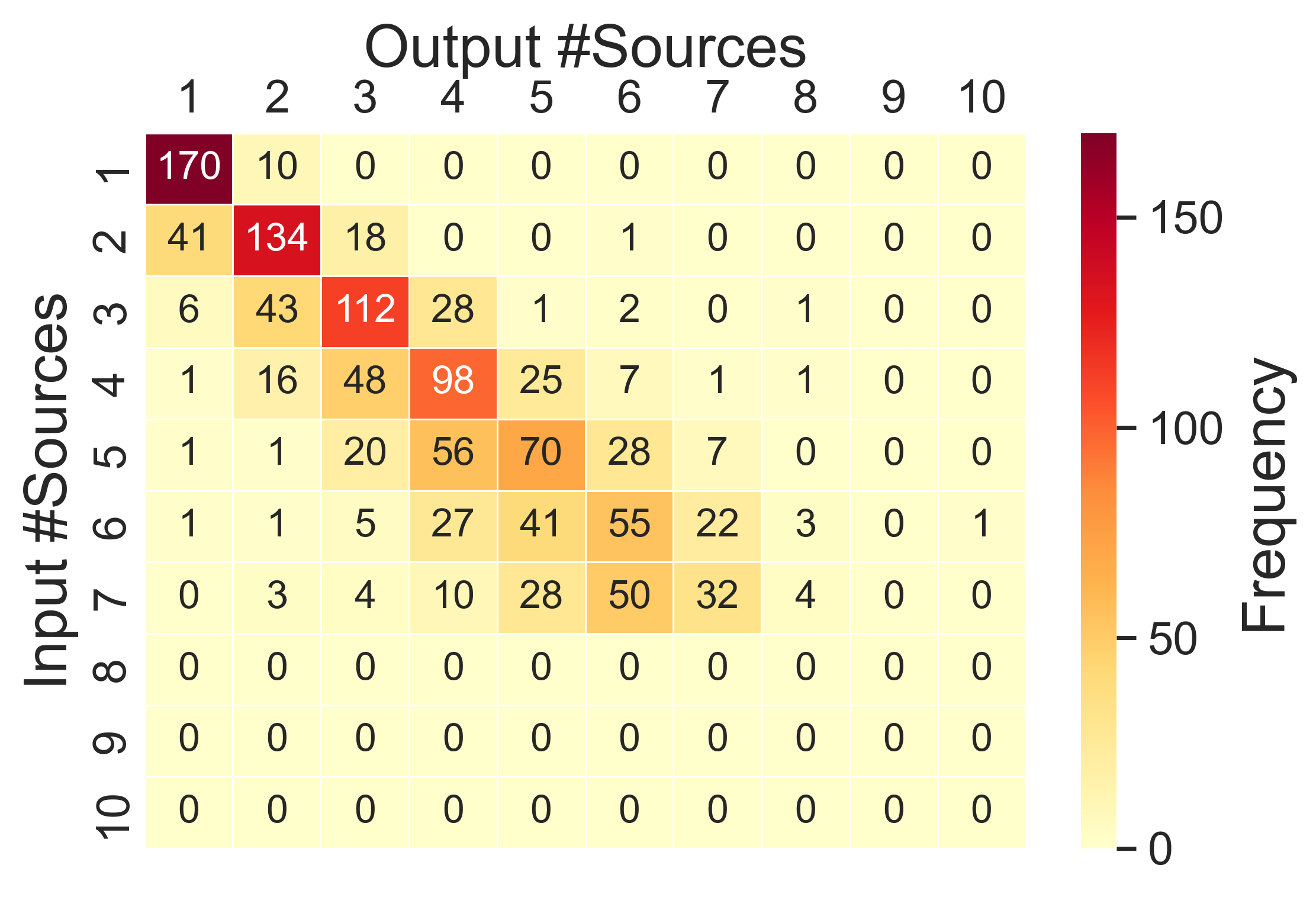}\includegraphics[width=0.32\textwidth]{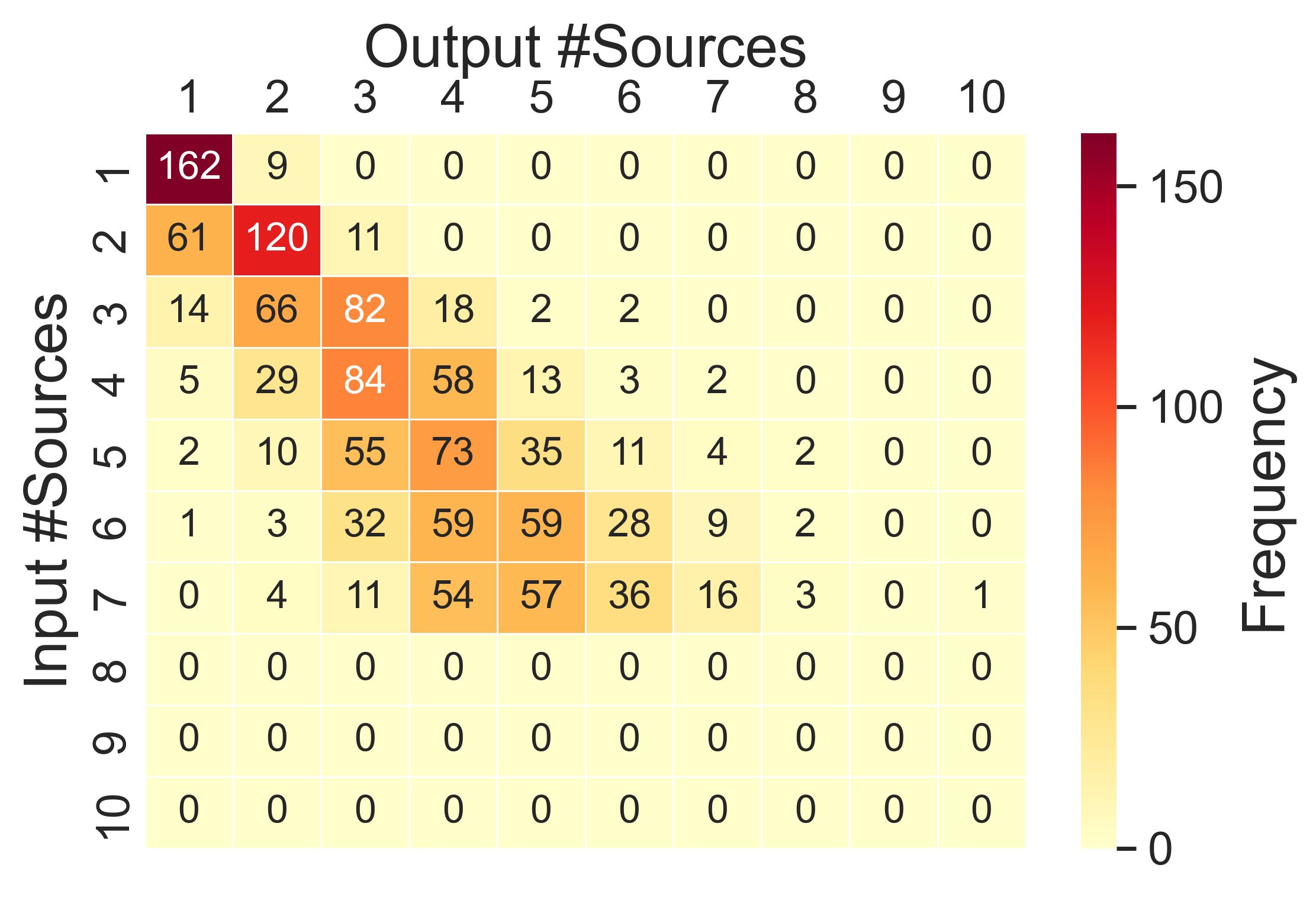}\includegraphics[width=0.32\textwidth]{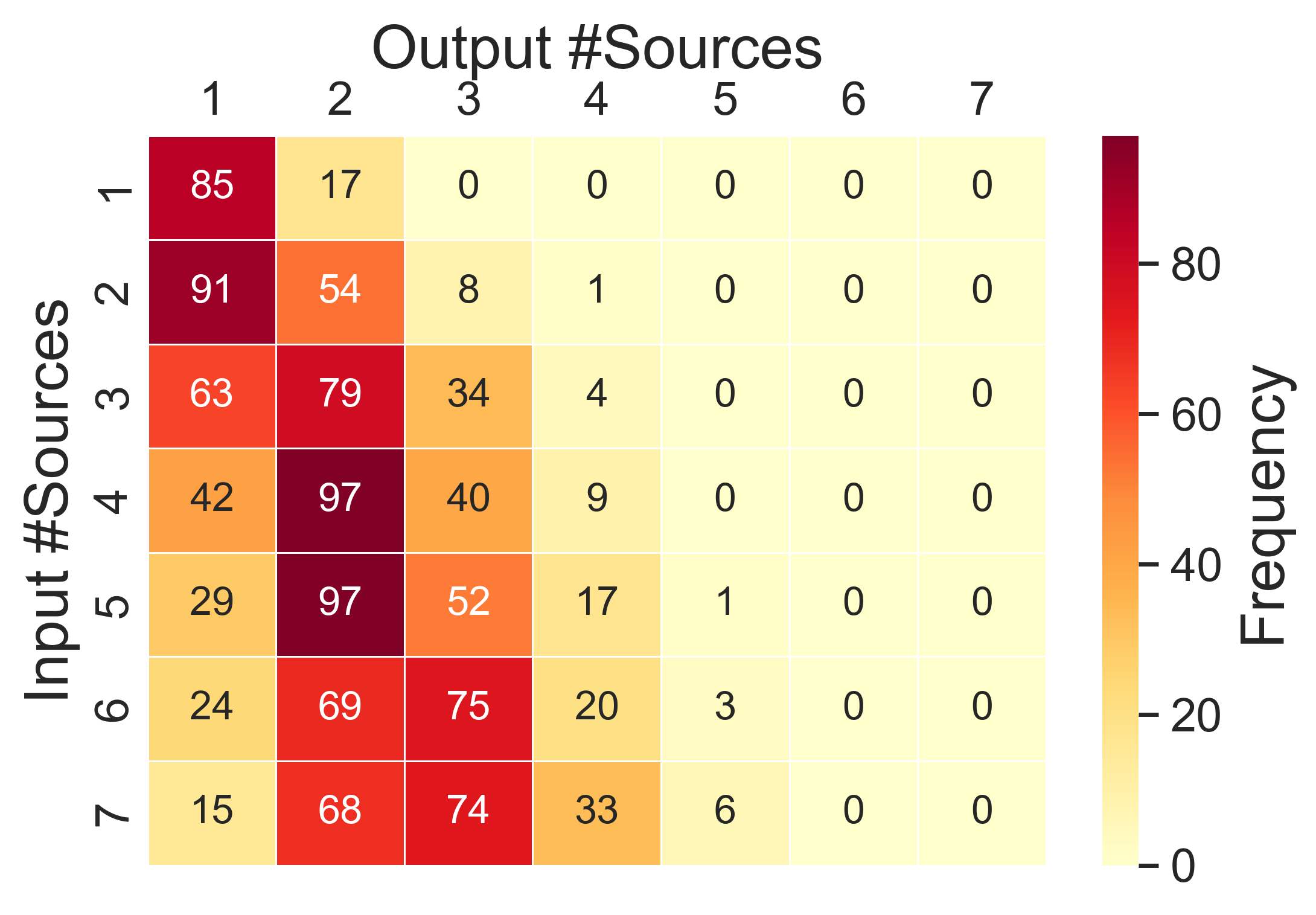}\label{fig_confmatrix}}
	\vspace{-1em}
	\subfigure[]{\includegraphics[width=0.32\textwidth]{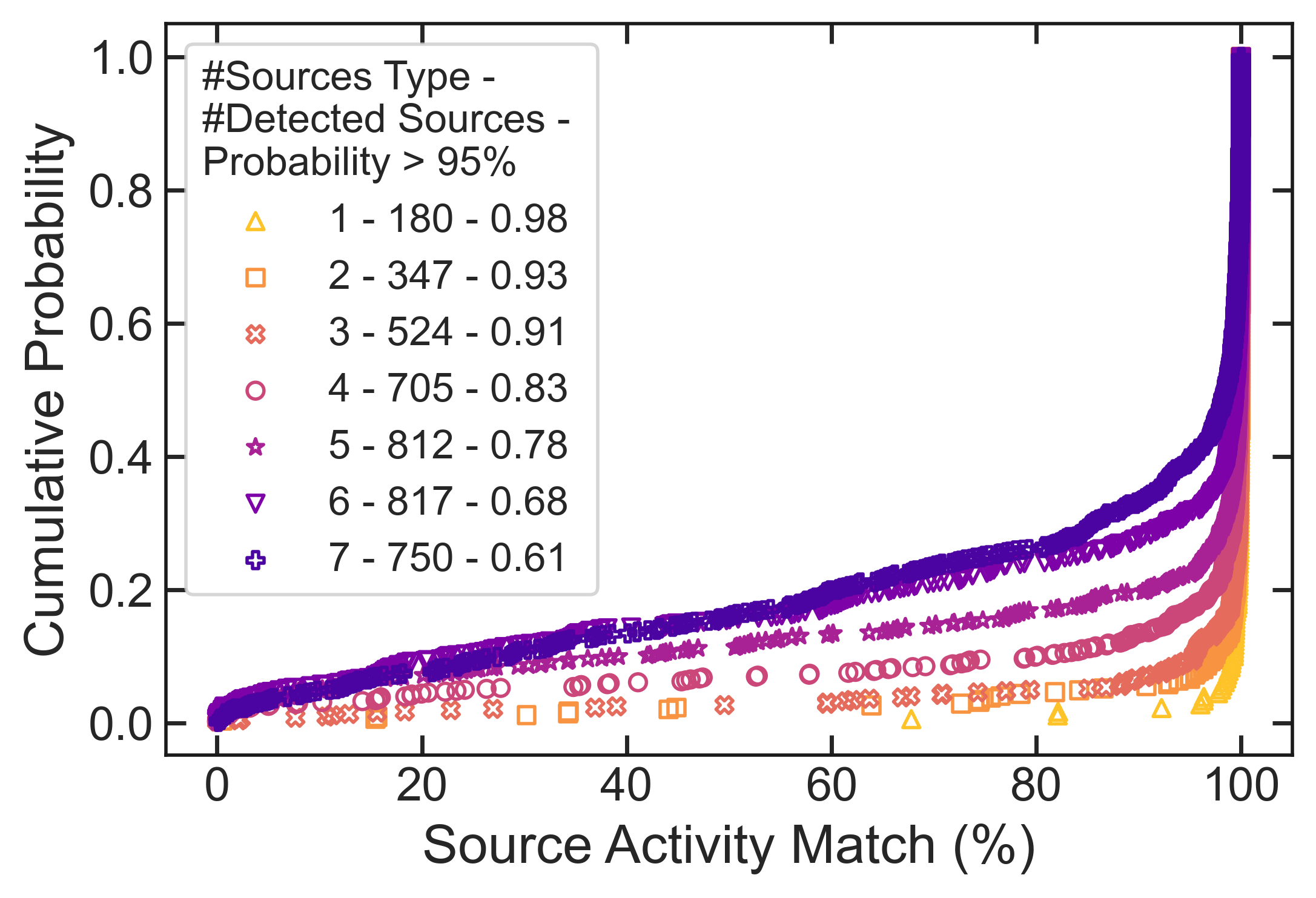}\includegraphics[width=0.32\textwidth]{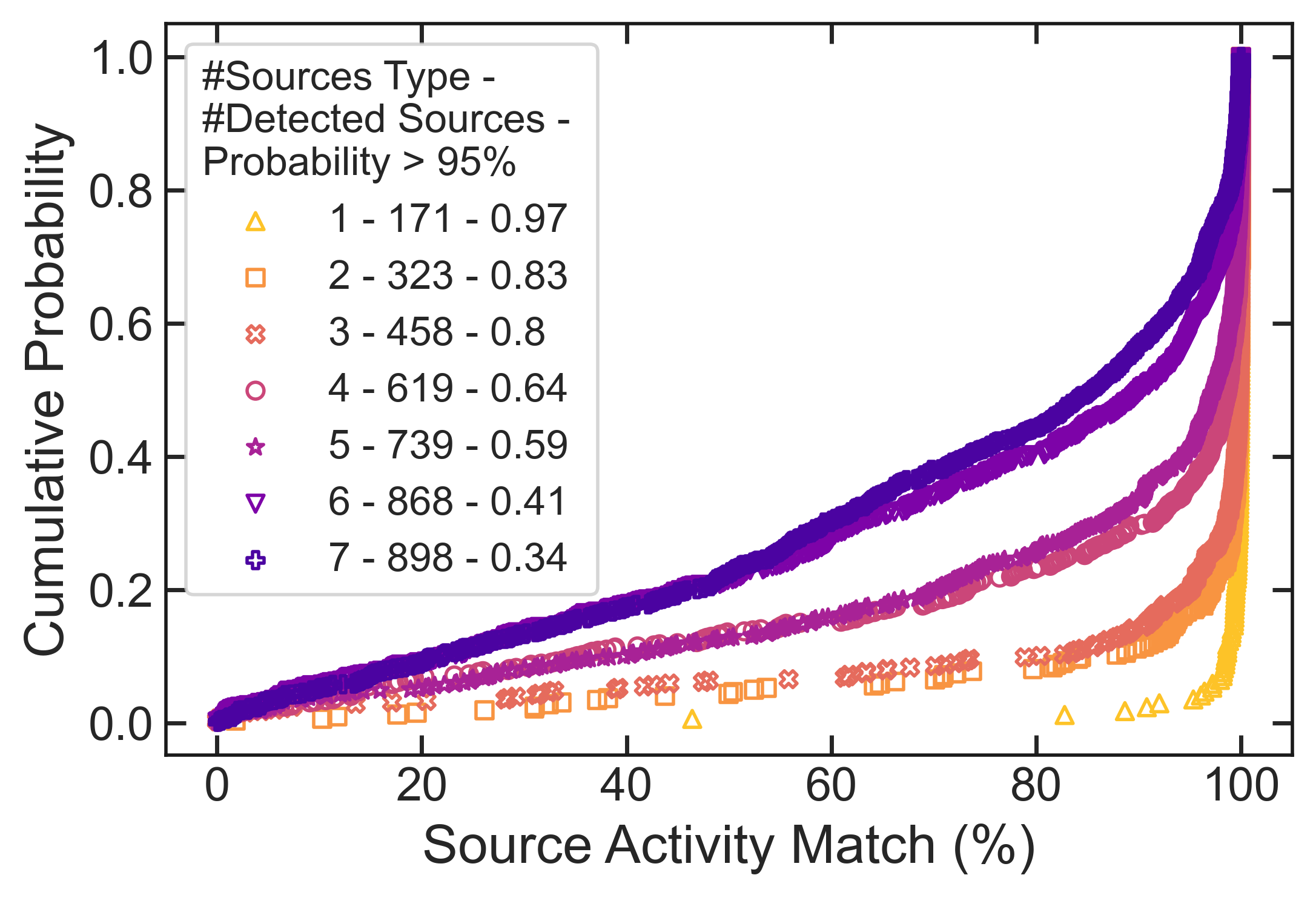}\includegraphics[width=0.32\textwidth]{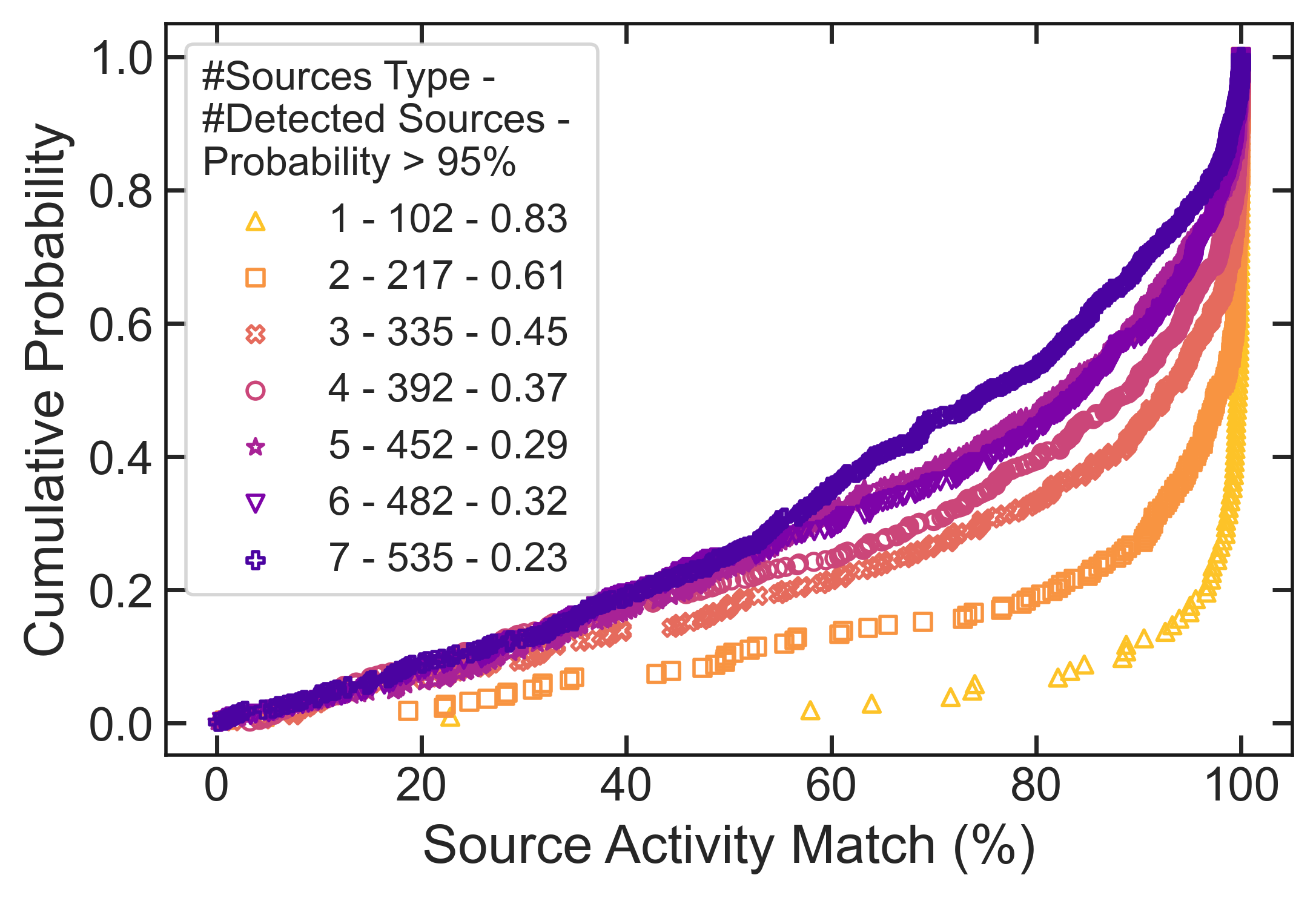}\label{fig_gof}}
	\vspace{-1em}
	\subfigure[]{\includegraphics[width=0.32\textwidth]{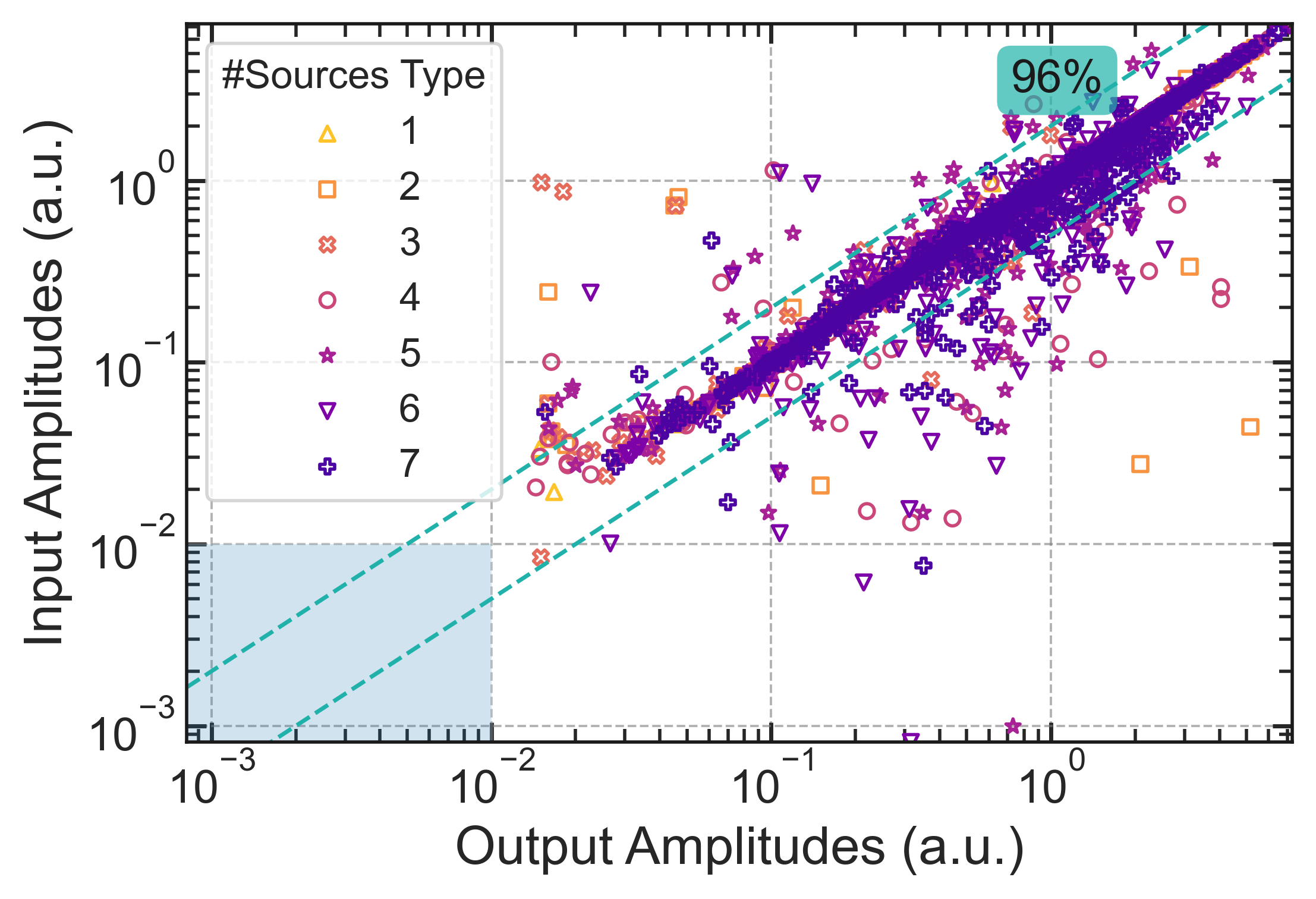}\includegraphics[width=0.32\textwidth]{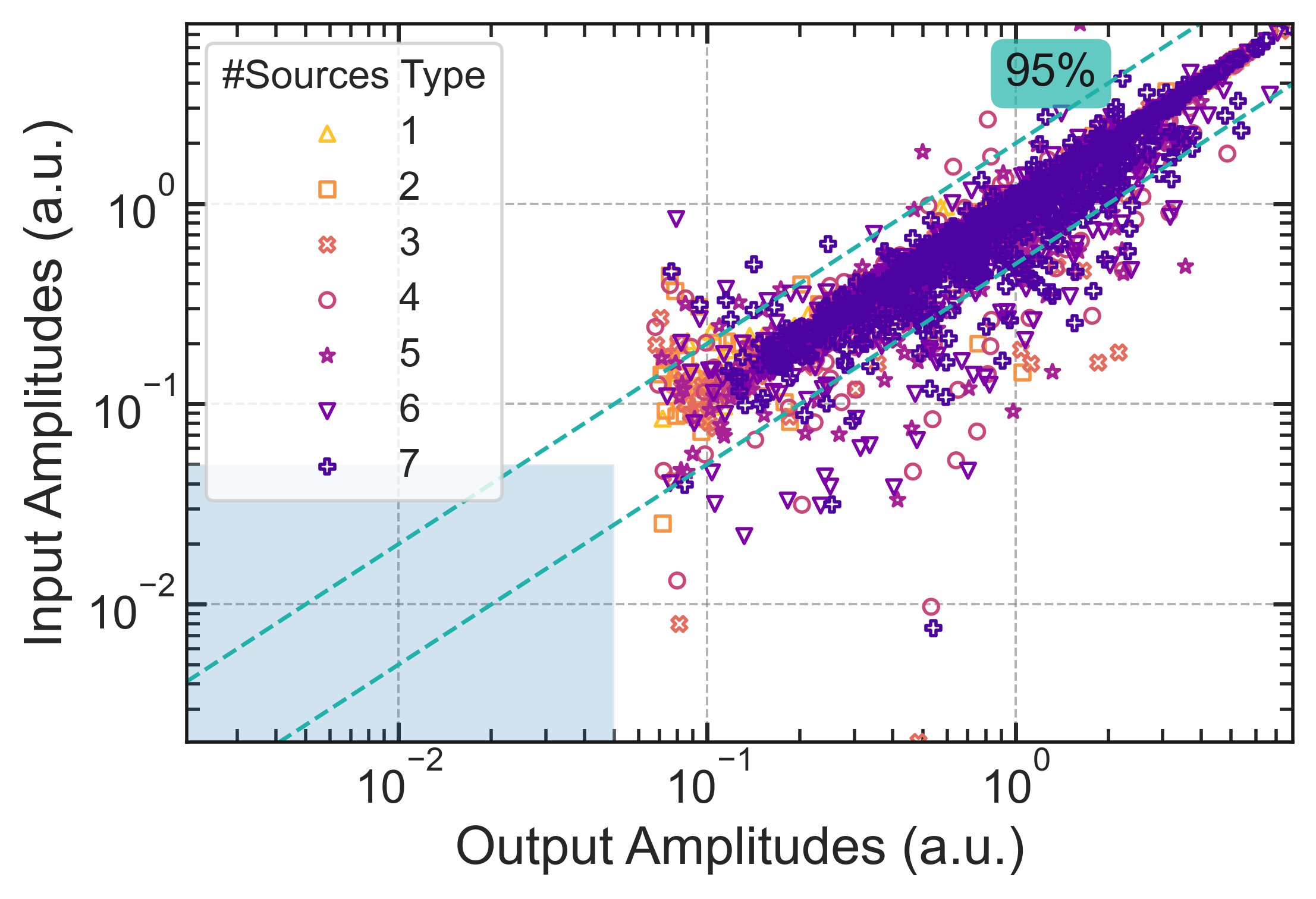}\includegraphics[width=0.32\textwidth]{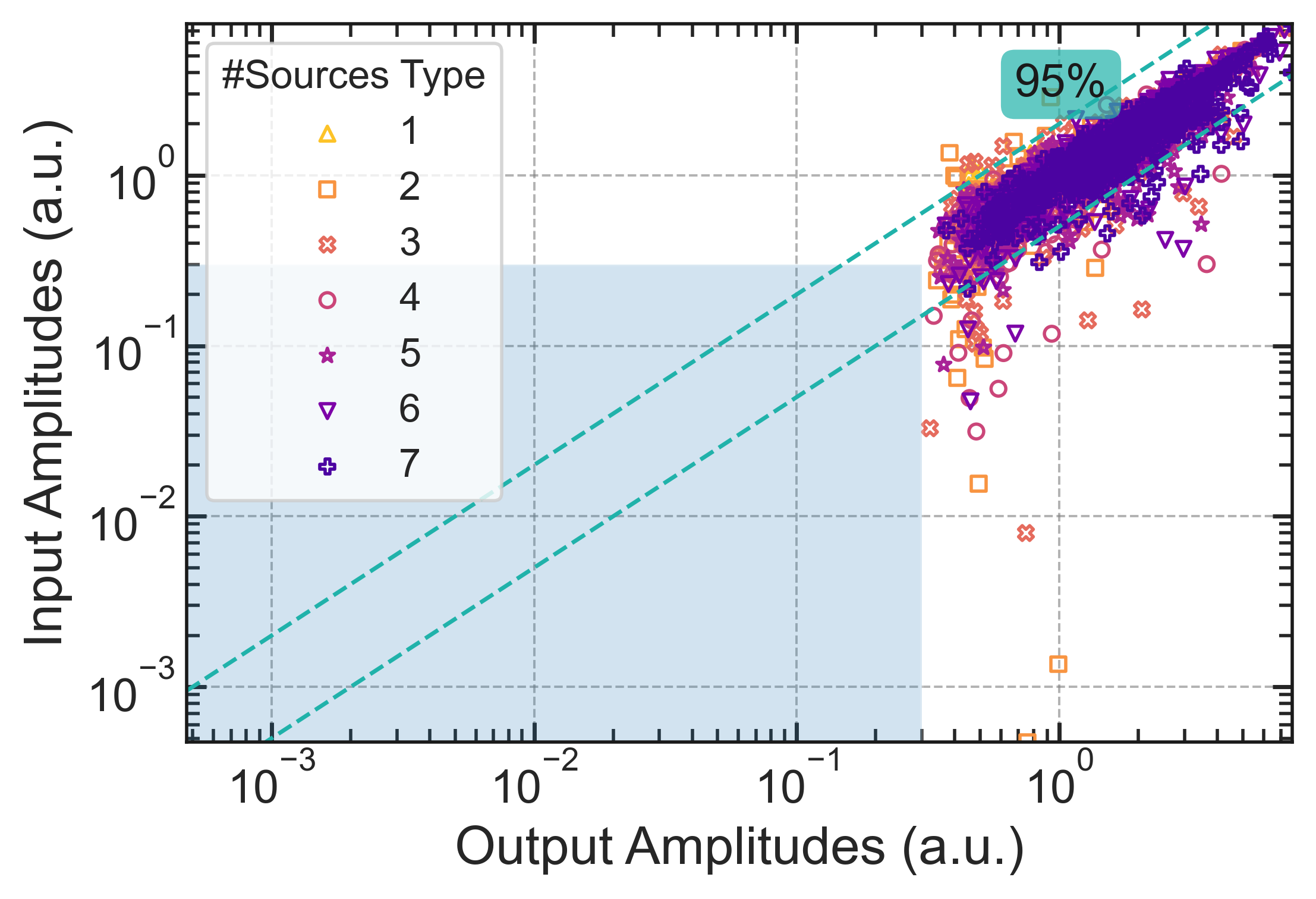}\label{fig_ampd}}
	\vspace{-1em}
	\subfigure[]{\includegraphics[width=0.32\textwidth]{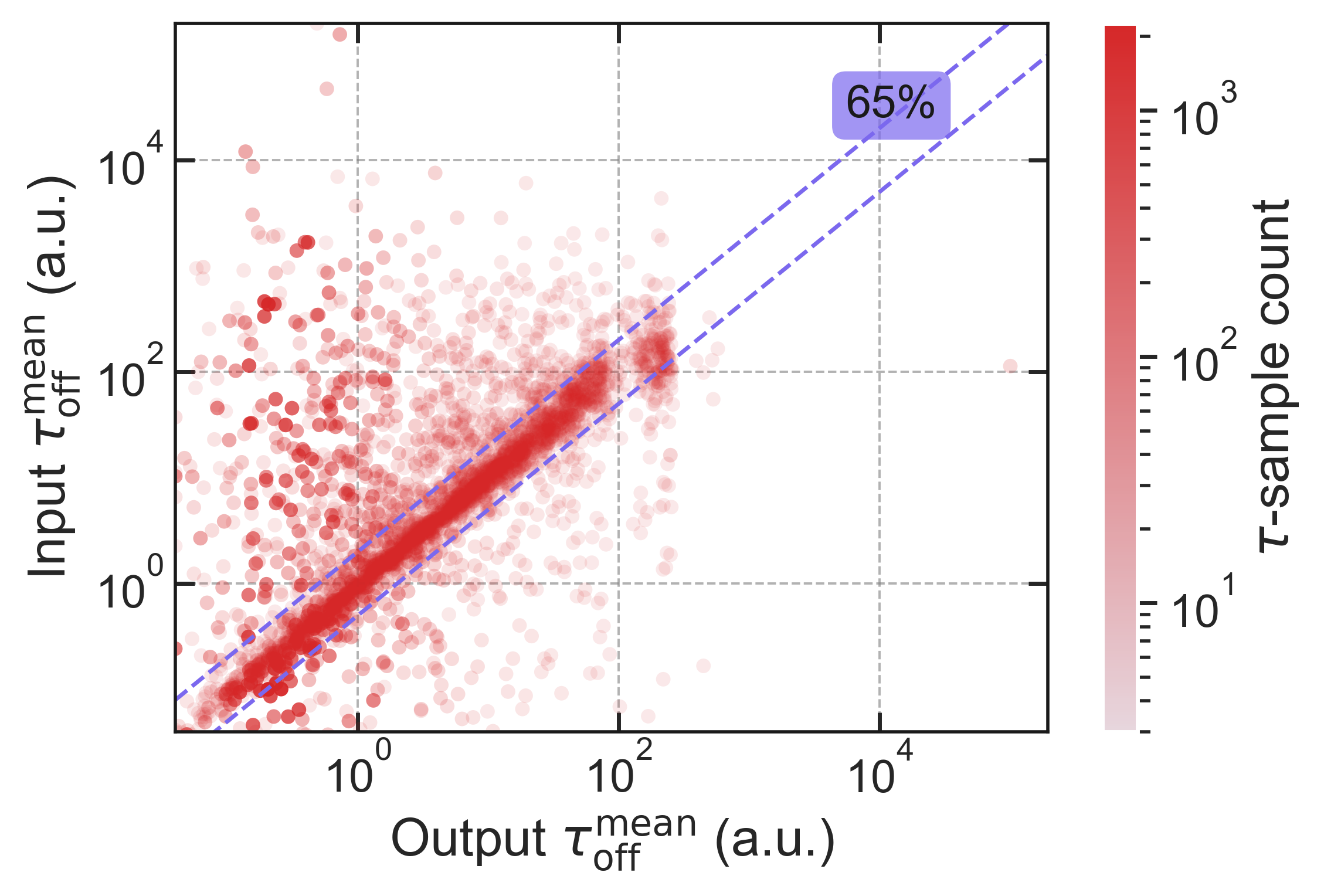}\includegraphics[width=0.32\textwidth]{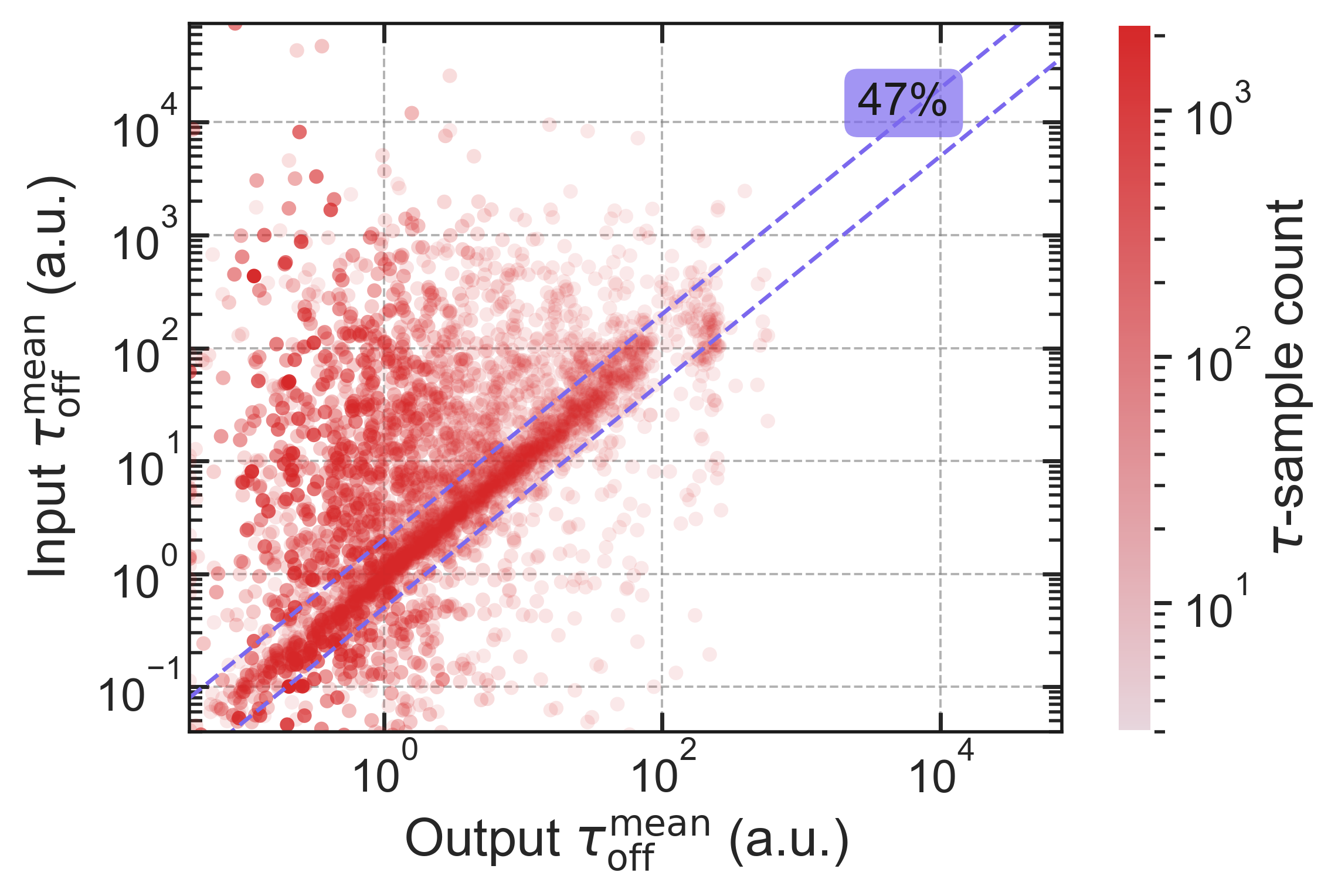}\includegraphics[width=0.32\textwidth]{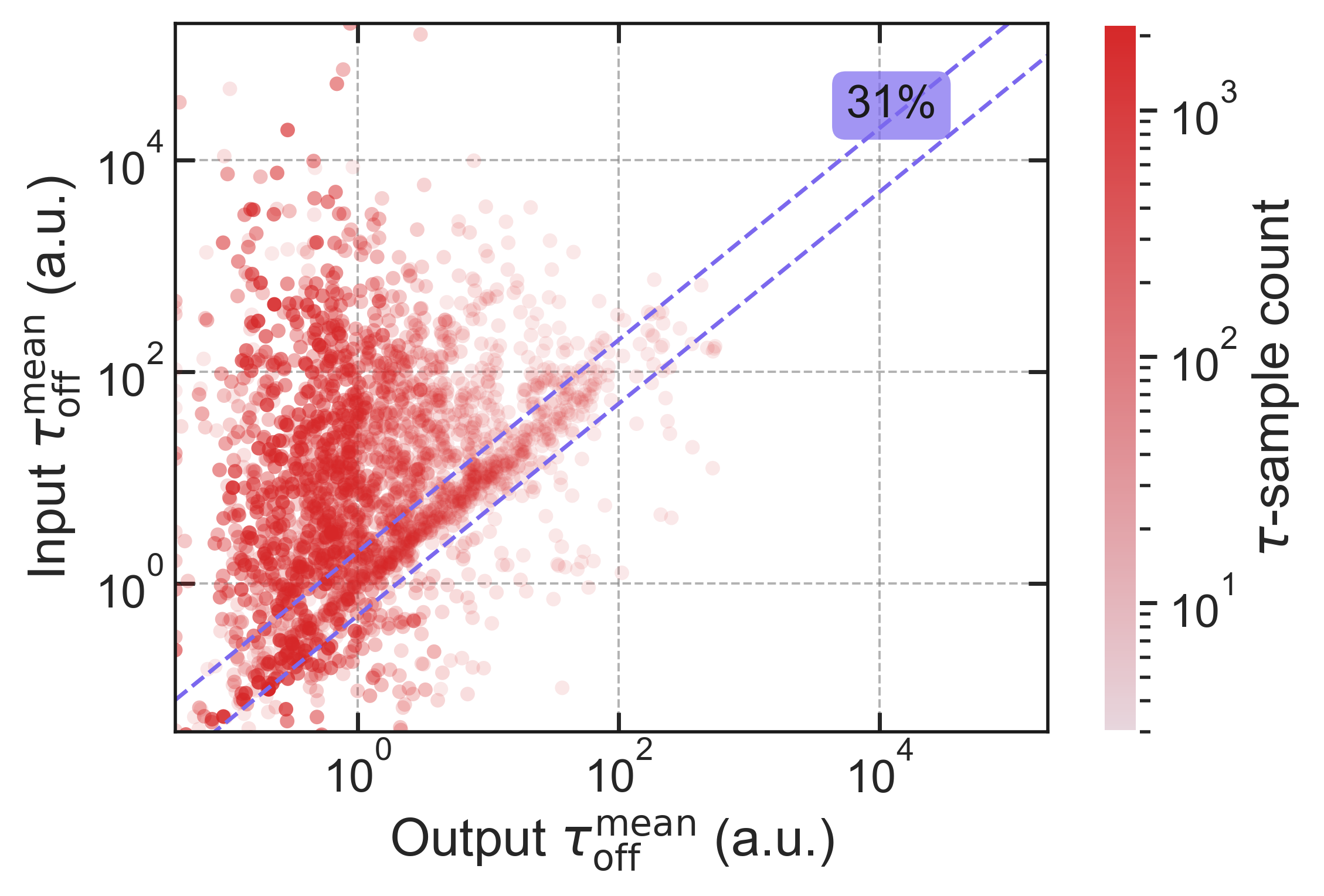}\label{fig_toff}}
	\vspace{-1em}	
	\subfigure[]{\includegraphics[width=0.32\textwidth]{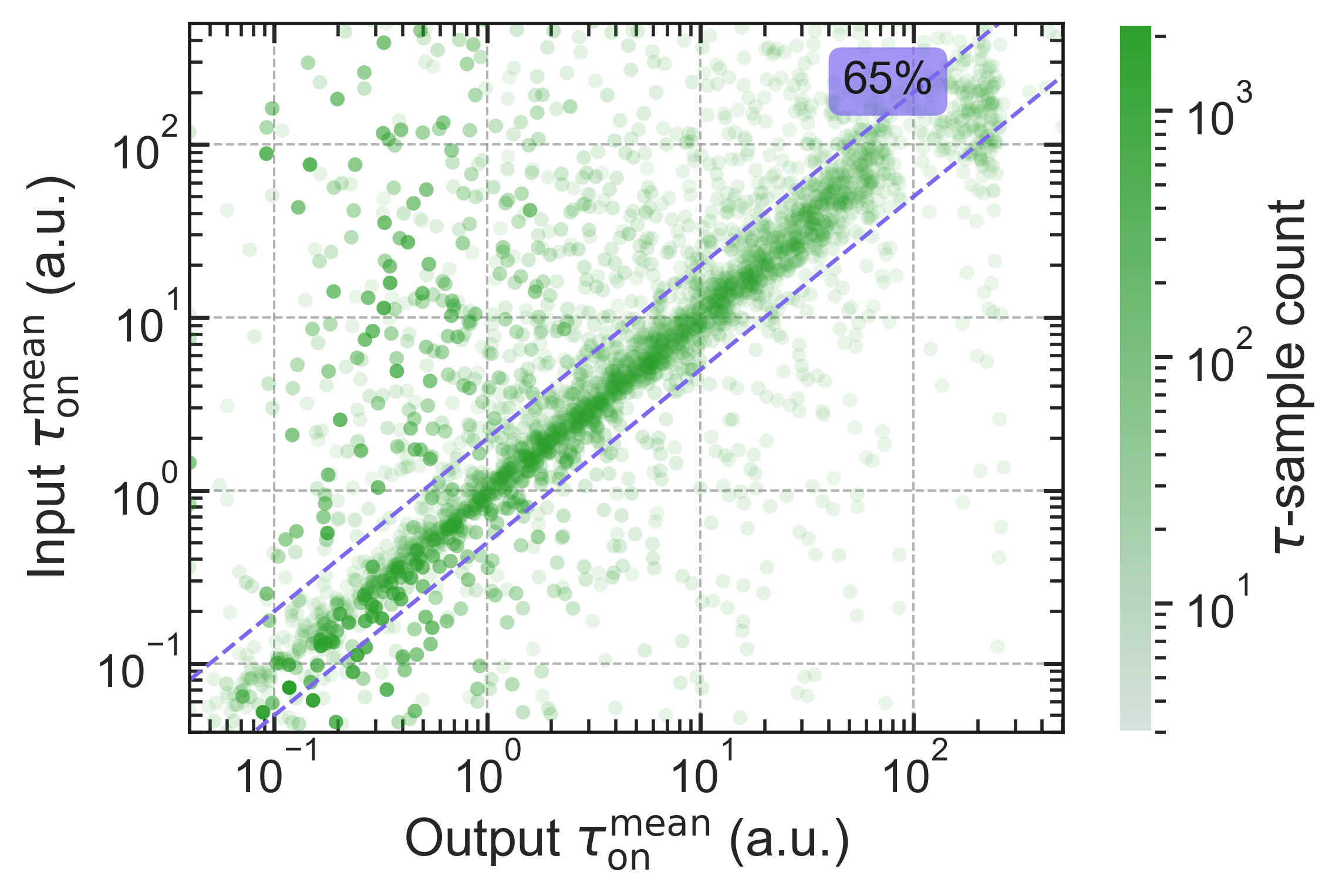}\includegraphics[width=0.32\textwidth]{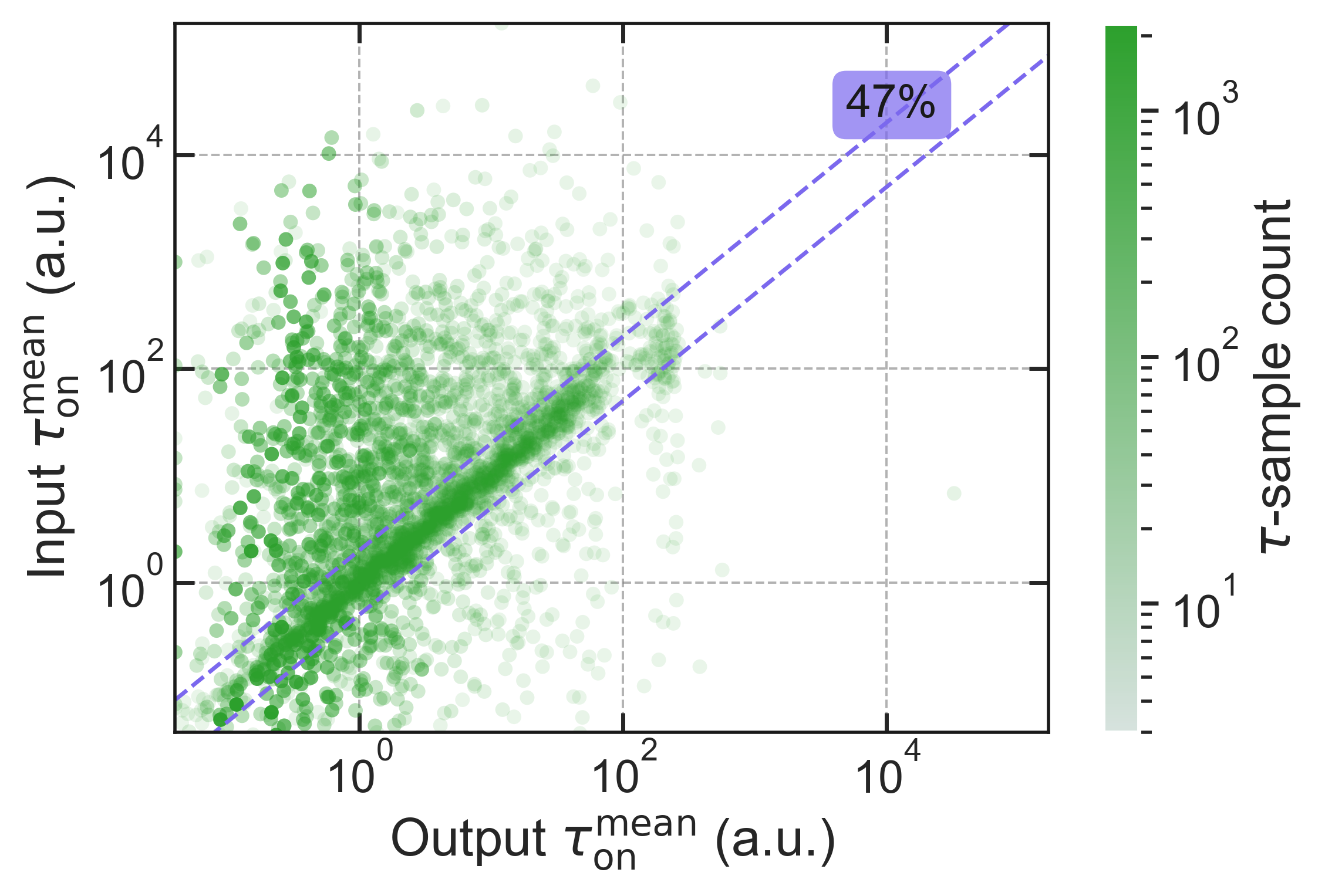}\includegraphics[width=0.32\textwidth]{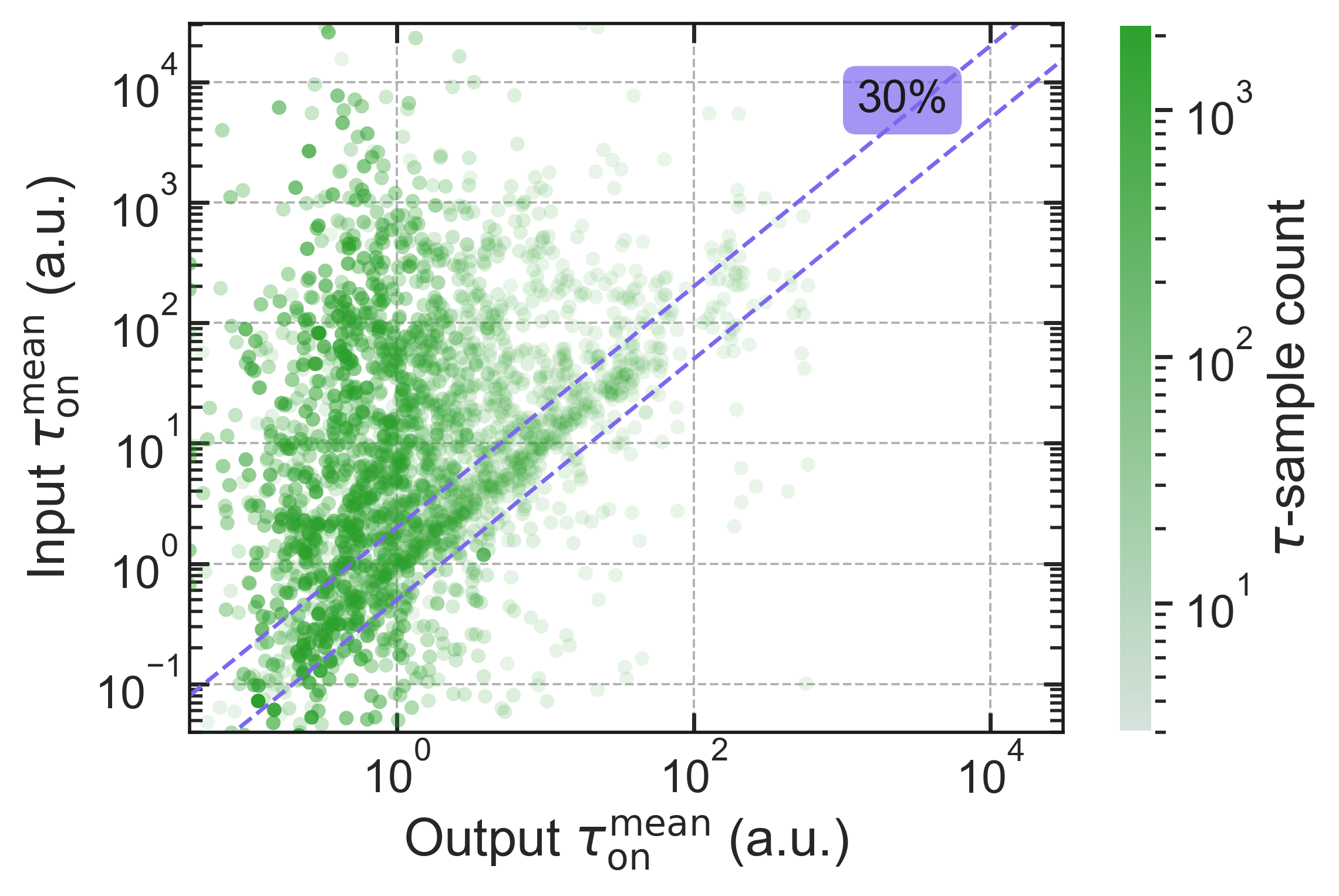}\label{fig_ton}}
	\vspace{-0.5em}
	\caption{\justifying Performance of the \textit{RTNinja} framework at 1\%, 5\%, and 30\% noise levels across all dataset complexities. (a)~Matrix showing the number of datasets corresponding to each combination of true and estimated source counts, with diagonal entries indicating correct matches. (b)~Activity match accuracy for mapped input--output sources, grouped by dataset complexity. The number of detected sources and probability of achieving $>$95\% activity match is indicated. (c)~Accurate amplitude extraction is demonstrated, independent of noise level and data complexity. Blue squares indicate noise bounds; dashed lines mark $0.5\times$ and $2\times$ amplitude thresholds, with percentages showing sources within bounds. (d)~Off-state and (e)~on-state mean durations, with the percentage of estimates within $0.5\times$$-$$2\times$ of true values. Accuracy is affected by inherent statistical sampling limitations and increasing noise, which can lead to underestimation.}\label{fig_stats}
\end{figure*}

A comprehensive evaluation of the \textit{RTNinja} framework’s performance across varying data complexities is presented in Figure~\ref{fig_stats} for three noise levels: 1\%, 5\%, and 30\%. Additional results corresponding to intermediate noise levels of 10\% and 20\% are provided in Supplementary Figure~2. Figure~\ref{fig_confmatrix} illustrates the comparison between the true number of sources in each dataset and the number estimated by \textit{RTNinja}. The diagonal entries in this matrix indicate the number of datasets with correctly identified source counts. As noise levels increase, accurate detection becomes progressively more challenging because weaker sources — i.e., those with low amplitudes — are increasingly suppressed by noise, often resulting in an underestimation of the true number of sources. Ideally, each row in the matrix should sum to 200 datasets, representing the complete batch for each source count. However, in practice, \textit{RTNinja} occasionally fails to return a result for certain datasets (Supplementary Table~1). This is particularly true for datasets with fewer sources, where all sources become indistinguishable from noise due to low amplitudes. Conversely, in datasets with a higher number of sources and low noise levels, overfitting may occur due to the Bayesian algorithm overestimating the number of levels, which can lead to non-convergence in the \textit{SourcesMapper} module. This is often caused by the initialization of the $N$-source hypothesis at an inflated value and the resulting computational burden of evaluating a large combinatorial space of candidate solutions. Hence, more datasets yield results at the highest noise level (30\%) than at the lowest (1\%) because the framework avoids overfitting in noisier conditions. Overall statistics on output yield across noise levels are provided in Supplementary Table~1.

%\captionsetup[subfigure]{font={normal}, skip=0pt, margin=0cm, singlelinecheck=false}

To rigorously assess \textit{RTNinja}’s performance at the level of individual sources, we mapped each true (input) source to its corresponding estimated (output) source. This mapping is straightforward in cases where the number of estimated sources matches the true number of sources in the dataset. However, discrepancies arise when \textit{RTNinja} underestimates or overestimates the number of sources — typically due to low-amplitude sources being missed in high-noise scenarios or due to noise being mistakenly modeled as an additional source in low-noise, overfitted conditions. In such instances, we adopt an amplitude-based matching strategy: for example, if a dataset contains four true sources but \textit{RTNinja} decomposes the signal into five sources, we select the four output sources whose amplitudes most closely match those of the true sources and discard the remaining unmatched component. This approach is similarly applied in cases of underestimation, where fewer sources are identified than are actually present. While the full temporal activity of each source is available, we deliberately rely solely on source amplitudes for input--output mapping to avoid artifacts that can arise from activity merging or splitting — a common side effect when the estimated number of sources deviates from the ground truth. This amplitude-centric matching ensures a consistent and interpretable evaluation even in the presence of estimation errors.

Once the input--output source pairs have been successfully mapped, we assess \textit{RTNinja}’s ability to accurately extract key characteristics of each individual source. The primary characteristic evaluated is the full activity profile of a source, measured using the activity match accuracy. To quantify this, we employ the Hamming distance, which calculates the dissimilarity between the binary activity patterns of the input and output, specifically whether a source is in the off- or on-state at each instance~\cite{hamming}. The Hamming distance is then converted into an activity match percentage by computing its complement. Figure~\ref{fig_gof} illustrates the distribution of activity match percentages plotted against cumulative probability for each $N$-source dataset configuration across varying noise levels. Overall, \textit{RTNinja} exhibits strong performance in recovering source activity. However, the accuracy of activity matching systematically decreases with increasing data complexity, and also declines with higher noise levels for any fixed number of sources. These trends highlight the increasing challenge of accurately capturing source-level dynamics under more complex and noisier signal conditions.

Figure~\ref{fig_ampd} presents the comparison of input and output amplitudes for mapped sources across different $N$-source dataset types. To evaluate the accuracy of amplitude extraction, we compute the percentage of mapped sources whose estimated amplitudes fall within a factor of two (i.e., between $0.5\times$ and $2\times$) of the corresponding true amplitude. \textit{RTNinja} demonstrates strong performance in accurately recovering source amplitudes, independent of noise levels. However, the overall detection window narrows as noise increases, resulting in a systematic reduction in the proportion of correctly identified sources — from 73.8\% at 1\% noise to 44.9\% at 30\% noise, as detailed in Supplementary Table~1. This drop is primarily attributed to low-amplitude sources being increasingly suppressed by the rising noise floor. Notably, the accuracy of amplitude estimation does not exhibit a clear dependency on data complexity, suggesting that noise, rather than the number of sources, is the dominant factor affecting amplitude detectability.

Since the full binary activity trace of each source is available, we further analyze the durations of isolated on-state and off-state intervals. The temporal lengths of these continuous state segments approximate an exponential distribution, as discussed in Figure~\ref{fig_mcsim}. We fit these distributions using a maximum likelihood estimation method to extract the primary parameters: the mean off-state duration ($\langle\tau_{off}^{n}\rangle$) and mean on-state duration ($\langle\tau_{on}^{n}\rangle$). The comparisons of these parameters between the ground truth and \textit{RTNinja} outputs are shown in Figures~\ref{fig_toff} and~\ref{fig_ton}, respectively. \textit{RTNinja} successfully recovers these distribution parameters for a substantial proportion of sources, as assessed by the percentage of mapped sources for which the extracted mean lies within a factor of two of the true value. However, accuracy declines systematically with increasing noise, accompanied by a growing tendency to underestimate the mean durations. This underestimation arises when some sources are not detected, causing their activity to be erroneously merged into the trajectories of detected sources — resulting in artificially faster dynamics. In addition, the reliability of mean duration estimates is inherently limited by the number of observed transitions: for example, if a source enters the on-state only twice during the observation window, the estimated mean is constrained by statistical sampling limitations.

Having evaluated \textit{RTNinja}’s performance at the level of individual sources, we now step back to examine the framework’s aggregate performance across entire datasets, focusing specifically on how varying noise levels affect the results when data complexity is pooled. For each noise level, this involves analyzing the ensemble of all converged datasets (up to 1400 per condition, as detailed in Supplementary Table 1). Such an aggregated perspective is particularly relevant in the context of semiconductor device research, where the statistical characterization of RTN behavior is typically drawn from the overall distributions of defect parameters — most notably, amplitude and state duration metrics — and summarized using statistical descriptors like the mean and standard deviation~\cite{grasser2020noise}. These metrics are routinely employed to benchmark and compare device technologies in terms of their RTN impact.

%\captionsetup[subfigure]{font={normal}, skip=0pt, margin=0cm, singlelinecheck=false}
\begin{figure*} %[htbp!]
	%\centering
	\subfigure[]{\includegraphics[width=0.325\textwidth]{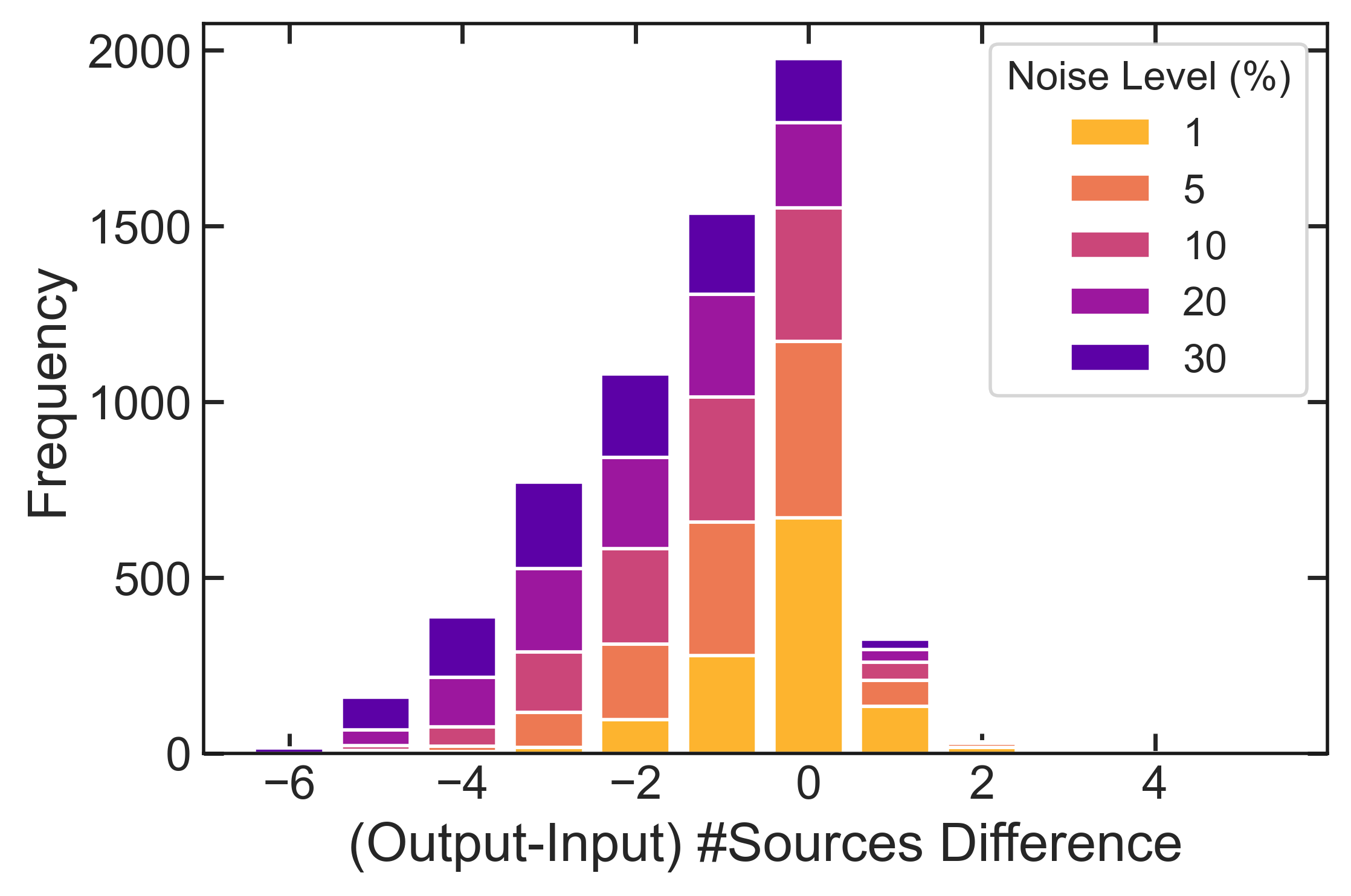}\label{fig_ndiff}}
	\subfigure[]{\includegraphics[width=0.32\textwidth]{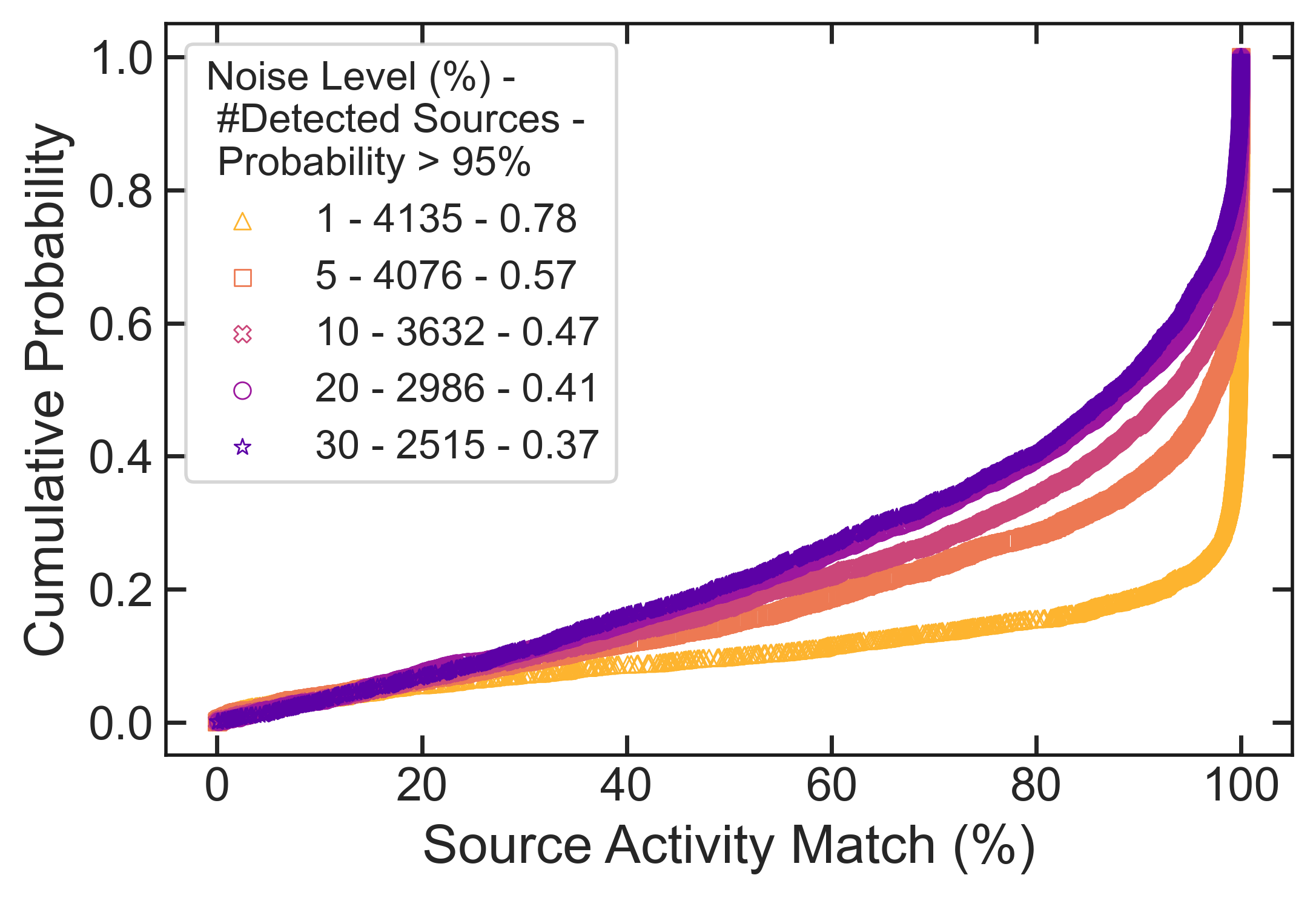}\label{fig_ngof}}
	\vfill%
	\vspace{-0.5em}
	\subfigure[]{\includegraphics[width=0.325\textwidth]{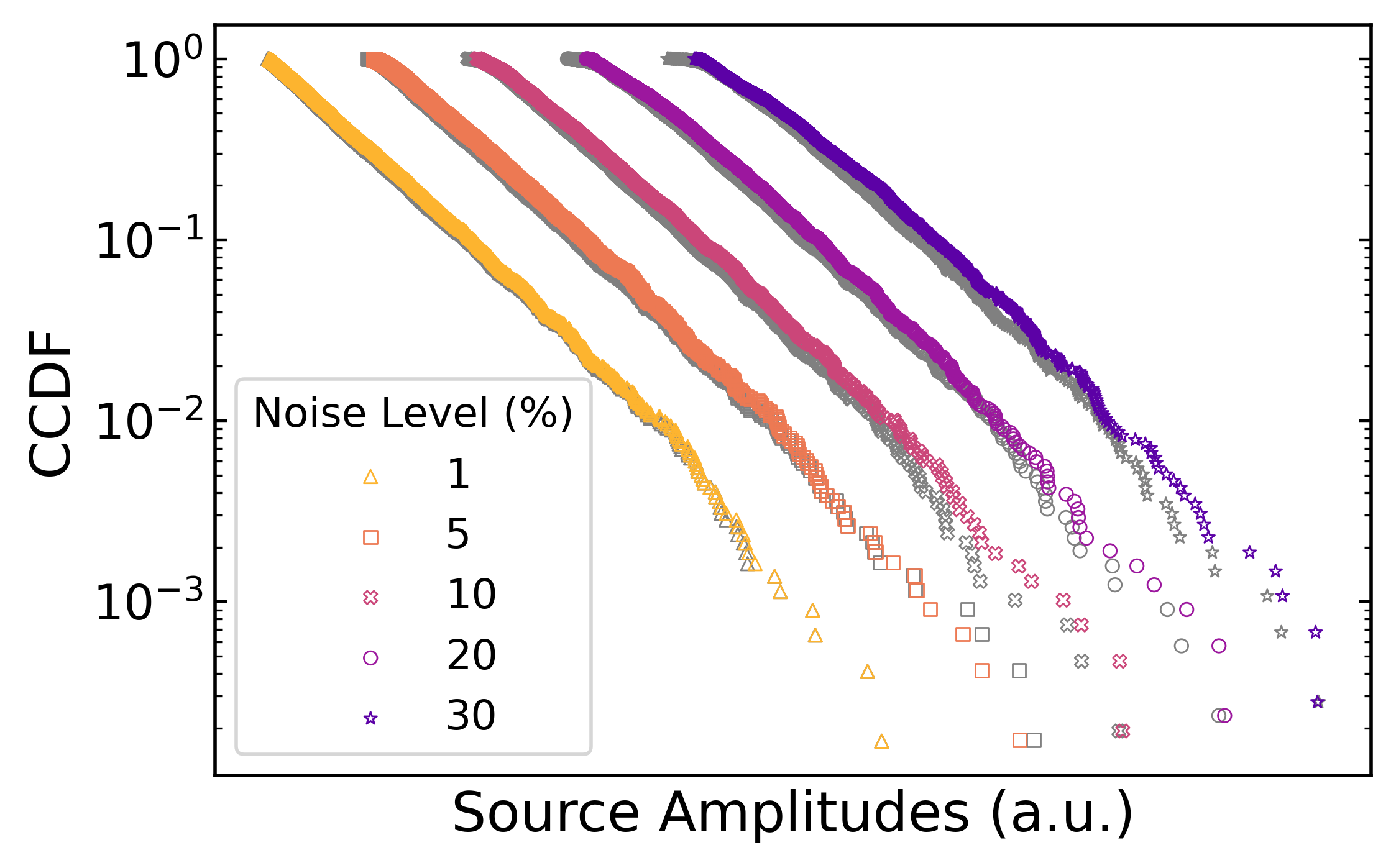}\label{fig_nampd}}
	\subfigure[]{\includegraphics[width=0.32\textwidth]{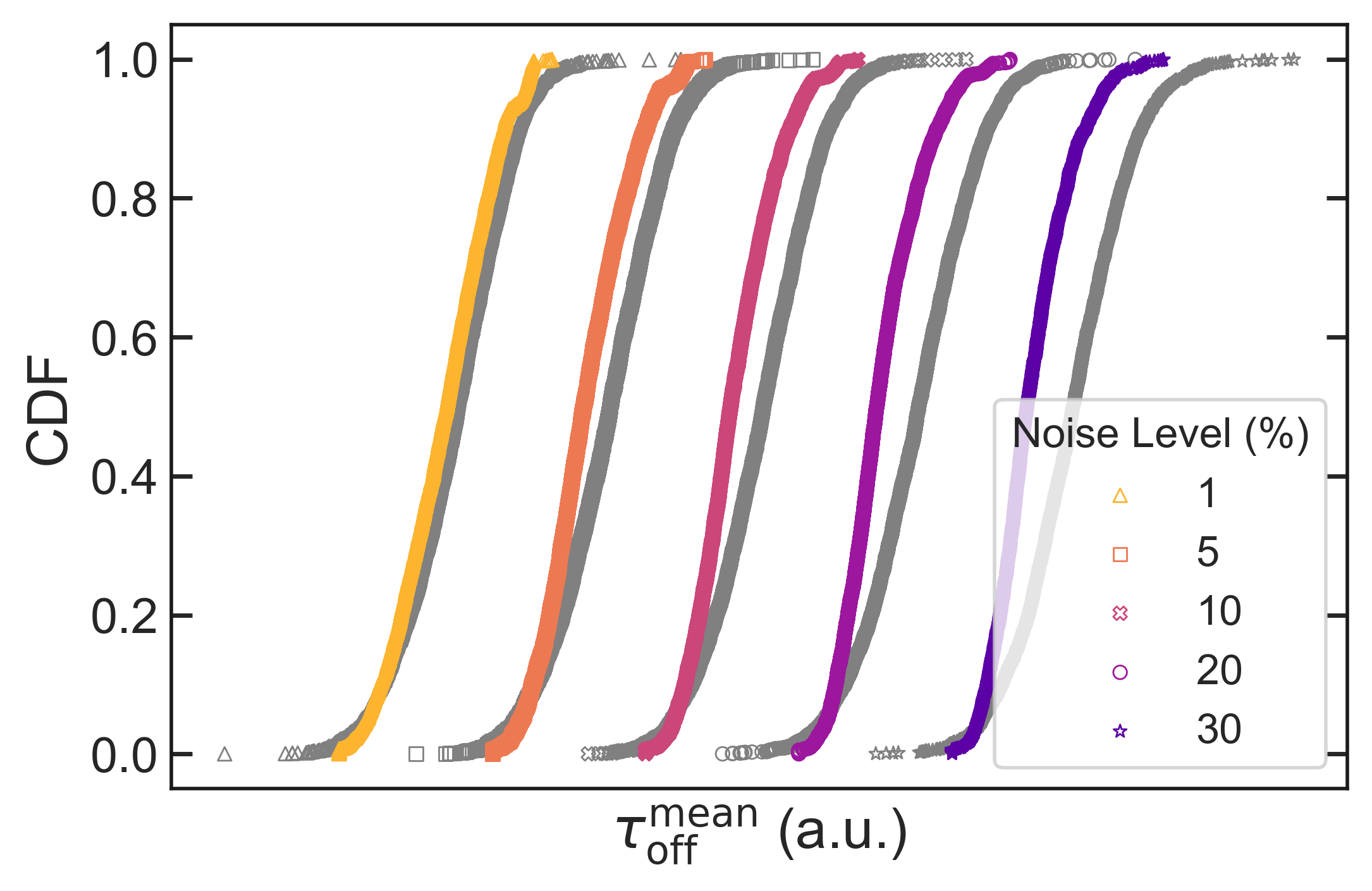}\label{fig_ntoff}}
	\subfigure[]{\includegraphics[width=0.32\textwidth]{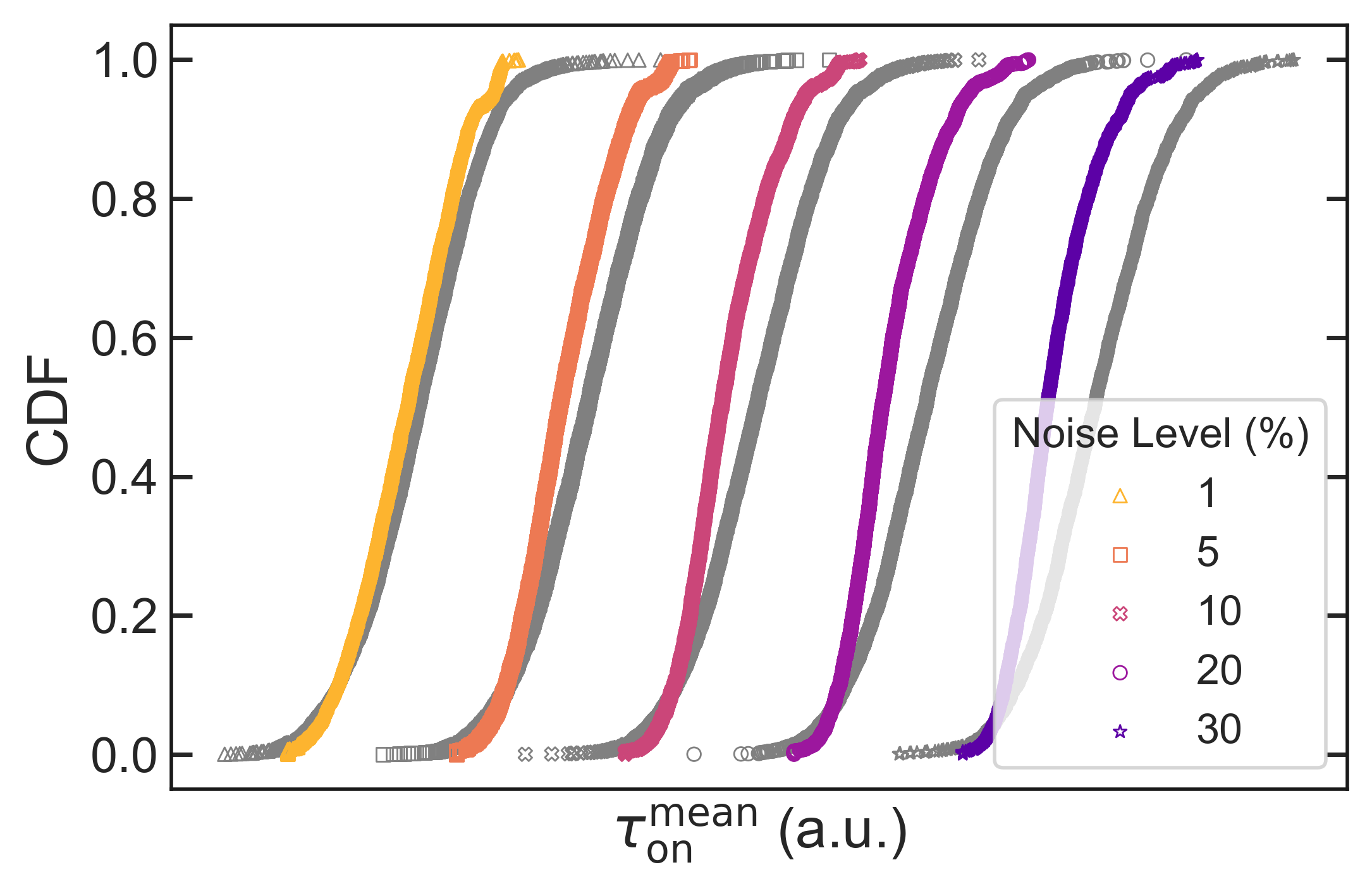}\label{fig_nton}}
	\vspace{-1em}
	\caption{\justifying Overall impact of noise level on \textit{RTNinja} performance. (a)~Histogram showing the frequency distribution of the difference between the estimated and actual number of sources across all datasets, grouped by noise level. A progressive leftward shift illustrates increasing underestimation as noise rises. (b)~Cumulative probability of source activity match accuracy, based on amplitude-centric input-output mapping. The total number of mapped sources and the fraction with $>$95\% activity match is provided. (c)~Complementary cumulative distribution function (CCDF) of estimated source amplitudes.  Input (true) amplitude distributions are shown in gray as a reference. (d--e)~Cumulative distributions of off-state and (e) on-state mean durations. True distributions are shown in gray. Deviations at higher noise levels reflect both missed sources and statistical compression due to limited sampling windows and temporal resolution.}\label{fig_statsall}
\end{figure*}

Figure~\ref{fig_ndiff} presents the cumulative frequency of the difference between the estimated and actual number of sources across noise levels. As anticipated, a systematic shift toward underestimation is observed with increasing noise, reflecting a higher fraction of missed sources as the noise floor rises. This trend is further supported by the activity match results shown in Figure~\ref{fig_ngof}, where activity accuracy — measured using the amplitude-based source matching strategy — declines inversely with noise. The primary source of this degradation lies in undetected sources whose activities are subsumed into the trajectories of detected ones. The overall distributions of mapped source characteristics — namely, amplitudes, off-state mean durations, and on-state mean durations — are summarized in Figures~\ref{fig_nampd}–\ref{fig_nton}. \textit{RTNinja} consistently recovers the underlying amplitude distributions with high accuracy, regardless of noise level. However, the total number of detected sources systematically declines as noise increases, particularly due to the suppression of low-amplitude components that fall below the elevated noise floor. At the other extreme, a minor deviation in amplitude estimation is occasionally observed for very high-amplitude sources. For both off-state ($\langle\tau_{off}^{n}\rangle$) and on-state ($\langle\tau_{on}^{n}\rangle$) mean durations, a clear trend of underestimation emerges with rising noise levels. This is primarily caused by the merging of undetected source activity into the traces of detected sources, leading to artificially faster dynamics. In addition, both the lower and upper tails of the extracted duration distributions are truncated across all noise conditions — a limitation imposed by the finite resolution and observation window of the RTN signal.

%\end{comment}

%\begingroup
%\color{red}
\section{Scope of applicability}
This section outlines the important conditions and constraints under which the \textit{RTNinja} framework can be effectively utilized. While the framework is designed to be flexible and robust across a range of tasks, its performance and reliability are contingent upon specific assumptions and data characteristics. Understanding these constraints is critical for practitioners to determine whether the framework is appropriate for a given problem and to anticipate potential limitations or challenges. The main guidelines that define the scope of applicability are as follows:\\
(i) Stationary signal: The framework assumes that the RTN signal remains stationary over the observation period. If device conditions drift over time or external interferences are present, the stationarity assumption may be violated, potentially compromising the accuracy of the analysis. To ensure reliable extraction of RTN parameters, such non-stationary components should be removed prior to applying the framework. A dedicated algorithm for drift correction has been recently reported and can be used in conjunction with this framework to enhance robustness~\cite{baseline}.\\
(ii) Two-state sources: The framework is well-suited for scenarios where each source behaves as a two-state system, allowing accurate identification and mapping of source dynamics. In cases where sources exhibit multi-state behavior, the \textit{SourcesMapper} module may interpret them as multiple independent two-state sources. Nevertheless, the \textit{LevelsExtractor} module remains applicable, as it can still reliably identify discrete levels. This consideration is particularly relevant for studies involving individual defects.\\
(iii) Independent source activity: The framework assumes that each source operates independently, with no electrical or physical coupling between them. This assumption enables reliable extraction and mapping of individual source dynamics, as the model treats each signal component as arising from a distinct, non-interacting source. Consequently, the framework does not account for coupling effects or correlated switching behavior. In experimental scenarios where source interactions are suspected, the interpretations of extracted parameters should therefore be made with caution.\\
The \textit{RTNinja} framework has already demonstrated its versatility and reliability through successful application to a variety of experimental RTN datasets across different device technologies. It has been effectively utilized for the analysis of nanoscale transistors and emerging device architectures, providing consistent and physically meaningful extraction of RTN parameters under diverse measurement conditions~\cite{piml,3dnand,cryocmos, wearout}. These successful applications highlight the framework’s robustness and adaptability while also validating the practical relevance of the assumptions outlined above. Collectively, they illustrate that within its defined scope, \textit{RTNinja} serves as a powerful and generalizable tool for RTN signal analysis in advanced device technologies.

%\endgroup

\section{Conclusion}
In this work, we presented \textit{RTNinja}, a generalized and fully automated machine learning framework for the analysis of random telegraph noise signals in nanoelectronic devices. By designing a modular architecture comprising the \textit{LevelsExtractor} and \textit{SourcesMapper} components, \textit{RTNinja} effectively deconvolves complex RTN signals into their constituent sources and activity profiles without requiring prior assumptions about the number or nature of these sources. A key strength of \textit{RTNinja} lies in its agnosticism to the underlying statistical distributions of source attributes and its independence from device-specific assumptions. This makes it broadly applicable across a wide range of nanoelectronic technologies and even beyond. The framework’s ability to operate without prior knowledge of the number or nature of RTN sources positions it as a powerful tool for both exploratory defect characterization and large-scale reliability benchmarking. Extensive Monte Carlo validation across 7000 datasets shows that \textit{RTNinja} delivers accurate, scalable, and automated analysis. It reliably reconstructs RTN signals and extracts source parameters, even in challenging scenarios with overlapping levels, undersampled sources, or high noise floors. \textit{RTNinja} unlocks new possibilities for applying RTN analysis in high-impact technology domains. In the semiconductor industry, it enables benchmarking of state-of-the-art devices — for instance, in dynamic random access memory components — by quantifying their susceptibility to defect-induced noise. This, in turn, guides circuit-level reliability assessments and memory architecture design. In quantum computing, where charge noise severely limits qubit coherence and fidelity, \textit{RTNinja} can uniquely characterize individual fluctuations, offering critical insights for designing noise-resilient gate operations and control protocols. These capabilities make \textit{RTNinja} a unique tool in RTN analysis but also a versatile platform for shaping the development, qualification, and adoption of next-generation semiconductor and quantum technologies.

\section*{Supplementary Material}
Please refer to the supplementary material for additional details on the \textit{RTNinja} framework. Supplementary Table 1 provides statistics on output yield across different noise levels. Supplementary Figure 1 gives additional information on the construction and clustering of the \(P_{T}-\Delta\) space. Supplementary Figure 2 provides additional information on the framework’s performance at 10\% and 20\% noise levels. Supplementary Figure 3 gives additional information on the difference between estimated and true source counts across noise levels and dataset types.

\begin{acknowledgments}
This work was part of the imec Industrial Affiliation Program. The authors thank Dr. Alexander Grill for helpful discussions during the \textit{RTNinja} framework development.
\end{acknowledgments}

\section*{Author Declarations}

\subsection*{Conflict of Interest}
The authors have no conflicts to disclose.

\subsection*{Author Contributions}
A.V. and R.D. conceptualized the work with assistance from C.M. A.V. generated the RTN data with support from R.D. A.V., R.D., and P.R. developed the \textit{RTNinja} framework. A.V. performed the data analysis, prepared the figures, and wrote the manuscript, with support from C.M. and R.D. All authors contributed to the discussions and interpretation of the results.

\section*{Code and Data Availability}
The synthetic datasets, \textit{RTNinja} output files, the code for generating datasets, and the code for generating statistics from output files are available at \href{https://doi.org/10.5281/zenodo.16750728}{10.5281/zenodo.16750728}. The code for the \textit{RTNinja} framework is currently part of a pending patent application (US 18/936,911; EP 23207819.6; CN 202411558787.3) and is therefore subject to legal restrictions.

\section*{References}
\bibliographystyle{unsrt} 
\bibliography{references}% Produces the bibliography via BibTeX.

@PREAMBLE{
 "\providecommand{\noopsort}[1]{}" 
 # "\providecommand{\singleletter}[1]{#1}%" 
}

@article{cmosscale,
	title={CMOS scaling trends and beyond},
	author={Bohr, Mark T and Young, Ian A},
	journal={IEEE Micro},
	volume={37},
	number={6},
	pages={20--29},
	year={2017},
	publisher={IEEE}
}

@article{ralls1984,
	title={Discrete resistance switching in submicrometer silicon inversion layers: Individual interface traps and low-frequency (1/f?) noise},
	author={Ralls, KS and Skocpol, WJ and Jackel, LD and Howard, RE and Fetter, LA and Epworth, RW and Tennant, DM},
	journal={Physical review letters},
	volume={52},
	number={3},
	pages={228},
	year={1984},
	publisher={APS}
}

@article{uren1985,
	title={1/f and random telegraph noise in silicon metal-oxide-semiconductor field-effect transistors},
	author={Uren, MJ and Day, DJ and Kirton, MJj},
	journal={Applied physics letters},
	volume={47},
	number={11},
	pages={1195--1197},
	year={1985},
	publisher={AIP Publishing}
}

@inproceedings{yaney1987meta,
	title={A meta-stable leakage phenomenon in DRAM charge storage-Variable hold time},
	author={Yaney, David S and Lu, Chih-Yuan and Kohler, Ross A and Kelly, Michael J and Nelson, James T},
	booktitle={1987 International Electron Devices Meeting},
	pages={336--339},
	year={1987},
	organization={IEEE}
}

@article{mueller1996conductance,
	title={Conductance modulation of submicrometer metal--oxide--semiconductor field-effect transistors by single-electron trapping},
	author={Mueller, HH and Schulz, M},
	journal={Journal of applied physics},
	volume={79},
	number={8},
	pages={4178--4186},
	year={1996},
	publisher={AIP Publishing}
}

@inproceedings{mueller1997statistics,
	title={Statistics of random telegraph noise in sub-$\mu$m MOSFETs},
	author={Mueller, HH and Schulz, M},
	booktitle={Proceedings of the 14th International Conference: Noise in Physical Systems and 1/f Fluctuations},
	volume={1},
	pages={195--200},
	year={1997},
	organization={World Scientific}
}

@inproceedings{fukuda2007random,
	title={Random telegraph noise in flash memories-model and technology scaling},
	author={Fukuda, Koichi and Shimizu, Yuui and Amemiya, Kazumi and Kamoshida, Masahiro and Hu, Chenming},
	booktitle={2007 IEEE International Electron Devices Meeting},
	pages={169--172},
	year={2007},
	organization={IEEE}
}

@inproceedings{qiu2015impact,
	title={Impact of random telegraph noise on write stability in Silicon-on-Thin-BOX (SOTB) SRAM cells at low supply voltage in sub-0.4 V regime},
	author={Qiu, Hao and Mizutani, Tomoko and Yamamoto, Yoshiki and Makiyama, Hideki and Yamashita, Tomohiro and Oda, Hidekazu and Kamohara, Shiro and Sugii, Nobuyuki and Saraya, Takuya and Kobayashi, Masaharu and others},
	booktitle={2015 Symposium on VLSI Technology (VLSI Technology)},
	pages={T38--T39},
	year={2015},
	organization={IEEE}
}

@inproceedings{goda2015time,
	title={Time dependent threshold-voltage fluctuations in NAND Flash memories: From basic physics to impact on array operation},
	author={Goda, Akira and Miccoli, Carmine and Compagnoni, Christian Monzio},
	booktitle={2015 IEEE International Electron Devices Meeting (IEDM)},
	pages={14--7},
	year={2015},
	organization={IEEE}
}

@book{grasser2020noise,
	title={Noise in nanoscale semiconductor devices},
	author={Grasser, Tibor},
	year={2020},
	publisher={Springer Nature}
}

@ARTICLE{fantini2007,
	author={Fantini, Paolo and Ghetti, Andrea and Marinoni, Andrea and Ghidini, Gabriella and Visconti, Angelo and Marmiroli, Andrea},
	journal={IEEE Electron Device Letters}, 
	title={Giant Random Telegraph Signals in Nanoscale Floating-Gate Devices}, 
	year={2007},
	volume={28},
	number={12},
	pages={1114-1116},
	doi={10.1109/LED.2007.909835}}

@inproceedings{tega2006anomalously,
	title={Anomalously large threshold voltage fluctuation by complex random telegraph signal in floating gate flash memory},
	author={Tega, Naoki and Miki, Hiroshi and Osabe, Taro and Kotabe, Akira and Otsuga, Kazuo and Kurata, Hideaki and Kamohara, Shiro and Tokami, Kenji and Ikeda, Yoshihiro and Yamada, Renichi},
	booktitle={2006 International Electron Devices Meeting},
	pages={1--4},
	year={2006},
	organization={IEEE}
}

@article{kurata2007random,
	title={Random telegraph signal in flash memory: Its impact on scaling of multilevel flash memory beyond the 90-nm node},
	author={Kurata, Hideaki and Otsuga, Kazuo and Kotabe, Akira and Kajiyama, Shinya and Osabe, Taro and Sasago, Yoshitaka and Narumi, Shunichi and Tokami, Kenji and Kamohara, Shiro and Tsuchiya, Osamu},
	journal={IEEE Journal of Solid-State Circuits},
	volume={42},
	number={6},
	pages={1362--1369},
	year={2007},
	publisher={IEEE}
}

@article{spinelli2021random,
	title={Random telegraph noise in 3D NAND flash memories},
	author={Spinelli, Alessandro S and Malavena, Gerardo and Lacaita, Andrea L and Monzio Compagnoni, Christian},
	journal={Micromachines},
	volume={12},
	number={6},
	pages={703},
	year={2021},
	publisher={MDPI}
}

@article{kaczer2016defect,
	title={The defect-centric perspective of device and circuit reliability—From gate oxide defects to circuits},
	author={Kaczer, Ben and Franco, Jacopo and Weckx, Pieter and Roussel, Ph J and Simicic, Marko and Putcha, Venkata and Bury, Erik and Cho, Moonju and Degraeve, Robin and Linten, Dimitri and others},
	journal={Solid-State Electronics},
	volume={125},
	pages={52--62},
	year={2016},
	publisher={Elsevier}
}

@article{simoen2011random,
	title={Random telegraph noise: From a device physicist's dream to a designer's nightmare},
	author={Simoen, Eddy and Kaczer, Ben and Toledano-Luque, Maria and Claeys, Cor},
	journal={ECS Transactions},
	volume={39},
	number={1},
	pages={3},
	year={2011},
	publisher={IOP Publishing}
}

@INPROCEEDINGS{matsumo2012,
	author={Matsumoto, Takashi and Kobayashi, Kazutoshi and Onodera, Hidetoshi},
	booktitle={2012 International Electron Devices Meeting}, 
	title={Impact of random telegraph noise on CMOS logic delay uncertainty under low voltage operation}, 
	year={2012},
	volume={},
	number={},
	pages={25.6.1-25.6.4},
	doi={10.1109/IEDM.2012.6479104}}

@ARTICLE{luo2015,
	author={Luo, Mulong and Wang, Runsheng and Guo, Shaofeng and Wang, Jing and Zou, Jibin and Huang, Ru},
	journal={IEEE Transactions on Electron Devices}, 
	title={Impacts of Random Telegraph Noise (RTN) on Digital Circuits}, 
	year={2015},
	volume={62},
	number={6},
	pages={1725-1732},
	doi={10.1109/TED.2014.2368191}}

@article{yuzhelevski2000random,
	title={Random telegraph noise analysis in time domain},
	author={Yuzhelevski, Y and Yuzhelevski, M and Jung, G},
	journal={Review of Scientific Instruments},
	volume={71},
	number={4},
	pages={1681--1688},
	year={2000},
	publisher={American Institute of Physics}
}

@inproceedings{nagumo2009new,
	title={New analysis methods for comprehensive understanding of random telegraph noise},
	author={Nagumo, T and Takeuchi, K and Yokogawa, S and Imai, K and Hayashi, Y},
	booktitle={2009 IEEE International Electron Devices Meeting (IEDM)},
	pages={1--4},
	year={2009},
	organization={IEEE}
}

@article{martin2014new,
	title={New weighted time lag method for the analysis of random telegraph signals},
	author={Martin-Martinez, Javier and Diaz, Javier and Rodriguez, Rosana and Nafria, Montserrat and Aymerich, Xavier},
	journal={IEEE Electron Device Letters},
	volume={35},
	number={4},
	pages={479--481},
	year={2014},
	publisher={IEEE}
}

@inproceedings{nagumo2010statistical,
	title={Statistical characterization of trap position, energy, amplitude and time constants by RTN measurement of multiple individual traps},
	author={Nagumo, Toshiharu and Takeuchi, Kiyoshi and Hase, Takashi and Hayashi, Yoshihiro},
	booktitle={2010 International Electron Devices Meeting},
	pages={28--3},
	year={2010},
	organization={IEEE}
}

@article{stampfer2018characterization,
	title={Characterization of single defects in ultrascaled MoS2 field-effect transistors},
	author={Stampfer, Bernhard and Zhang, Feng and Illarionov, Yury Yuryevich and Knobloch, Theresia and Wu, Peng and Waltl, Michael and Grill, Alexander and Appenzeller, Joerg and Grasser, Tibor},
	journal={ACS nano},
	volume={12},
	number={6},
	pages={5368--5375},
	year={2018},
	publisher={ACS Publications}
}

@article{stampfer2020semi,
	title={Semi-automated extraction of the distribution of single defects for nMOS transistors},
	author={Stampfer, Bernhard and Schanovsky, Franz and Grasser, Tibor and Waltl, Michael},
	journal={Micromachines},
	volume={11},
	number={4},
	pages={446},
	year={2020},
	publisher={MDPI}
}

@inproceedings{miki2012statistical,
	title={Statistical measurement of random telegraph noise and its impact in scaled-down high-$\kappa$/metal-gate MOSFETs},
	author={Miki, H and Tega, N and Yamaoka, M and Frank, David J and Bansal, Aditya and Kobayashi, M and Cheng, K and D'Emic, CP and Ren, Zhibin and Wu, S and others},
	booktitle={2012 International Electron Devices Meeting},
	pages={19--1},
	year={2012},
	organization={IEEE}
}

@article{puglisi2014factorial,
	title={Factorial hidden Markov model analysis of random telegraph noise in resistive random access memories},
	author={Puglisi, Francesco Maria and Pavan, Paolo},
	journal={ECTI Transactions on Electrical Engineering, Electronics, and Communications},
	volume={12},
	number={1},
	pages={24--29},
	year={2014}
}

@article{grill2019electrostatic,
	title={Electrostatic coupling and identification of single-defects in GaN/AlGaN Fin-MIS-HEMTs},
	author={Grill, Alexander and Stampfer, Bernhard and Im, Ki-Sik and Lee, J-H and Ostermaier, C and Ceric, Hajdin and Waltl, Michael and Grasser, Tibor},
	journal={Solid-State Electronics},
	volume={156},
	pages={41--47},
	year={2019},
	publisher={Elsevier}
}

@inproceedings{xiao2024new,
	title={A New Method of Automatic Extraction of RTN and OMI-Friendly Implementation},
	author={Xiao, Yu and Zhang, Chenyang and Wang, Da and Xue, Yongkang and Ren, Pengpeng and Ji, Zhigang},
	booktitle={2024 IEEE International Reliability Physics Symposium (IRPS)},
	pages={P75--TX},
	year={2024},
	organization={IEEE}
}

@article{gonzalez2020neural,
	title={Neural network based analysis of random telegraph noise in resistive random access memories},
	author={Gonz{\'a}lez-Cordero, Gerardo and Gonz{\'a}lez, MB and Morell, A and Jim{\'e}nez-Molinos, Francisco and Campabadal, Francesca and Rold{\'a}n, Juan B},
	journal={Semiconductor science and technology},
	volume={35},
	number={2},
	pages={025021},
	year={2020},
	publisher={IOP Publishing}
}

@inproceedings{xu2024deep,
	title={Deep Learning-Assisted Trap Extraction Method from Noise Power Spectral Density for MOSFETs},
	author={Xu, Jinghan and Zhou, Zheng and Fan, Mengqi and Sun, Zixuan and Wang, Shuhan and Tang, Zili and Liu, Fei and Liu, Xiaoyan},
	booktitle={2024 IEEE International Reliability Physics Symposium (IRPS)},
	pages={1--7},
	year={2024},
	organization={IEEE}
}

@incollection{fergus2022performance,
	title={Performance evaluation metrics},
	author={Fergus, Paul and Chalmers, Carl},
	booktitle={Applied Deep Learning: Tools, Techniques, and Implementation},
	pages={115--138},
	year={2022},
	publisher={Springer}
}

@inproceedings{vecchi2023unified,
	title={A Unified Framework to Explain Random Telegraph Noise Complexity in MOSFETs and RRAMs},
	author={Vecchi, Sara and Pavan, Paolo and Puglisi, Francesco Maria},
	booktitle={2023 IEEE International Reliability Physics Symposium (IRPS)},
	pages={1--6},
	year={2023},
	organization={IEEE}
}

@article{apc,
	author = {Brendan J. Frey  and Delbert Dueck },
	title = {Clustering by Passing Messages Between Data Points},
	journal = {Science},
	volume = {315},
	number = {5814},
	pages = {972-976},
	year = {2007},
	doi = {10.1126/science.1136800}
}

@article{hamming,
	title={Error detecting and error correcting codes},
	author={Hamming, Richard W},
	journal={The Bell system technical journal},
	volume={29},
	number={2},
	pages={147--160},
	year={1950},
	publisher={Nokia Bell Labs}
}

@inproceedings{baseline,
	title={Algorithm for Robust Correction of Long-Term Drift Components in Gate Leakage Current RTN Data},
	author={Varanasi, Anirudh and Degraeve, Robin and Roussel, Philippe J and Merckling, Clement},
	booktitle={2025 IEEE International Reliability Physics Symposium (IRPS)},
	pages={1--6},
	year={2025},
	organization={IEEE}
}

@inproceedings{piml,
	title={Physics-informed machine learning to analyze oxide defect-induced RTN in gate leakage current},
	author={Varanasi, Anirudh and Degraeve, Robin and Roussel, Philippe J and Vici, Andrea and Merckling, Clement},
	booktitle={2024 IEEE International Reliability Physics Symposium (IRPS)},
	pages={1--7},
	year={2024},
	organization={IEEE}
}

@inproceedings{3dnand,
	title={Advanced RTN Analysis on 3D NAND Trench Devices Using Physics-Informed Machine Learning Framework},
	author={Higashi, Y and Varanasi, A and Roussel, PJ and Canflanca, PS and Bastos, JP and Grill, A and Catapano, E and Asanovski, R and Franco, J and Kaczer, B and others},
	booktitle={2025 IEEE International Reliability Physics Symposium (IRPS)},
	pages={1--6},
	year={2025},
	organization={IEEE}
}

@article{cryocmos,
	title={Investigation of Low Frequency Noise in CryoCMOS devices through Statistical Single Defect Spectroscopy},
	author={Catapano, Edoardo and Varanasi, Anirudh and Roussel, Philippe and Degraeve, Robin and Higashi, Yusuke and Asanovski, Ruben and Kaczer, Ben and Fortuny, Javier Diaz and Waltl, Michael and Afanasiev, Valeri and others},
	journal={arXiv preprint arXiv:2505.04030},
	year={2025}
}

@article{wearout,
	title = {Markov model describing progressive degradation of local percolation path in thin oxides},
	author = {Sara Sacchi and Anirudh Varanasi and Robin Degraeve and Andrea Vici and Giorgio Molinaro and Jacopo Franco and Philippe Roussel and Ben Kaczer and Clement Merckling},
	journal = {Solid-State Electronics},
	volume = {231},
	pages = {109266},
	year = {2026},
	publisher={Elsevier}
}
%\printbibliography

\clearpage

\onecolumngrid
\begin{center}
	{\large\bfseries Supplementary Material}
\end{center}
\vspace{1em}

%\captionsetup[table]{skip=10pt} % Add vertical space between caption and table
\begin{table*}[htbp!]
	\centering
	\vspace{5pt} % Optional: add more space below the caption if needed
	\begin{tabular}{ccccc}
		\toprule
		Noise Level (\%) & \makecell{\#Datasets with\\ \textit{RTNinja} Convergence} & \#Datasets Yield (\%) & \#Detected Sources & \makecell{\#Detected Sources\\Yield(\%)} \\
		\midrule
		1   & 1234 & 88.1 & 4135 & 73.8 \\
		5   & 1308 & 93.4 & 4076 & 72.8 \\
		10  & 1303 & 93.1 & 3632 & 64.9 \\
		20  & 1254 & 89.6 & 2986 & 53.3 \\
		30  & 1207 & 86.2 & 2515 & 44.9 \\
		\bottomrule
	\end{tabular}
	\caption*{\justifying Supplementary Table 1: Statistics on \textit{RTNinja} output yield across different noise levels. The total number of datasets is 1400, while the total count of individual sources is 5600.} % Manually named caption
\end{table*}

%\renewcommand{\thefigure}{S1}
%\captionsetup[subfigure]{font={normal}, skip=0pt, margin=0cm, singlelinecheck=false}
\begin{figure*} %[htbp!]
	\subfigure[]{\includegraphics[width=0.49\textwidth]{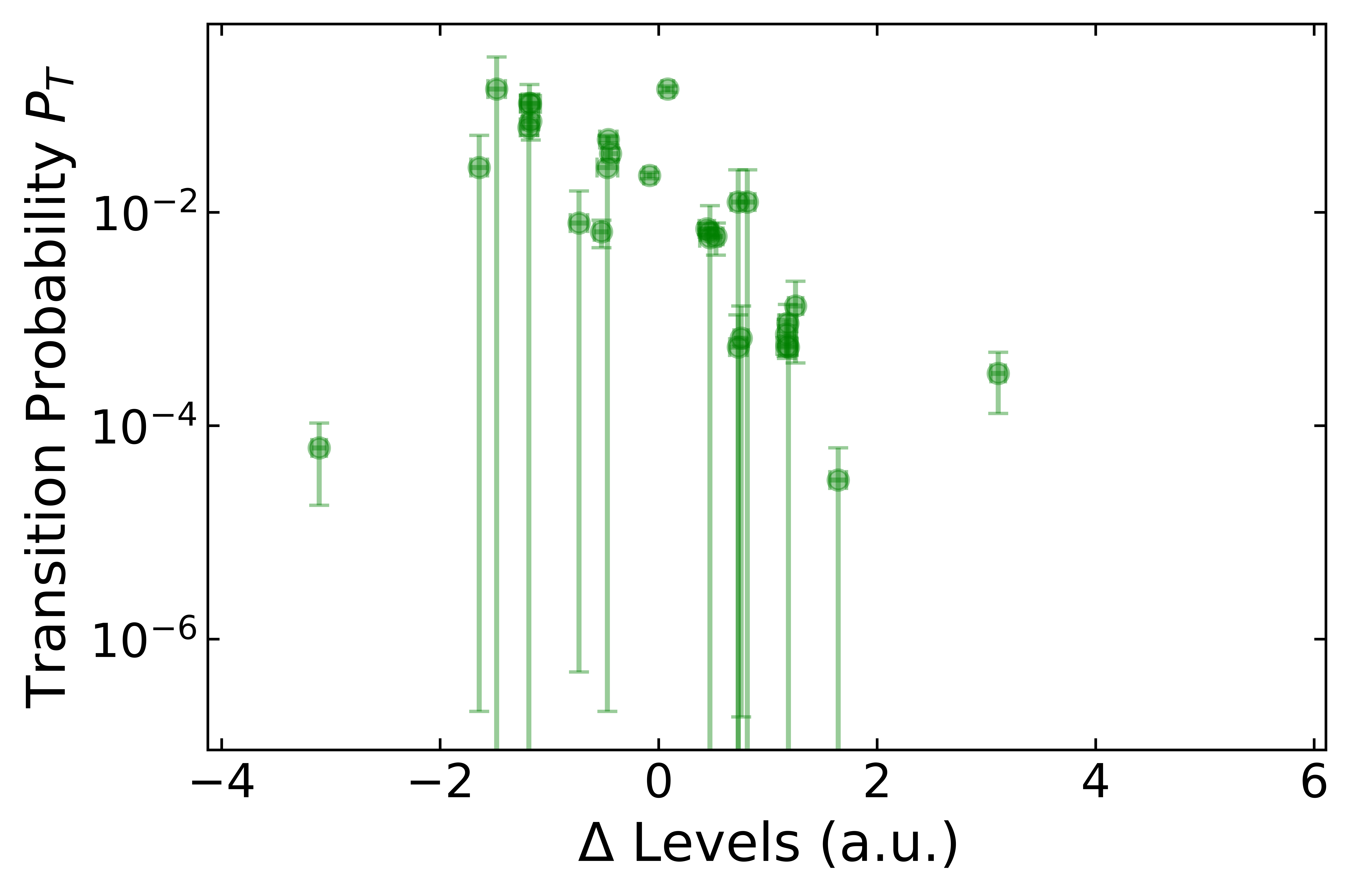}\label{fig_cluster2}}
	\subfigure[]{\includegraphics[width=0.49\textwidth]{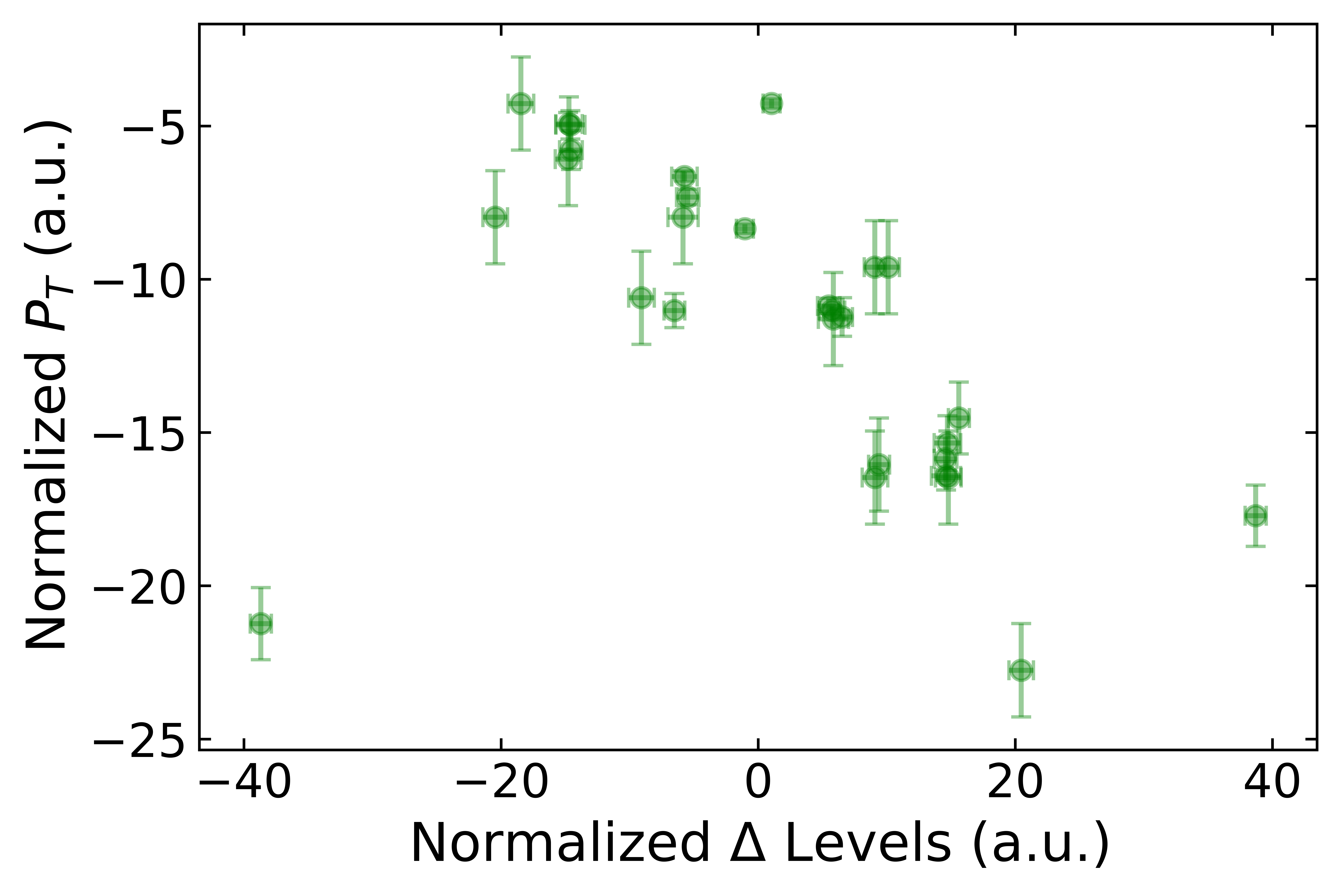}\label{fig_cluster3}}
	\subfigure[]{\includegraphics[width=0.99\textwidth]{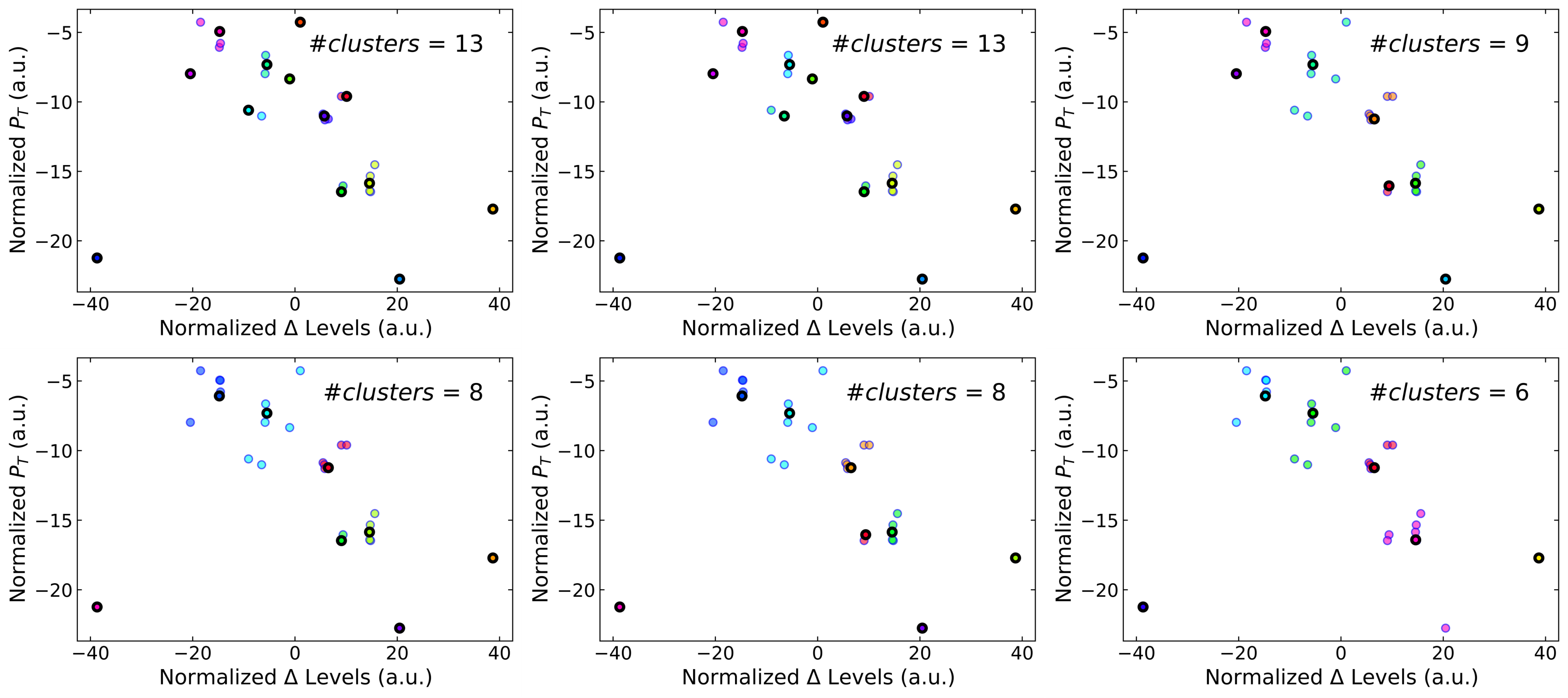}\label{fig_cluster4}}
	\caption*{\justifying Supplementary Figure 1: Construction and clustering of the \(P_{T}-\Delta\) space for source amplitude estimation.(a)~Two-dimensional \(P_{T}-\Delta\) space showing transition probabilities between levels and potential source amplitudes, defined as the difference between mean values of the distinct levels. (b)~Normalized \(P_{T}-\Delta\) space, where both dimensions are scaled by the median average error to ensure equal weighting, enabling unbiased clustering. (c)~Affinity propagation clustering is applied to the normalized space using a range of hyperparameter settings. The resulting slight variations in cluster assignments are aggregated into an ensemble of one-dimensional amplitude estimates, reducing sensitivity to clustering hyperparameters and enhancing robustness.}\label{fig_clustering2}
\end{figure*}

\setcounter{subfigure}{0}
%\renewcommand{\thefigure}{S2}
%\captionsetup[subfigure]{font={normal}, skip=0pt, margin=0cm, singlelinecheck=false}
\begin{figure*} %[htbp!]
	\centering
	\makebox[\textwidth][c]{%
		\begin{minipage}{0.33\textwidth}
			\centering Noise Level: 10\%
		\end{minipage}%
		\begin{minipage}{0.33\textwidth}
			\centering 20\%
	\end{minipage}}
	%\vspace{-1em}
	\subfigure[]{\includegraphics[width=0.32\textwidth]{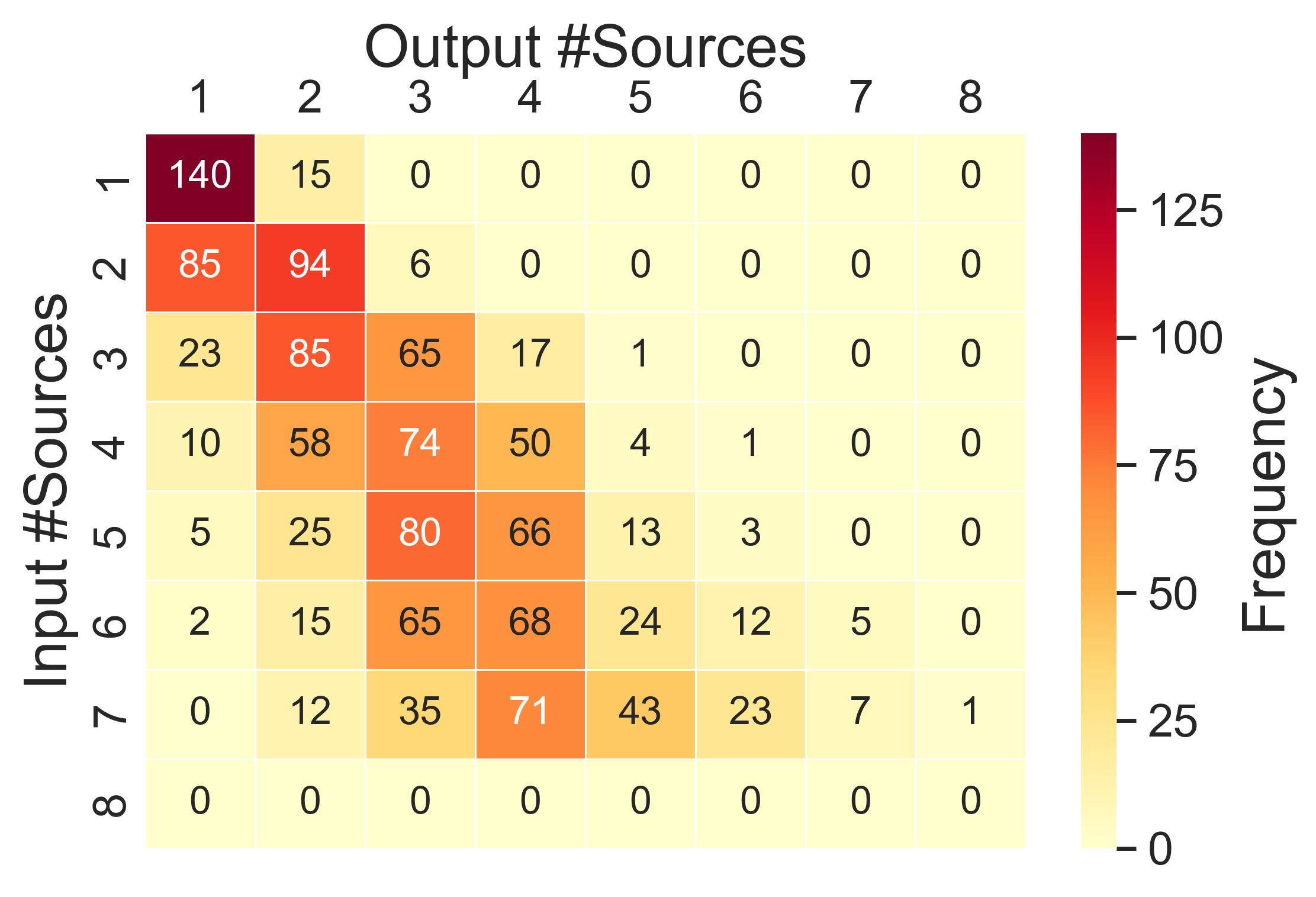}\includegraphics[width=0.32\textwidth]{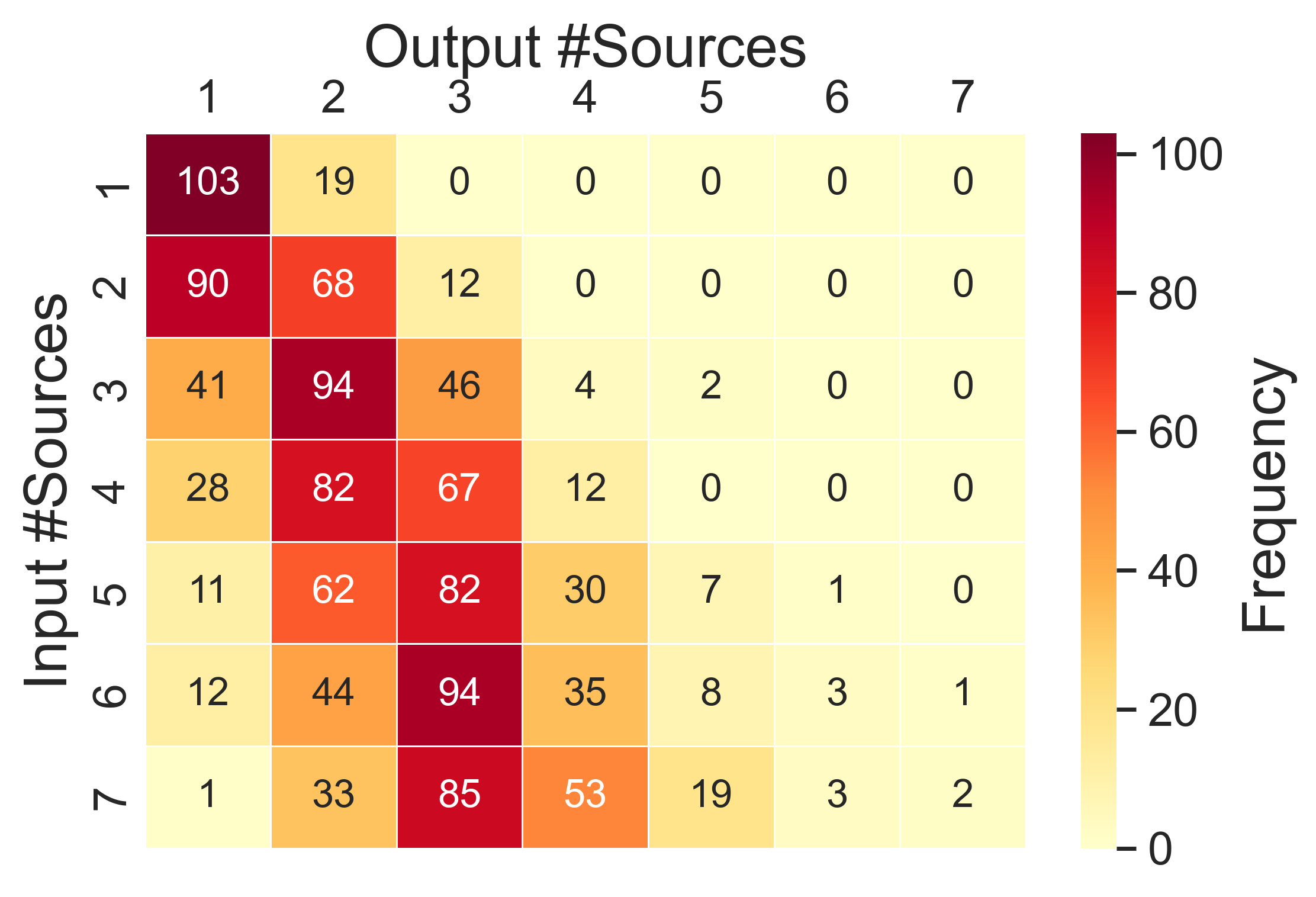}}
	%\vspace{-1em}
	\subfigure[]{\includegraphics[width=0.32\textwidth]{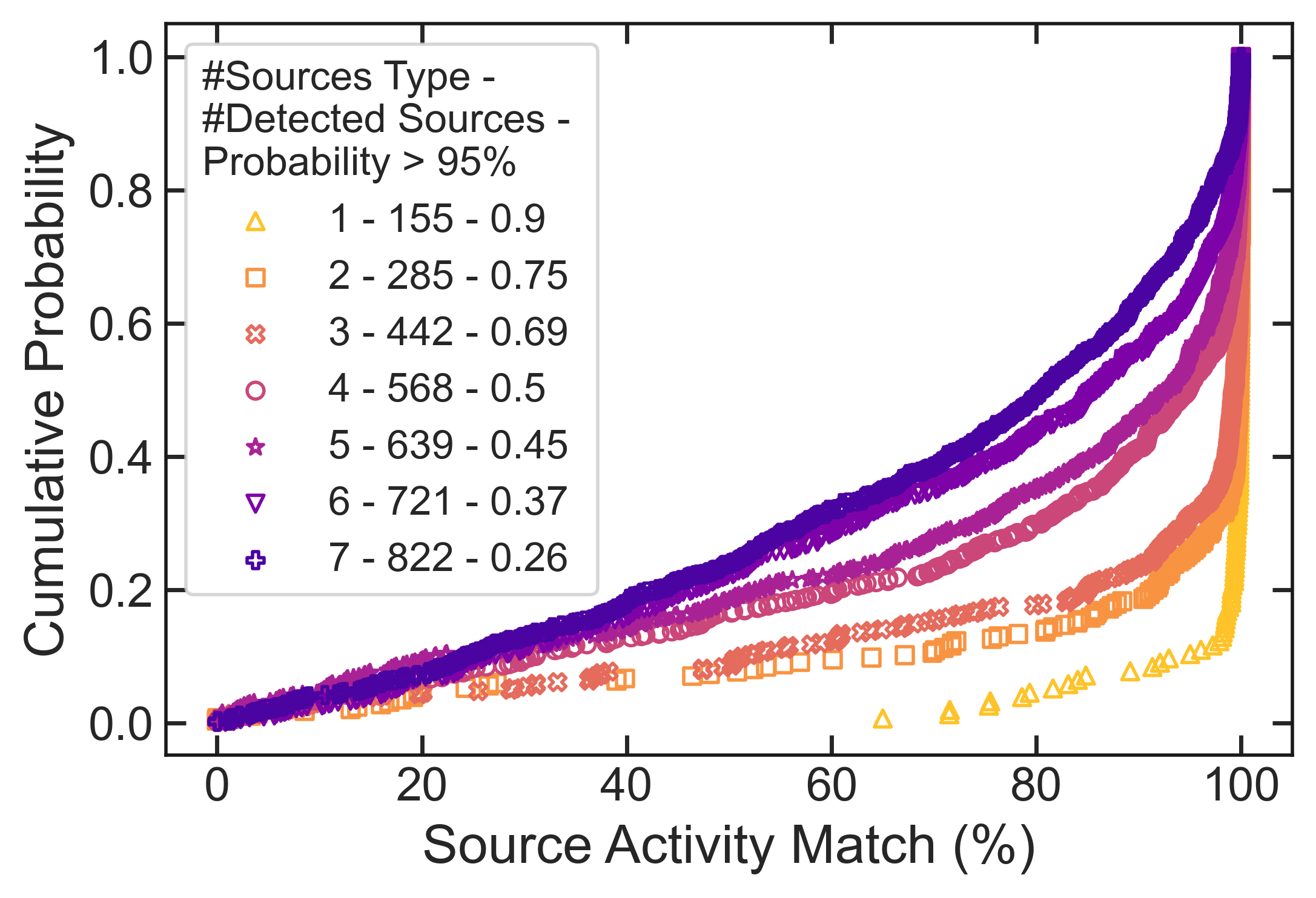}\includegraphics[width=0.32\textwidth]{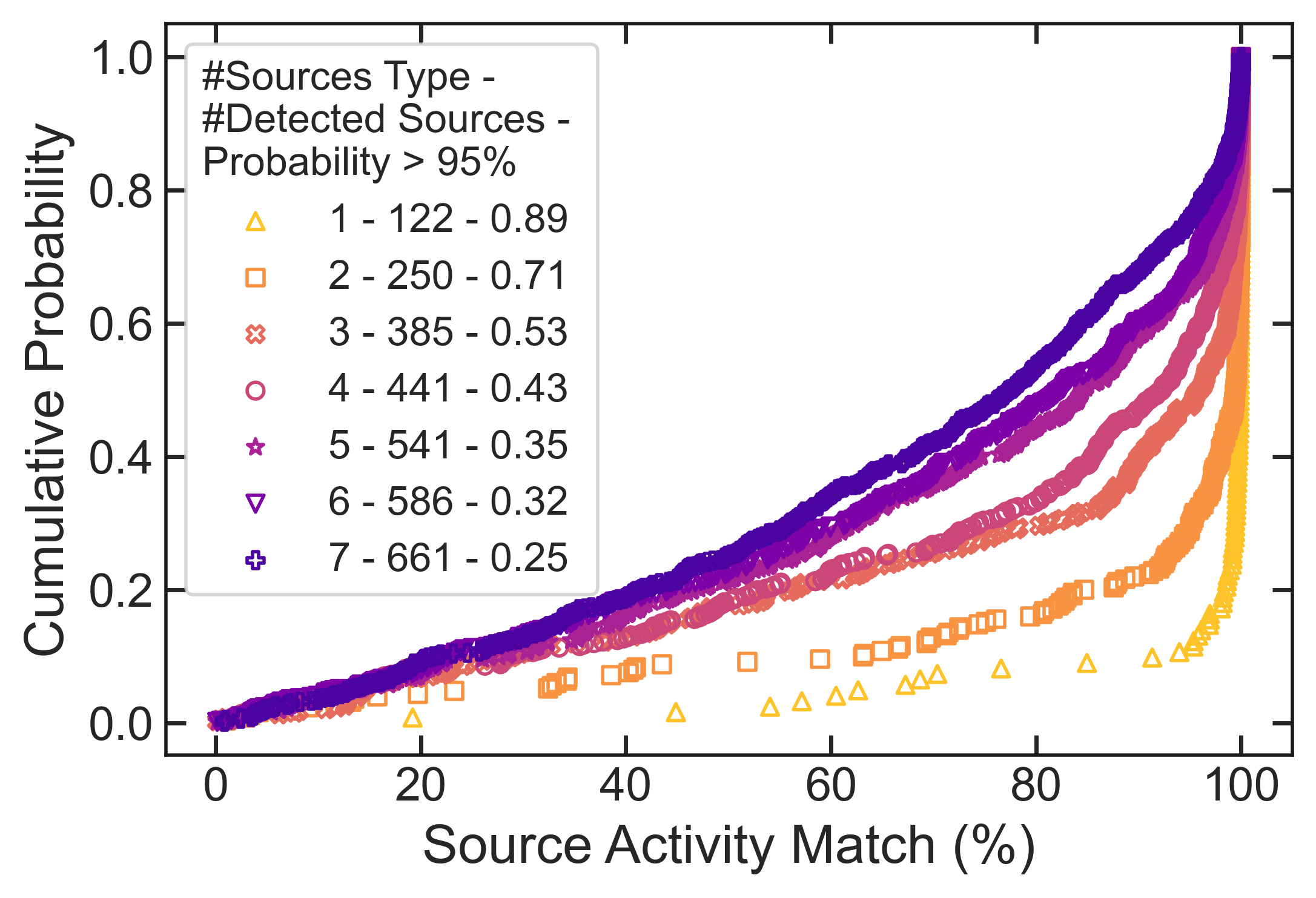}}
	%\vspace{-1em}
	\subfigure[]{\includegraphics[width=0.32\textwidth]{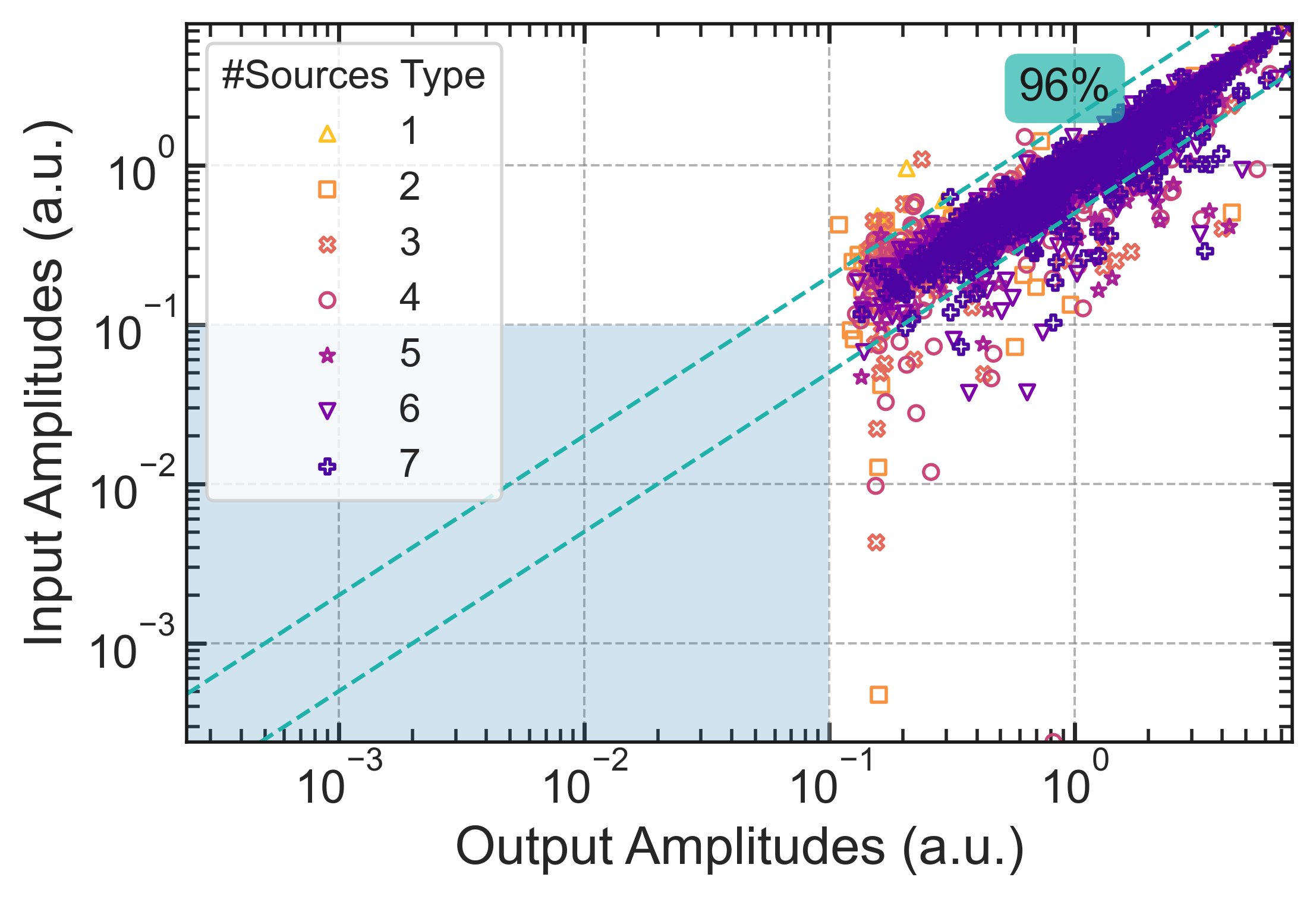}\includegraphics[width=0.32\textwidth]{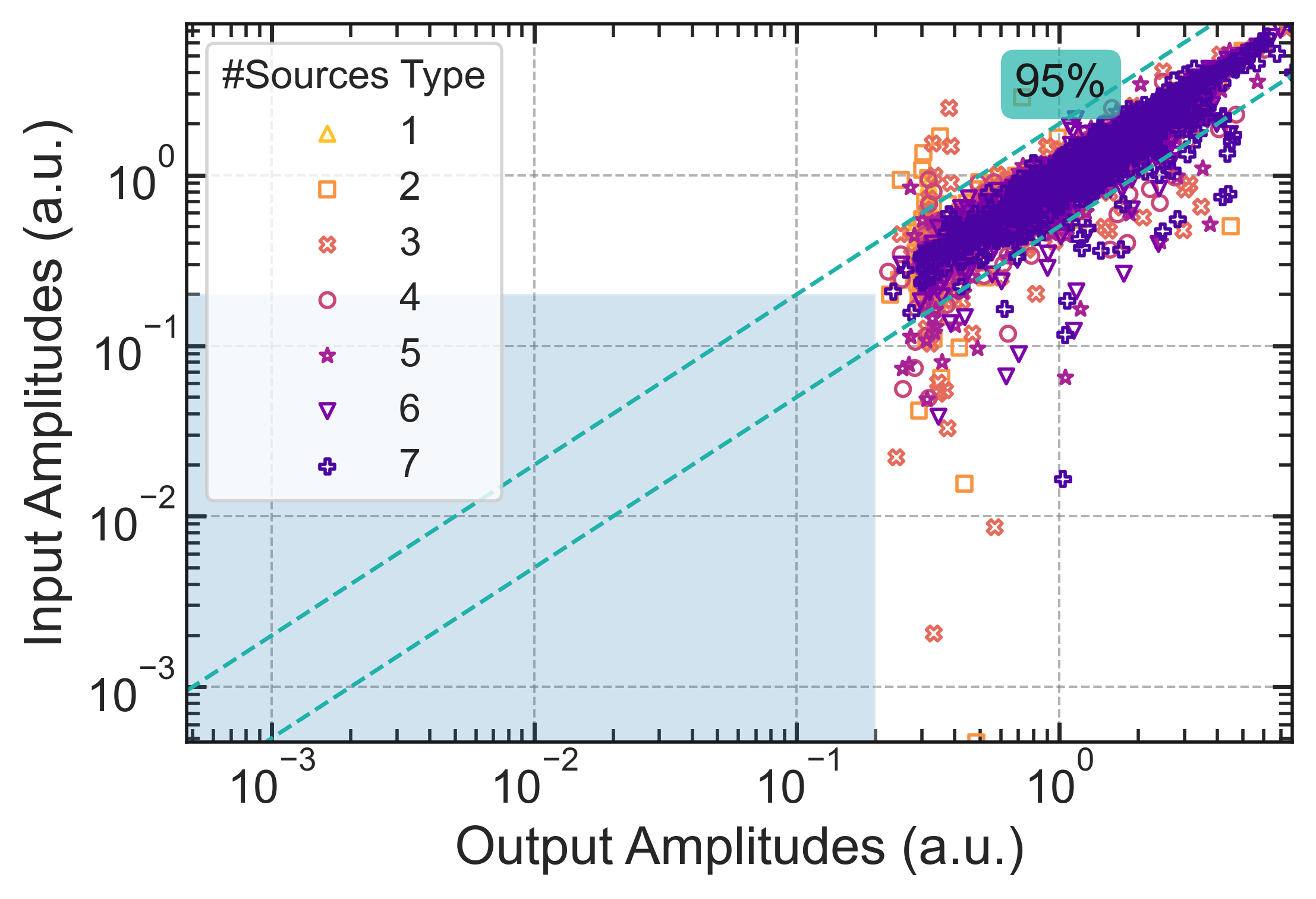}}
	%\vspace{-1em}
	\subfigure[]{\includegraphics[width=0.32\textwidth]{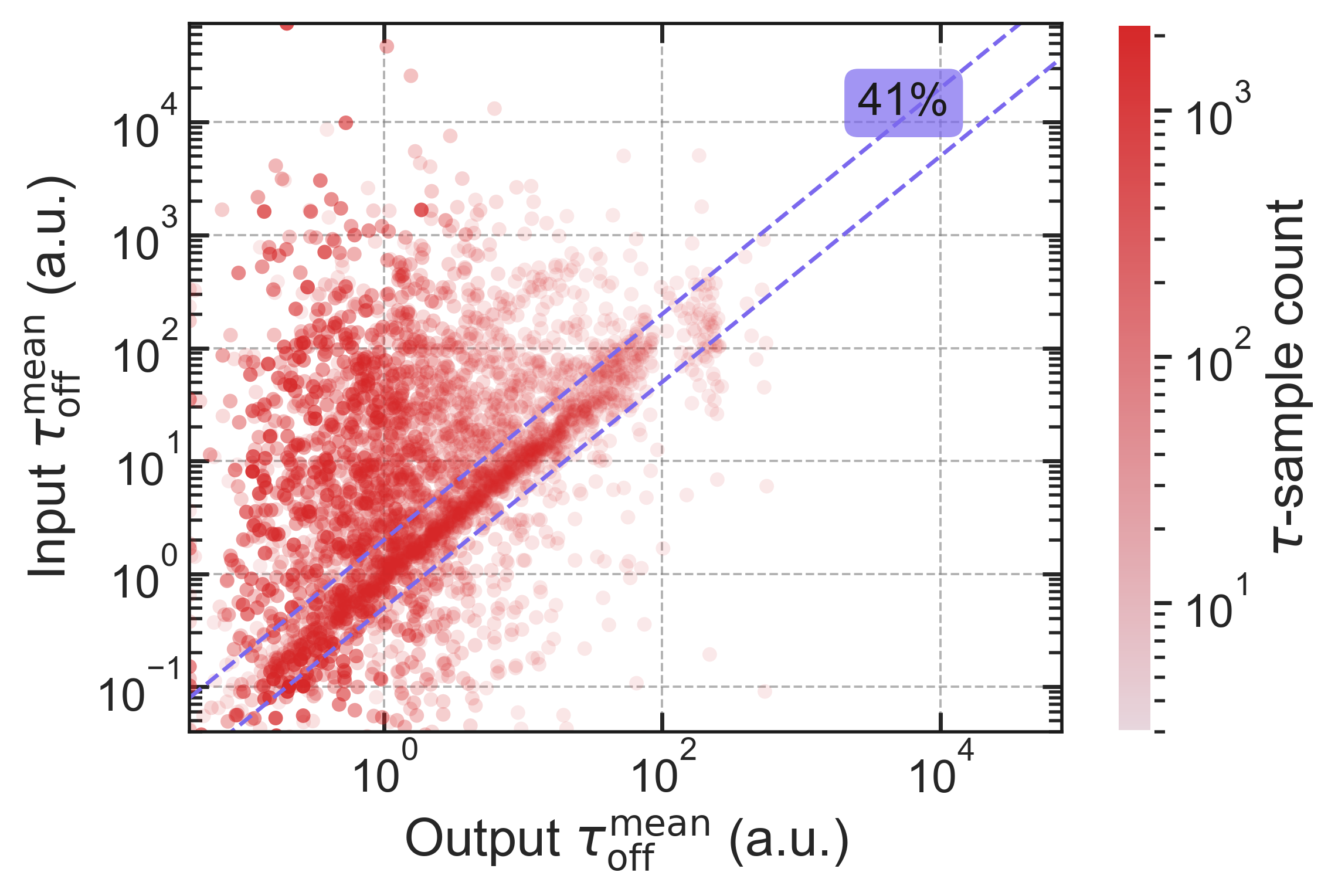}\includegraphics[width=0.32\textwidth]{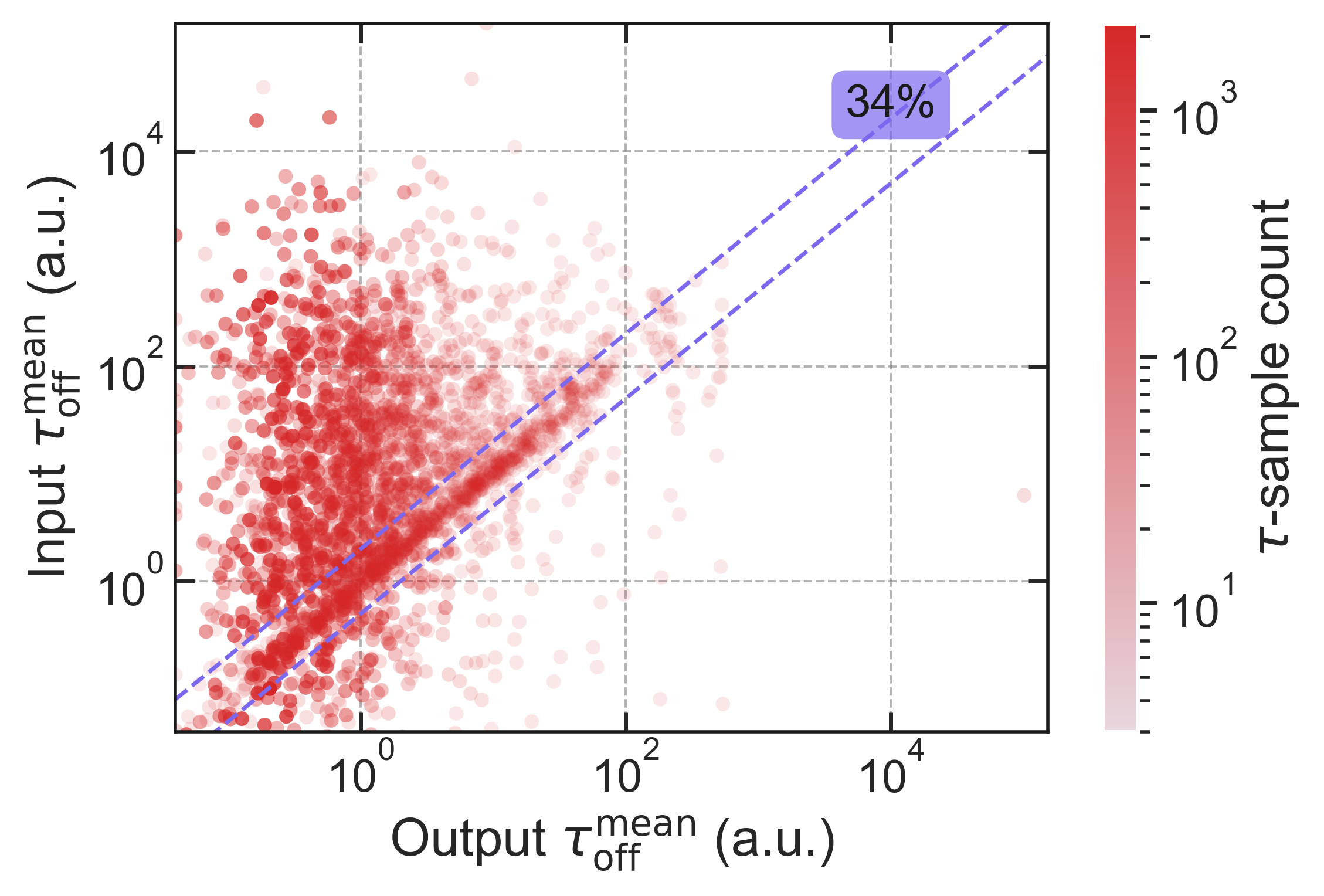}}
	%\vspace{-1em}	
	\subfigure[]{\includegraphics[width=0.32\textwidth]{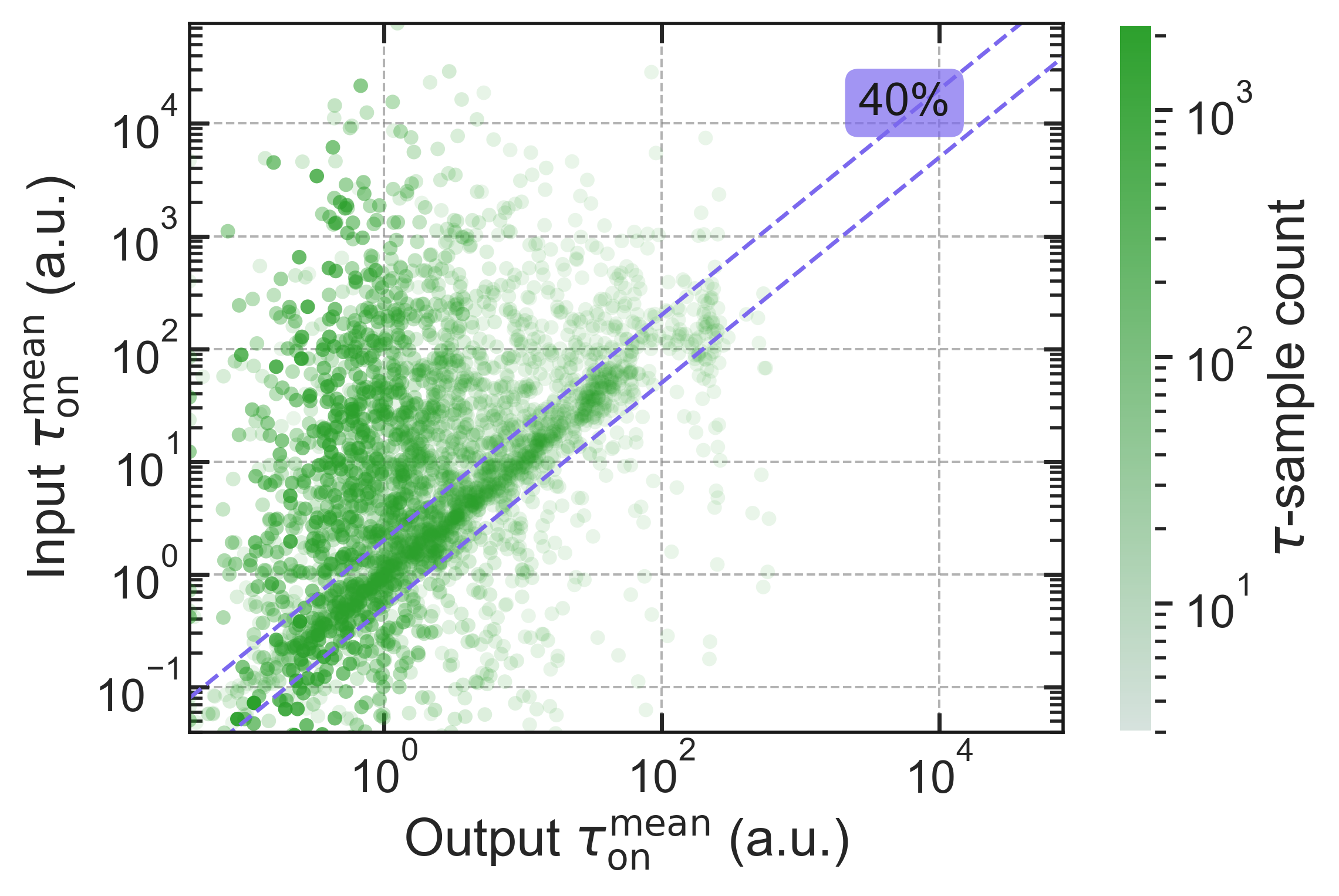}\includegraphics[width=0.32\textwidth]{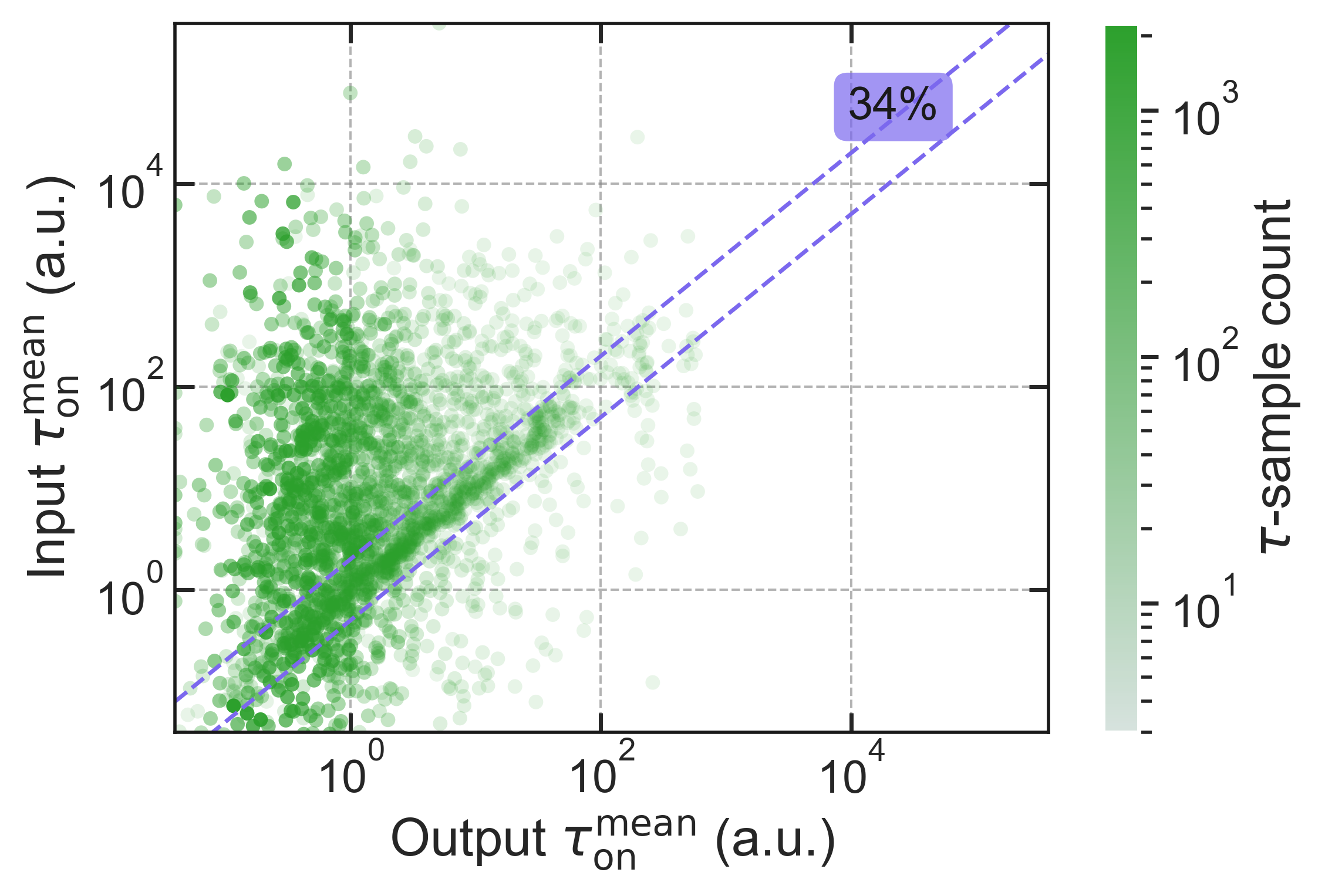}}
	%\vspace{-0.5em}
	\caption*{\justifying Supplementary Figure 2: Performance of the \textit{RTNinja} framework at 10\% and 20\% noise levels. The results align with the overall trends discussed in the main text, in line with observations at 1\%, 5\%, and 30\% noise levels.}\label{fig_stats2}
\end{figure*}

\setcounter{subfigure}{0}
%\renewcommand{\thefigure}{S3}
%\captionsetup[subfigure]{font={normal}, skip=0pt, margin=0cm, singlelinecheck=false}
\begin{figure*} %[htbp!]
	\subfigure[Noise Level: 1\%]{\includegraphics[width=0.32\textwidth]{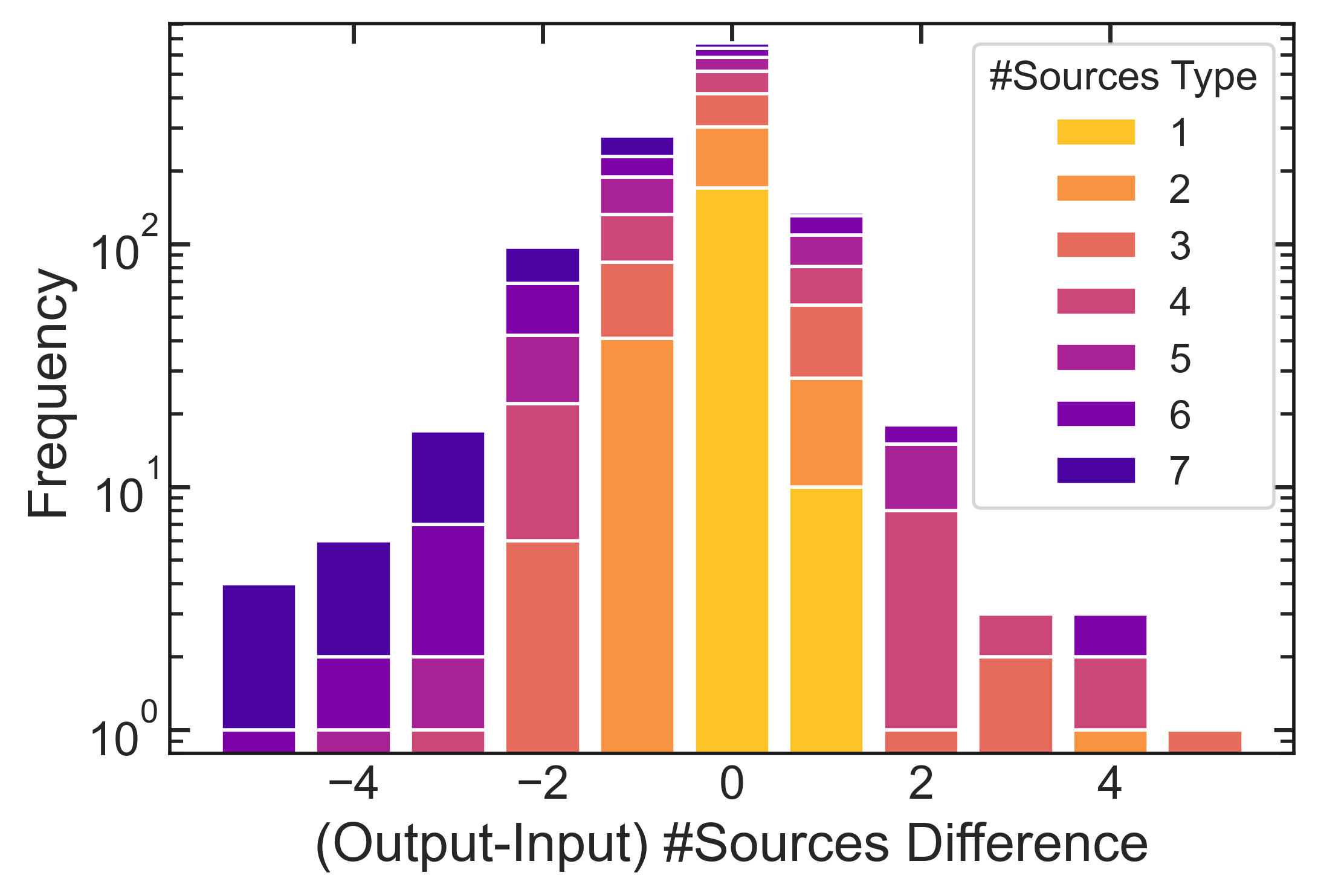}}
	\subfigure[5\%]{\includegraphics[width=0.32\textwidth]{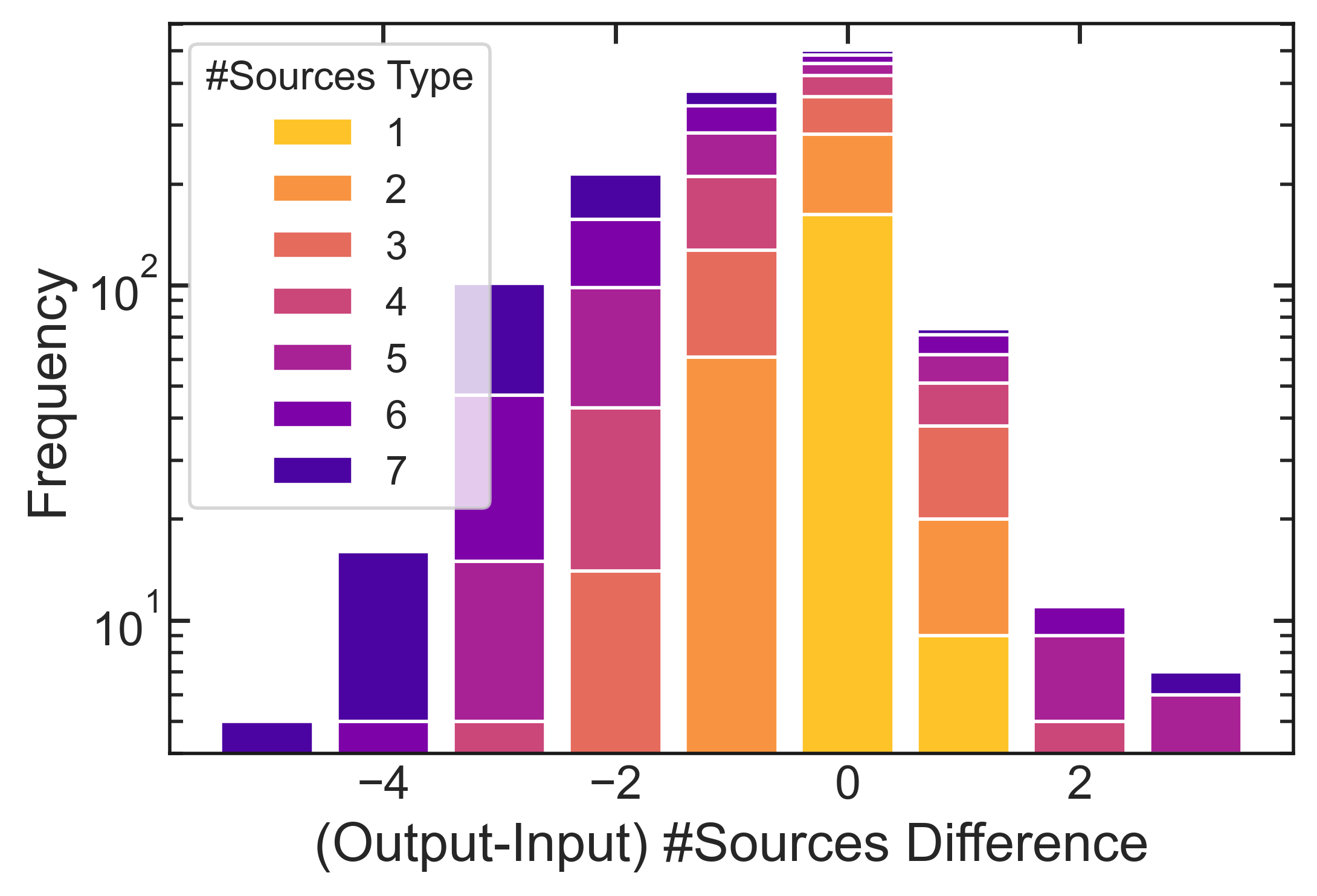}}
	\subfigure[10\%]{\includegraphics[width=0.32\textwidth]{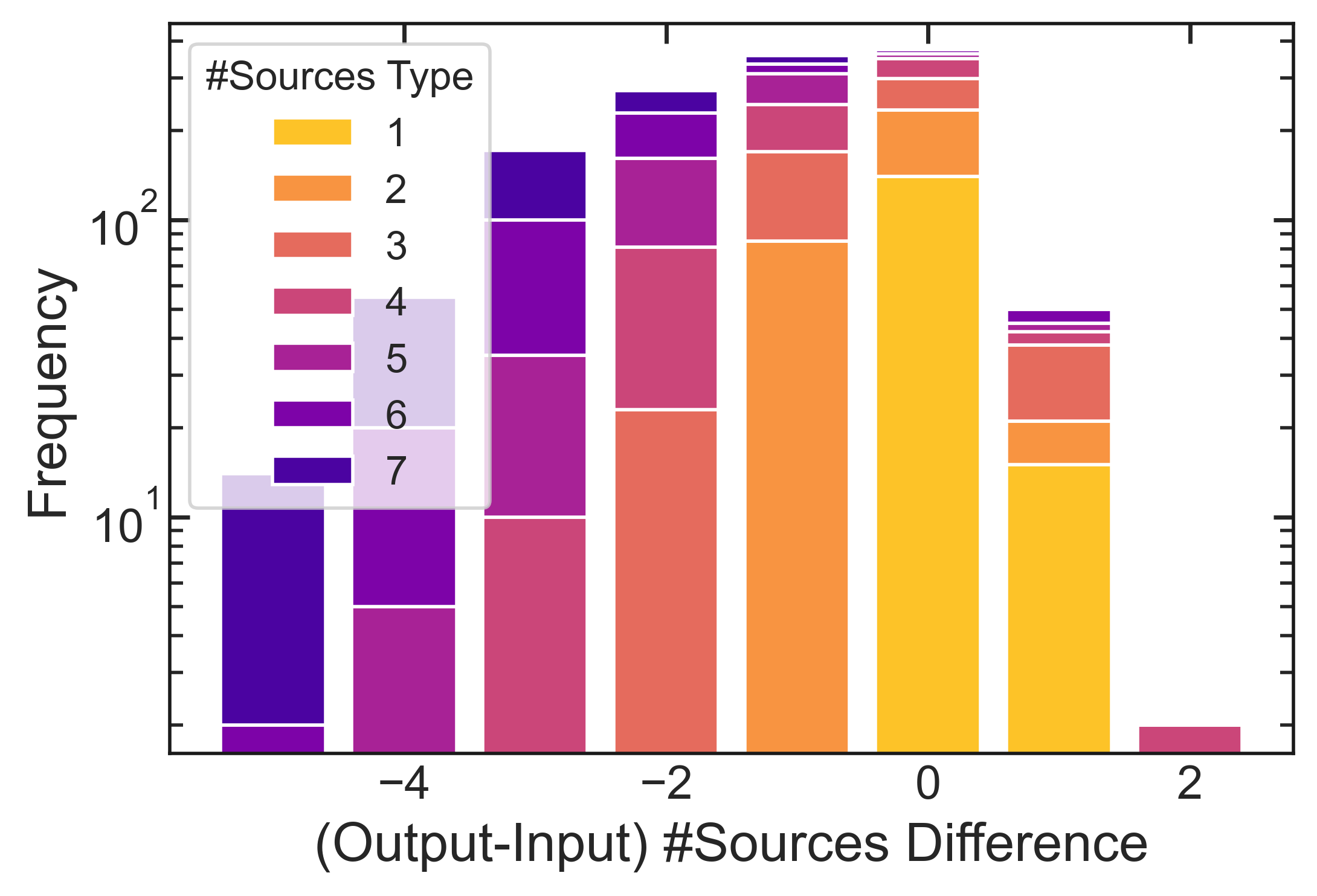}}
	\subfigure[20\%]{\includegraphics[width=0.32\textwidth]{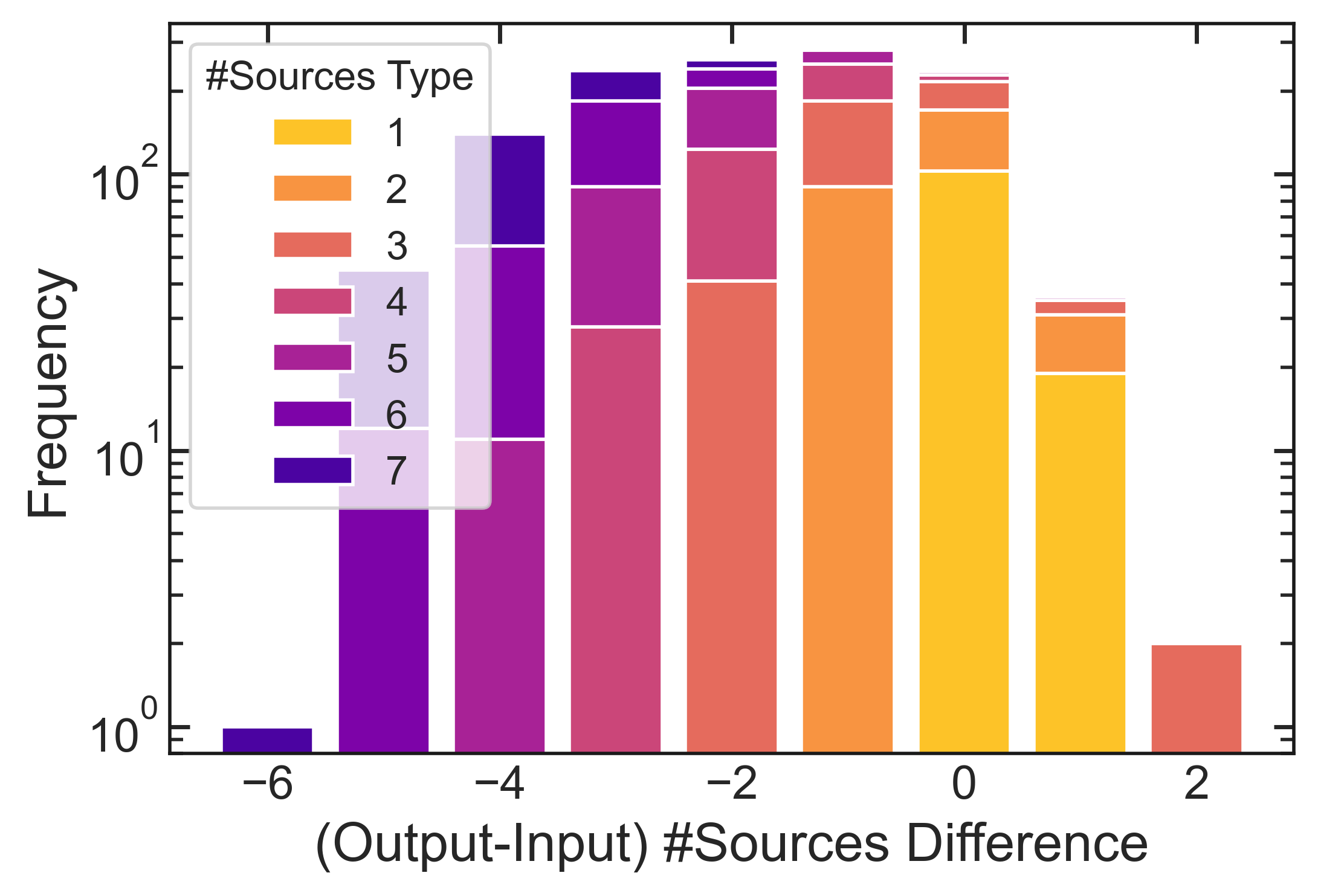}}
	\subfigure[30\%]{\includegraphics[width=0.32\textwidth]{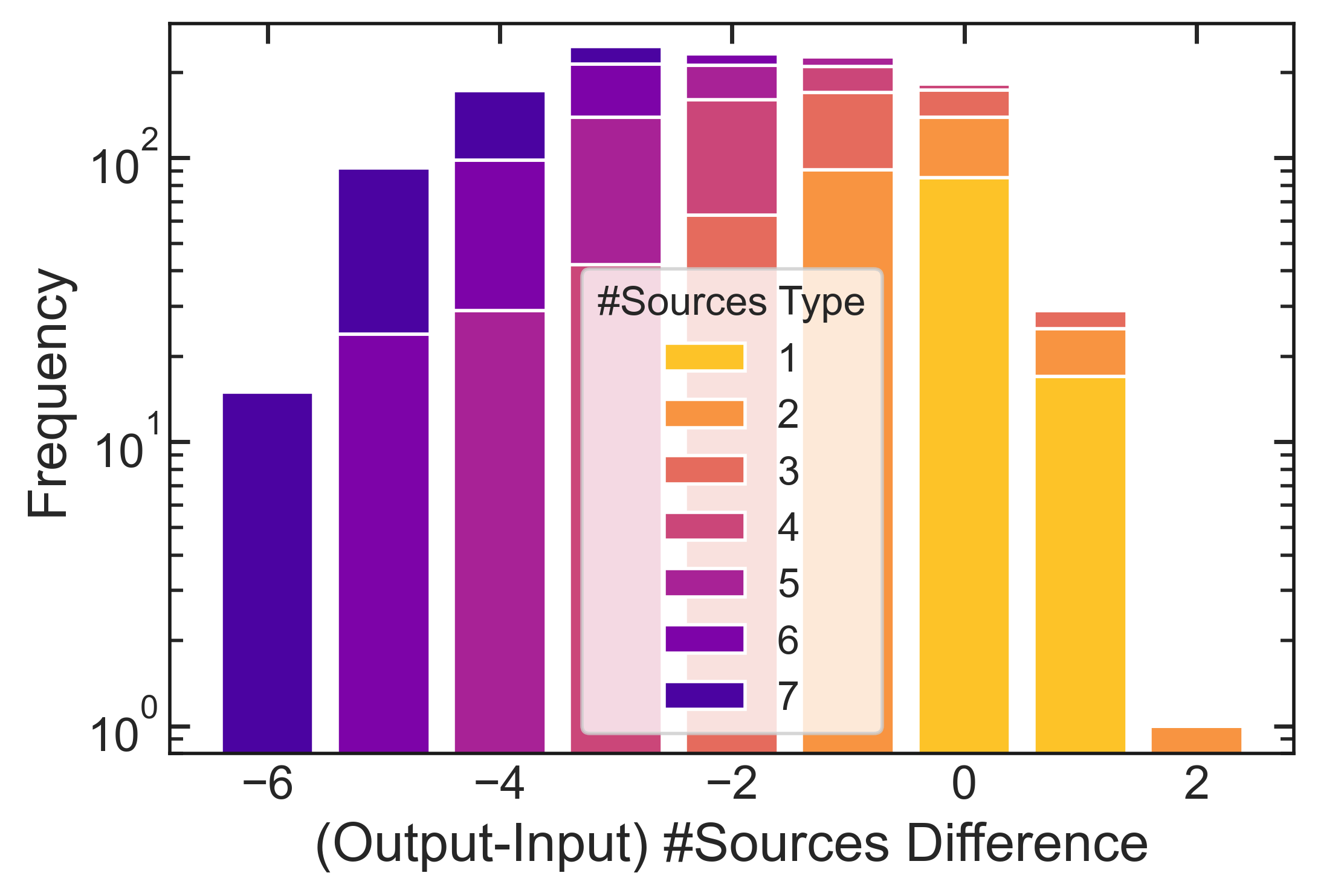}}
	\caption*{\justifying Supplementary Figure 3: Difference between the number of sources estimated by \textit{RTNinja} and the ground truth, shown across noise levels and grouped by $N$-source dataset type. A systematic shift toward negative differences is observed, indicating an increasing fraction of undetected sources as the noise level rises.}\label{fig_stats3}
\end{figure*}

\end{document}